%% file: main.tex
\documentclass[11pt,a4paper]{report}
\usepackage{setspace}

\usepackage[left=3cm, right=2.5cm]{geometry}
\usepackage{graphicx}
\usepackage{enumitem}
\usepackage[
  backend=biber,
  sorting=nyt, 
  maxcitenames=2, 
  maxbibnames=99  
]{biblatex}
\usepackage{subfiles}

\usepackage{commands}

\newcommand{\unchapter}[1]{%
  \begingroup
  \let\@makechapterhead\@gobble 
  \chapter{#1}
  \endgroup
}
\makeatother

\graphicspath{ {./img/} }
\begin{document}

\pagenumbering{roman}
\thispagestyle{empty}
\begin{figure}
\includegraphics[width=0.3\columnwidth]{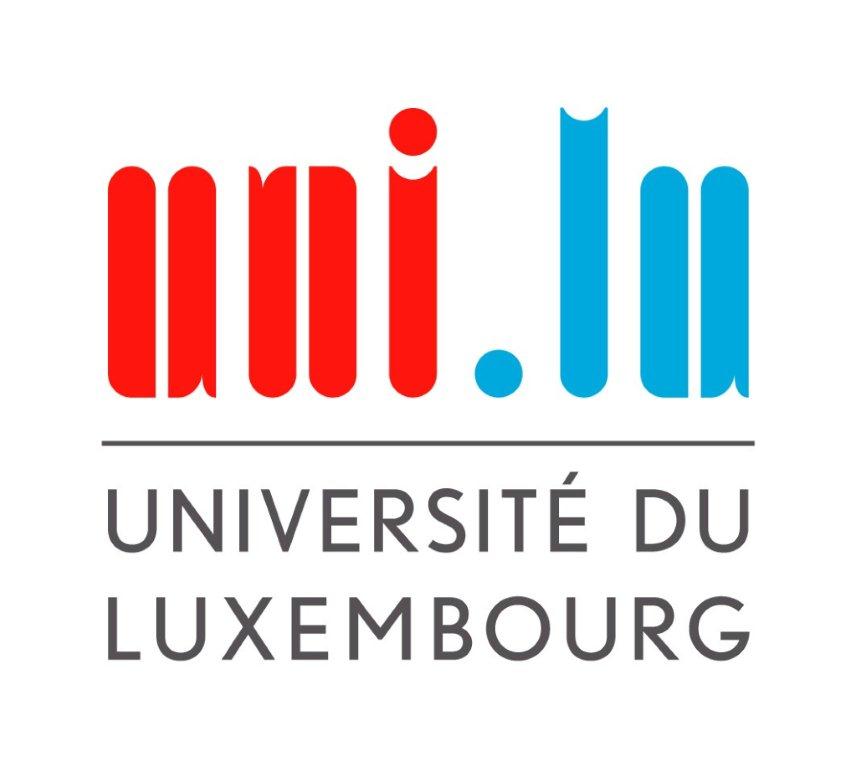}
\centering
\end{figure}

\begin {center}
\textbf{\footnotesize{PhD-No.....XX}}\\
\vspace{0.4cm}
\textbf{\footnotesize{Faculty of Sciences, Technology, and Medicine}}\\
\vspace{2.2cm}
\large{DISSERTATION}\\
\vspace{0.3cm}
Presented on 16th March in Luxembourg\\
\vspace{0.3cm}
to obtain the degree of\\
\vspace{0.3cm}
\textbf{\Large{DOCTEUR DE L'UNIVERSITE DU LUXEMBOURG}}\\
\vspace{0.5cm}
\textbf{\Large{EN INFORMATIQUE}}\\
\vspace{1.1cm}
by\\
\vspace{0.3cm}
\textbf{Benoît Alcaraz}\\
\vspace{0.3cm}
Born on 13th January 1998 in Bourgoin-Jallieu (France)\\
\vspace{1.6cm}

\textbf{\Large{What if Pinocchio Were a Reinforcement Learning Agent: A Normative End-to-End Pipeline}}

\end {center}

\newpage
\thispagestyle{empty}

\begin{figure}
\includegraphics[width=0.3\columnwidth]{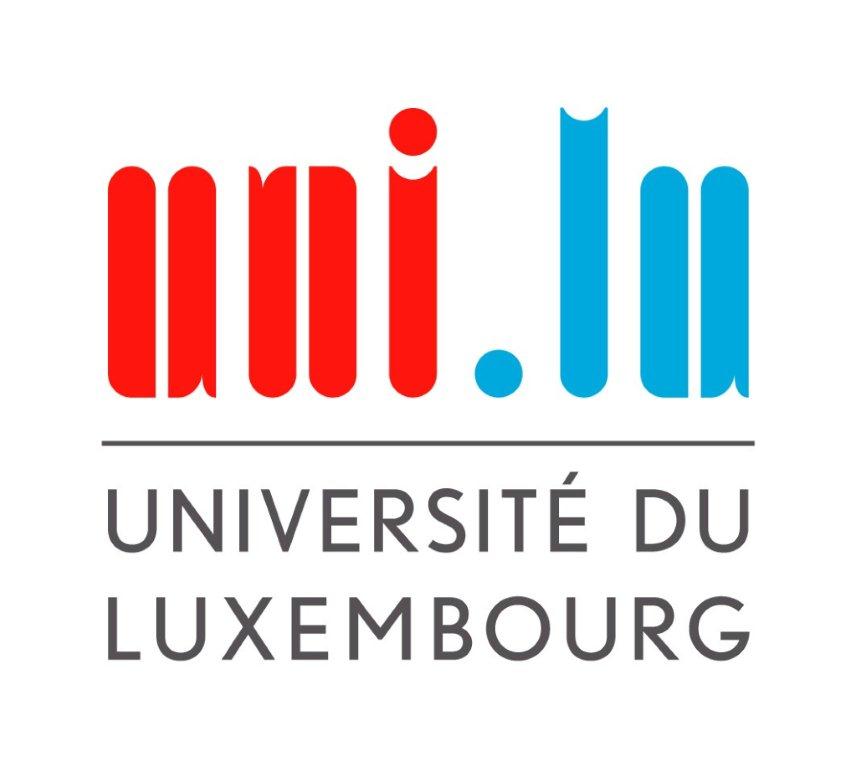}
\centering
\end{figure}

\begin{figure}
\includegraphics[width=0.3\columnwidth]{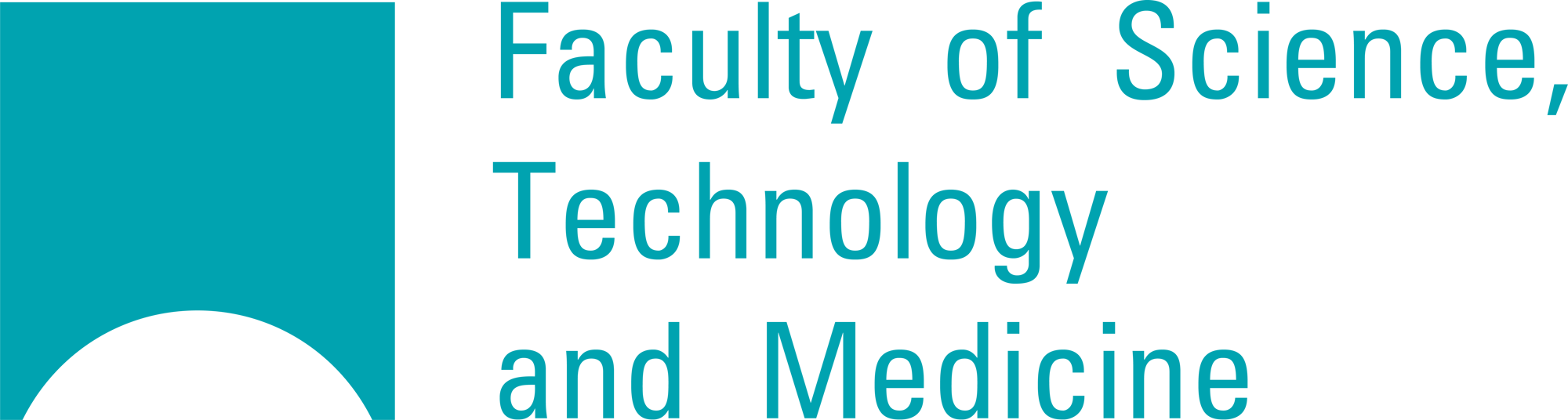}
\centering
\end{figure}
\vfill
\begin{center}
Doctoral School in Science and Engineering\\
\end{center}

\vfill
\noindent \textbf{Dissertation Defence Committee:}
\vspace{0.2cm}
\begin{itemize}[leftmargin=0cm]
\item[] Committee members: \hspace{2cm} Prof. Decebal Mocanu
	\item[] \hspace{5.8cm} Prof. Marija Slavkovik
	\item[] \hspace{5.8cm} Prof. Laetitia Matignon
	\item[] \hspace{5.8cm} Dr. Amro Najjar
	\item[] \hspace{5.8cm} Prof. Leendert WN van der Torre
\end{itemize}
\begin{itemize}[leftmargin=0cm]
	\item[] Supervisor: \hspace{3.6cm} Leendert WN van der Torre, Professor
\end{itemize}

\newpage

\noindent \textbf{\large{Affidavit}}\\
\newline
I hereby confirm that the PhD thesis entitled “What if Pinocchio Were a Reinforcement Learning Agent: A Normative End-to-End Pipeline” has been written independently and without any other sources than cited.
\vspace{2.5cm}
\newline
\noindent Luxembourg, \rule{4cm}{0.4pt}  \hspace{2cm} \rule{6cm}{0.4pt}
\newline 
\hspace*{9cm} Name

\newpage

\topskip0pt
\vspace*{\fill}

\begin{center}
    \textbf{Remerciements}
\end{center}

\textit{Je tiens à remercier mon superviseur, le Professeur Leon van der Torre, pour m'avoir permis de réaliser cette thèse, ainsi que pour m'avoir fait confiance dans le choix de ma thématique. J'ai apprécié cette collaboration et je suis aujourd'hui fier de pouvoir lui présenter ce manuscrit. Je remercie aussi Dr Amro Najjar pour son aide et ses conseils tout au long de ma thèse, ainsi que pour son soutien dans mon travail.}

\textit{À mes amis, Salma et Aria, je vous remercie sincèrement pour m'avoir accompagné dans cette aventure et de m'avoir soutenu tout du long. Merci à vous pour votre bienveillance. J'ai apprécié passer ces moments à vos côtés et j'espère poursuivre cette amitié bien au-delà de ces quatre années.}

\textit{Je remercie mes collègues et co-auteurs, Adam, David, Emery, Alexandre et tous les autres que je ne peux citer, pour leurs conseils. Votre aide m'a été précieuse et je garde un souvenir indélébile de nos collaborations.}

\textit{Christopher, je te remercie de m'avoir fait découvrir le monde de la recherche ainsi que de m'y avoir donné goût. Tout a débuté avec nos travaux sur AJAR, et quatre ans plus tard, cette thèse en est l'aboutissement. J'espère que tu la trouveras à la hauteur de notre collaboration.}

\textit{Enfin, je te remercie Gloria. À mes côtés du début à la fin, je te suis profondément reconnaissant. Merci d'avoir partagé cet instant avec moi. Ce ne fut pas un long fleuve tranquille, mais comme le disait si bien Paul Valéry, ``Le vent se lève !... Il faut tenter de vivre !'', Et ainsi dans les moments de doute, tu as su m'épauler, me soutenir, et m'écouter avec passion. Si nul ne peut prédire l'avenir, je peux dire avec certitude de par ces dernières années qu'à jamais dans mon coeur tu resteras.}

\vspace*{\fill}

\newpage

\topskip0pt
\vspace*{\fill}

\textit{Je dédis cette thèse à mes parents, Joséphine et Éric, sans qui rien de tout cela n'aurait été possible. Je vous suis reconnaissant de toujours m'avoir poussé et soutenu, non seulement ces quatre années, mais aussi les vingt-quatre qui ont précédé. Malgré la distance, vous avez su être présents, et même si le titre de cette thèse peut vous sembler quelque peu obscur, vous avez toujours su écouter avec attention et fierté lorsque j'en parlais. Je suis aujourd'hui fier, à mon tour, de pouvoir écrire ces mots dans ce manuscrit, rendant ainsi à César ce qui de droit lui revient. Mes parents : je vous aime.}

\vspace*{\fill}

\newpage
\renewcommand*\contentsname{Index}
\tableofcontents
\newpage
\listoffigures
\listoftables
\newpage

\topskip0pt
\vspace*{\fill}

\begin{center}
    \textbf{Abstract}
\end{center}

In the past decade, artificial intelligence (AI) has developed quickly. With this rapid progression came the need for systems capable of complying with the rules and norms of our society so that they can be successfully and safely integrated into our daily lives. Inspired by the story of Pinocchio in ``Le avventure di Pinocchio - Storia di un burattino'', this thesis proposes a pipeline that addresses the problem of developing norm compliant and context-aware agents. Building on the AJAR, Jiminy, and NGRL architectures, the work introduces \pino, a hybrid model in which reinforcement learning agents are supervised by argumentation-based normative advisors. In order to make this pipeline operational, this thesis also presents a novel algorithm for automatically extracting the arguments and relationships that underlie the advisors' decisions. Finally, this thesis investigates the phenomenon of \textit{norm avoidance}, providing a definition and a mitigation strategy within the context of reinforcement learning agents. Each component of the pipeline is empirically evaluated. The thesis concludes with a discussion of related work, current limitations, and directions for future research.

\vspace*{\fill}

\newpage

\pagenumbering{arabic} 
\setcounter{page}{1}

\chapter{Introduction}\label{cha:intro}

\subfile{sections/1_introduction}

\chapter{Preliminaries}\label{cha:background}

\subfile{sections/1.1_preliminaries}

\chapter{\pino: A Context-Aware Normative Architecture}
\label{cha:pino}

\subfile{sections/2_normative_architecture}

\chapter{Dynamically Modelling the Norms of Stakeholders}
\label{cha:aria}

\subfile{sections/3_dynamically_modelling_norms}

\chapter{Proposing a Robust Approach to Norm Avoidance}
\label{cha:normavoidance}

\subfile{sections/4_norm_avoidance}

\chapter{Related Work}

\subfile{sections/5_related_work}

\chapter{Future Work}\label{cha:future}

\subfile{sections/6_future_work}

\chapter{Summary}

\subfile{sections/7_summary}

\chapter*{References}
\addcontentsline{toc}{chapter}{References}
\begin{sloppypar}
\printbibliography[heading=none]
\end{sloppypar}


\end{document}

%% file: sections/1_introduction.tex
\section{Context and Motivations}

\lettrine[lines=3, lraise=0.25, findent=3pt]{P}{inocchio} is the main protagonist of ``Le avventure di Pinocchio - Storia di un burattino''. This book and its adaptations tell the story of a sentient puppet that faces morally relevant situations along its wanderings. His companion, Jiminy Cricket, plays the role of its consciousness, advising the puppet on what is right and wrong. Eventually, the story ends well as Pinocchio learns from his experiences and he is then transformed into a real, flesh-and-blood little boy.
Although Pinocchio is often seen as an allegory for naughty children, we can also interpret it as an artificial agent~\cite{russell1995modern,wooldridge2009introduction,botti2025agentic}. 
An artificial agent is an autonomous system capable of perceiving its environment, reacting to changes, pursuing its own goals proactively, and interacting socially with others in the pursuit of individual or collective tasks.
We can regard Pinocchio as an agent that will do anything that gives a reward, this without accounting for any ethical or normative considerations. For this reason, it requires this external consciousness that is Jiminy Cricket and that provides some normative guidance to it.

In the past, Artificial Intelligence (AI)\footnote{\citet{nilsson2009quest} defined AI as follows: ``\textit{Artificial intelligence is that activity devoted to making machines intelligent, and intelligence is that quality that enables an entity to function appropriately and with foresight in its environment}''.} had a limited field of applications due to technical constraints when dealing with the real world. This resulted in very few scenarios in which AI was interacting with humans. Today, AI is developing rapidly, and it will be common in the future to engage with AI systems in our daily life. Already, several of these AI systems can be found in various real-world applications, such as autonomously driving cars on our roads~\cite{shreyas2020self}, automated job applicant selection and shortlisting~\cite{pal2022resume}, smart grids regulating energy distribution~\cite{krontiris2021smart}, audiovisual and textual content generation for entertainment or customer support~\cite{george2023review}, adaptive robots learning while operating alongside warehouse workers~\cite{aliotam2020smart}, or conversational agents~\cite{brown2020language}.
Because these artificial agents are now evolving in our world, they need to follow our norms~\cite{tomic2019robby}.

In human societies, norms serve as implicit or explicit guidelines that govern interactions to ensure coordination, safety, and efficiency~\cite{morris2015normology,pham2024multi}. These norms can be legal (\emph{e.g.}, traffic laws for autonomous vehicles), social (\emph{e.g.}, queuing in public spaces) or ethical (\emph{e.g.}, fairness in AI decision-making). These norms can sometimes be made explicit through text or they can emerge dynamically through repeated interactions. Norms may vary across cultural and situational contexts~\cite{mc2011linking}.

A system or an environment in which norms are in force can be called a \emph{normative system}. Not respecting the relevant norms can lead to misunderstandings, lack of trust, or dangerous behaviours, ultimately resulting, at best, in inefficiency and, at worst, in moral or physical harm. However, several questions arise. How can agents behave optimally if they have to adhere to norms that restrain their actions? Where would the norms in question come from? How should we shape agents so that they integrate into these systems? The aforementioned are just a few of many questions and challenges we are currently facing.

In the scientific literature, one can find two major kinds of approaches. The first kind, often referred to as \emph{top-down} approaches, usually relies on the concept of norms as defined in deontic logic. Deontic logic is a subfield of logic that classifies norms as obligations, permissions, or prohibitions~\cite{gabbay2021handbook}. Doing so allows us to use logical tools to, for example, assess whether an agent is behaving well according to a set of norms. It can also serve to determine what an agent ought to do even if that is not stated explicitly. It can also be extended by considering the cases where a norm does not have to be respected as the situation is exceptional, in the sense that it fits a set of criteria that makes adherence to the norm less relevant.
To some extent, the Jiminy architecture~\cite{liao2019building,liao2023jiminy}\footnote{This architecture is more deeply detailed in Section~\ref{sec:bg_jiminy}.} can be seen as employing a top-down approach. The architecture serves to determine, in a given context, which actions an agent has the right to perform. To do this, it considers the preferences and arguments of several stakeholders, such as the user, the law enforcement, or the manufacturer. It also resolves dilemmas where no action is fully compliant with the norms.
This ensemble of techniques enables reasoning about the duties of an agent, but it often struggles in uncertain, large, and continuous environments where the consequence of an action is not clear and where there is no way to avoid breaking the rules.

The second group of approaches, known as \emph{bottom-up} approaches, involves analysing large quantities of unstructured data, or learning through trial and error, so that the agent's behaviour is optimal in terms of the task it was designed for. But when agents engage in a prohibited action, they can be penalised. This incentivises agents to complete their given task in a way that complies with the system's norms. Although these techniques work well in complex environments, they do not provide any guarantee of what exactly is learnt by the system. Is the agent learning malicious behaviours? Does it favour certain norms over others in cases of conflict? Can it deliberately violate a norm? Because of the opacity of such systems, it is not realistic to imagine we can obtain solid answers. Since regulations on autonomous systems increasingly emphasise transparency (see the European AI Act~\cite{goodman2017european}), this problem can no longer be tolerated.

Consequently, a third class of approaches exists. It aims to address the problems associated with both top-down and bottom-up approaches while preserving their good points---top-down approaches are strong on transparency and deductive power while bottom-up approaches are strong on adaption capability and computational efficiency. The third class of approaches are often referred to as \emph{hybrid}.
One example of such a hybrid approach is the NGRL (Norm-Guided Reinforcement Learning) agent\footnote{This approach is presented in more detail in Section~\ref{sec:bg_neufeld}.} proposed by \citet{neufeld2022reinforcement} (and further developed in subsequent publications~\cite{neufeld2023norm,neufeld2024learning}). In this reinforcement learning~\cite{sutton1999reinforcement} architecture, an agent learns to achieve its main task by learning a policy that optimises its reward while prioritising compliance with the norms in its environment. It evolves in accordance with a labelled Markov decision process that returns a set of labels for each state visited by the agent. Using constitutive norms~\cite{searle1969derive}, these labels serve to determine whether the agent committed a violation. Then, such information is returned to the agent so that it can then update its policy.
Another example is AJAR (A Judging Agents-based RL-framework)\footnote{See Section~\ref{sec:bg_ajar} for more details.} proposed by \citet{alcaraz2023ajar,alcaraz2026combining}. Whereas the previous approach emphasises the learning algorithm, this approach focuses on the governor component that determines whether the agent acted ethically\footnote{AJAR was designed for applications in ethics rather than normative systems specifically.} or reprehensibly, evaluated against a designated moral value. This judgement is then scalarised into a numerical value and combined with the reward for accomplishing the main task. The final value is then given to the reinforcement learning agent so it can learn to optimise it.

In the story of Pinocchio, both Pinocchio and Jiminy can be modelled using the agent metaphor~\cite{boella12004agent,boella2002attributing}. This metaphor attributes mental attitudes to actors\footnote{Here, an actor is a general term that encompasses a single social entity, a group~\cite{boella2004groups}, a virtual community~\cite{boella2003local}, a normative system~\cite{boella2003attributing,boella2005constitutive}, a contract~\cite{boella2004contracts} or an organisation~\cite{boella2002attributing,boella2004structuring}.} such as goals or beliefs. With respect to the previously introduced hybrid architectures, Pinocchio's goal is simply to optimise its own reward. It believes that performing a certain action in a particular context will produce a favourable outcome. In contrast, Jiminy Cricket's goal is to ensure that the perspectives of multiple stakeholders are taken into account. Its beliefs correspond to the arguments put forward by all of them. When Jiminy advises Pinocchio and Pinocchio follows the advice, we can see it working as a hybrid system where Pinocchio is the agent\footnote{Possibly a learning agent.}, and Jiminy is the moral or normative governor.

This thesis aims at creating a novel hybrid approach called \emph{\pino}~\cite{alcaraz2026pinocchio} (where $\pi$ refers to policy in reinforcement learning, and No refers to Normative). It is intended to complement rather than replace existing approaches. This thesis proposes a base architecture that makes use of recent top-down and bottom-up techniques. This architecture serves as an interface for training an agent to follow the norms of various stakeholders. These stakeholders may or may not share the same points of view.\footnote{As Perelman~\cite{perelman2012justice} puts it, ``If men (\textit{sic}) oppose each other concerning a decision to be taken, it is not because they commit some error of logic or calculation. They discuss apropos the applicable rule, the ends to be considered, the meaning to be given to values, the interpretation and characterisation of facts.'' These factors are why, even though the stakeholders discuss the same set of norms with the same observations, they may still disagree.} For the sake of transparency and explainability, the process of determining whether a norm must be respected in a certain context~\cite{aarts2003silence,malle2017networks} is carried out by a symbolic reasoning engine. Meanwhile, to ensure the agent's ability to adapt to unknown environments and a broad range of situations, we make use of reinforcement learning (RL), a common choice among bottom-up approaches.

The proposed architecture requires the norms to be represented in a structured and explicit manner, as they require to be manipulated by the top-down part. But such representation of these norms is not always directly available and often need to be handcrafted by those who wish to use this approach. Finding the reasons that render a norm irrelevant and the reasons that reinstantiate it~\cite{aarts2003silence,malle2017networks} (\emph{i.e.}, a reason that creates an exception to the exception) may prove to be even more complex. This limited scalability can drastically limit the applicability of the approach if not handled properly. For this reason, we propose to enhance this approach with an algorithm that aims to extract the reasons which may render a norm inapplicable, or reinstantiate it for a given context.

Furthermore, when developing this system, we identified an emerging issue raising serious concerns, not only for the proposed approach but also the ones developed within the literature of normative agents. This issue, termed \emph{norm avoidance}, is a phenomenon that shares similarities with reward hacking, as it can mislead the agent into gaming the norms.\footnote{One counter-argument is that one should just create the norms more cautiously so that norm avoidance does not happen. We claim that this is not a plausible solution as conceptual flaws have passed undetected many times in the history of computer science, \emph{e.g.}, phantom attacks~\cite{nassi2020phantom}, integer overflow~\cite{le1996ariane}, or biased data~\cite{yucer2024racial}.} Then, the behaviour learnt by the normative agent may not correspond to that expected by the designer, potentially leading to behavioural hazards or safety issues. As such, we formalise this concept and propose, as a first step, a way to address it with \pino.

Since this thesis draws on techniques from various subfields of AI, the following section provides a summary of the essential background required for proper understanding.

\section{Research Questions}
\label{sec:rqs}

Research questions were formulated to guide the development of the thesis and clarify its core contributions. They are as follows:

\begin{enumerate}[start=1,label=\textbf{RQ\arabic*.}, itemsep=-5pt, leftmargin=*]\tightlist
    \item How can we design an artificial agent that adheres to context-specific norms?
        \begin{enumerate}[start=1,label=\textbf{RQ1.\arabic*.}, itemsep=-5pt, leftmargin=*]\tightlist
            \item How can we design an agent that learns behaviours integrating heterogeneous normative viewpoints?
            
            \item Which methods can mitigate reward-gaming that compromises norm adherence by RL agents?
        \end{enumerate}
        
    \item What is the most suitable method for gathering norms to support the functioning of the proposed architecture?
        \begin{enumerate}[start=1,label=\textbf{RQ2.\arabic*.}, itemsep=-5pt, leftmargin=*]\tightlist
            \item How to learn the context-dependent exceptions to the application of a norm?
            
            \item In what way can norms and their exceptions be modelled to enhance intelligibility?
        \end{enumerate}
\end{enumerate}

In order to answer these research questions, this thesis proposes an architecture, \pino~\cite{alcaraz2026pinocchio}, that is based on a combination of the Jiminy architecture~\cite{liao2019building,liao2023jiminy}, the NGRL agent~\cite{neufeld2022reinforcement,neufeld2023norm,neufeld2024learning} and the AJAR framework~\cite{alcaraz2023ajar,alcaraz2026combining}. Even though these previous works have attempted to address the aforementioned research questions, by now they each suffer several limitations due to certain considerations that were out of scope at the time of their design.

First, the Jiminy architecture, while valuable for modelling heterogeneous stakeholder viewpoints, presents difficulties when integrated with reinforcement learning agents as it was originally agnostic to supervised agents. Its reasoning process is computationally costly, both online and offline, requiring extensive inference rules and the construction of argumentation graphs, which makes its use in stochastic environments challenging. Moreover, it provides no mechanism for handling uncertainty, as it exclusively selects actions prior to execution and does not adapt to address their consequences, which can lead to short-sighted behaviours. The approach also raises privacy concerns since stakeholder models stored within the system may expose sensitive information. The architecture assumes that all knowledge must be made explicit, which risks discouraging stakeholders from disclosing their argument, and eventually not accepting the use of such a system.

Second, the AJAR framework was originally designed for machine ethics rather than normative systems\footnote{While Jiminy was also designed for machine ethics, it is specifically focused on deontology. It can be argued that deontology, and normative systems in general, can be seen as a subfield of ethics, but with very specific requirements that make it different to the other branches of ethics. For example, there is a clear emphasis on the binary nature of compliance with norms, whereas ethical domains emphasise compromises.}, which limits its applicability in the particular context of normative systems. Its aggregation of the judges' outputs is largely arbitrary, and it assumes that each moral value is evaluated in isolation by a single judge, without considering how different values or norms might conflict with others. More importantly, AJAR treats ethical evaluations as scalar rewards that can be balanced against task performance. Although this is acceptable in ethical decision-making, it is problematic in normative settings where compliance with norms is treated as an obligation rather than a negotiable preference.

Finally, the NGRL agent, although providing a solid foundation for incorporating norms into RL, also suffers from notable limitations. It assumes that the set of constitutive and regulative norms and the thresholds for lexicographic selection are both predefined, which requires a significant amount of expert knowledge and severely limits scalability. The choice of thresholds itself is arbitrary, and the framework only offers a limited way of handling conflicts, leaving it vulnerable in situations where violations are unavoidable or where norms clash in complex ways. In particular, it is also prone to causing agents to commit norm avoidance. As a result, while NGRL represents an important step towards hybrid normative agents, it is less robust in dynamic, uncertain environments.

The aim of the architecture proposed in this thesis, \pino, is to make an agent learn to comply with context-dependent norms (\textbf{RQ1} and \textbf{RQ1.1}). Since this architecture requires that the data is collected (\textbf{RQ2}) and structured in a way that meets specific requirements and the intelligibility needs of normative systems (\textbf{RQ2.1} and \textbf{RQ2.2}), a method is also proposed to meet this requirement, the ARIA algorithm. Finally, since the phenomenon of norm avoidance can emerge within this framework (\textbf{RQ1.2}), a set of definitions is provided to characterise it along with strategies to limit its occurrence in a reinforcement learning setup.

\section{Methodology}
\label{sec:methodology}

Normative reasoning approaches, formal argumentation-based approaches, reasons-first-based approaches, and architectural approaches are different tools that serve different purposes. The field of logic has seen interest in combining these approaches as part of a larger toolbox~\cite{blackburn1997combine,yu2025bdi}. This thesis follows the same intuition to construct the \pino architecture, merging different methods from the literature,---reinforcement learning, moral supervisors, argumentation---to accommodate the particular requirements of normative systems.

The use of reinforcement learning~\cite{sutton1999reinforcement} allows for flexibility when faced with uncertain outcomes or situations that have not been encountered before. It also makes it possible to see beyond the direct consequences of an action while maintaining a low computational cost. As we face norms, we use the NGRL agent~\cite{neufeld2022reinforcement,neufeld2023norm,neufeld2024learning} to account for norms and their violations, but we improve the reasoning part by using an adjusted version of the Jiminy architecture~\cite{liao2019building,liao2023jiminy}. This allows the status of the norms to account for not only the ever-changing context but also the different viewpoints and arguments that the stakeholders hold. In order to guarantee a modular basis to convert the symbolic output of Jiminy into a reward that can be used to train the NGRL agent, the \pino architecture is based on the AJAR framework~\cite{alcaraz2023ajar,alcaraz2026combining}.

\section{Evaluation}
\label{sec:intro_eval}

This thesis is organised in such a way that each chapter is self-contained. Consequently, each approach presented in these chapters is evaluated independently, following their own testing protocol.

In Chapter~\ref{cha:pino}, the \pino architecture is evaluated in a custom reinforcement learning environment centred on an autonomous taxi agent. The evaluation focuses on the capacity of the agent to learn its task while adhering to relevant norms. Then, two variants of this agent are compared in order to assess the benefits of each.

In Chapter~\ref{cha:aria}, the evaluation focuses on the ability of the proposed algorithm to accurately extract an argumentation graph while still providing good predictive accuracy. First, a quantitative comparison of the accuracies of the proposed approach and its variants is conducted on benchmark tabular datasets from the classification literature. This is followed by a qualitative study of the explanatory potential of two of the variants.

In Chapter~\ref{cha:normavoidance}, approaches for mitigating norm avoidance are first defined and then evaluated by running them over a set of MDP (Markov decision process) environments representing various situations where norm avoidance can occur. The different approaches are compared in terms of performance against one another and against a normative agent architecture from the literature. The evaluation then proceeds to assess operational effectiveness and identify the advantages and disadvantages of each approach.

\section{Contributions}
\label{sec:contributions}

This thesis contributes to the field of normative systems by proposing an end-to-end pipeline that combines formal argumentation and normative reasoning with model-free normative reinforcement learning. This pipeline can be broken down into two main components.

The first component, the \pino architecture, responds to \textbf{RQ1} and \textbf{RQ1.1}. It aims to teach an artificial agent to follow norms while performing its assigned task. It combines and builds a normative RL framework upon three works in the literature, namely Neufeld's normative agent~\cite{neufeld2022reinforcement,neufeld2023norm,neufeld2024learning} which is a solid ground for training an agent to follow norms within an unknown stochastic environment, the Jiminy architecture~\cite{liao2019building,liao2023jiminy} which proposes a model for a normative supervisor, and the AJAR framework~\cite{alcaraz2023ajar,alcaraz2026combining}, which we developed in a prior project, and constitutes the base framework, since it was originally designed to combine symbolic supervision with reinforcement learning. The challenge of developing such an architecture lies in the translation of the symbolic normative reasoning into a numeric value that can be treated by the reinforcement learning part. Jiminy was not designed such that it returns a numerical value, and AJAR was not made in a way that it allows dialogues and collective reasoning between the judges. For these reasons, these two architecture need to be adapted so that they overcome these limitations.

The second component, answering \textbf{RQ2.1} and \textbf{RQ2.2}, aims at identifying the norms, and more precisely the exceptions to these norms, within a normative system. It does that by learning over behavioural datasets. Then, it represents these norms in the form of an argumentation graph, allowing for the first component to use them. This is meant to lessen the burden on the designer, rendering the approach less prone to human mistakes and more scalable. This component requires the norm to be known beforehand. As we do not want to reinvent the wheel, we prefer to let the user select one of the many methods to norm identification. However, in order to guide this choice, we provide a review of the literature on these methods and identify their specificities, such that it answers \textbf{RQ2}.

Finally, some empirical observations allowed me to identify a problem transversal to the field of normative agents, namely norm avoidance. Consequently, the last part of this thesis attempts to define this novel challenge by presenting definitions and examples and subsequently proposes an ad hoc solution for the specific case of normative reinforcement learning so that it answers \textbf{RQ1.2}. Here, the challenge lies in correctly framing the cases that can be categorised as norm avoidance and ensuring that the proposed methods to mitigate it address it without blocking behaviours that would be considered as acceptable.

A transversal challenge to the making of this thesis is the interdisciplinary aspect, as it combines philosophy and engineering. In consequence, it has to comply with the philosophical ideas developed over the years about the ethical and deontological principles or how dialogues should be led, while still ensuring correct functioning with resource complexity and scalability in mind, or technicalities such as the collection and use of data, or making of user-friendly interfaces.

Together, the various elements of this thesis should form a robust and scalable end-to-end pipeline for normative reinforcement learning.

\section{Layout of this Thesis}
\label{sec:layout}

This thesis is organised as follows. First, the technical background for the correct understanding of the dissertation, as well as the necessary knowledge about the reused approaches, is provided in Chapter~\ref{cha:background}. Then, Chapter~\ref{cha:pino} details an architecture for normative reinforcement learning that combines and adapts several approaches from the literature.
Chapter~\ref{cha:aria} provides an extensive analysis of the literature on norm identification and then proposes a novel algorithm that can be combined with existing norm mining methods to extract norms, as well as their potential exceptions, from observations. 
Then, Chapter~\ref{cha:normavoidance} proposes a preliminary definition of norm avoidance. Then, it partially addresses this problem by proposing a solution for the approaches based on reinforcement learning.
Finally, the remaining sections of this thesis discuss the results obtained with a higher-level reflection and present some related work. The manuscript is concluded with a presentation of the challenges for future research, as well as a summary of the content.

Each chapter is self-contained, and Chapter~\ref{cha:background} provides all the necessary knowledge about the tools we used inside of this thesis.

%% file: sections/1.1_preliminaries.tex
This section of the thesis introduces in Section~\ref{sec:technical} the necessary technical preliminaries for the understanding of the following chapters. Then, in Section~\ref{sec:background} it presents the main three works on which it builds.

\section{Technical Background}
\label{sec:technical}

This section provides the necessary background for understanding the content of this dissertation. Section~\ref{sec:rl} details what reinforcement learning and Markov decision process are, while Section~\ref{sec:norms} explains the way norms are defined across the Deontic Logic literature. Finally, Section~\ref{sec:argumentation} introduces formal argumentation, and more specifically, abstract argumentation.

\subsection{Reinforcement Learning}
\label{sec:rl}

Reinforcement learning (RL) is used to train an agent to perform a certain task in an environment that may or may not be stochastic. The agent learns through trial and error. Each time it performs an action, it receives a reward, represented by a numerical value. It features several advantages over the other approaches, such as its capacity to deal with stochastic (\emph{i.e.}, non-deterministic) environments, the small amount of expert knowledge required to have it working, or the fact that it converges towards optimal solutions.

More formally, it takes place in an environment formalised as a Markov Decision Process (MDP), defined as follows:

\begin{definition}\label{def:mdp}
    A Markov Decision Process is a tuple $\langle S, A, P, R \rangle$ where:
    \begin{itemize}\tightlist
        \item $S$ is a set of states
        \item $A$ is a function $A : S \rightarrow 2^{Act}$ from states to a set of possible actions (with $Act$ being the set of all the actions available to the agent)
        \item $R : S \times Act \times S \rightarrow \mathbb{R}$ is a scalar reward function over states and actions. It can be simplified to $R : S \times Act \rightarrow \mathbb{R}$ in a deterministic environment
        \item $P : S \times Act \times S \rightarrow [0, 1]$ is a probability function giving the probability of transitioning from the state $s$ to the state $s'$ when doing action $a$.
    \end{itemize}
\end{definition}

\begin{remark}
    A state represents how the agent perceives the environment. It can be provided explicitly, for example, by specifying that the agent is in state $s_i$, or in a raw form. In the latter case, the state may consist of pixels from an image, atomic propositions, numerical feature vectors, or a combination of these elements.
\end{remark}

\begin{figure}
    \centering
    \includegraphics[width=0.65\linewidth]{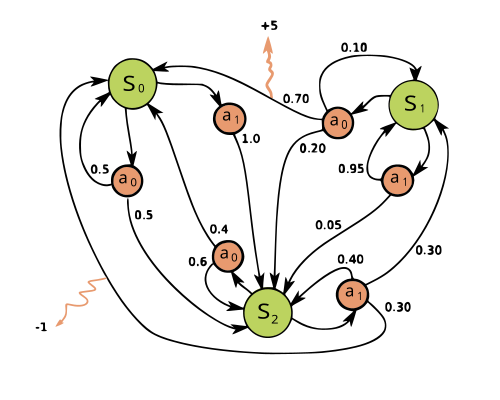}
    \caption{Example of an MDP.}
    \label{fig:mdp}
\end{figure}

To better understand what it corresponds to, a representation of an MDP containing three states ($S_{0-2}$) and two actions ($a_{0-1}$) is given in Fig.~\ref{fig:mdp}. We can see in this example that performing the action $a_1$ in state $S_1$ has a $5\%$ chance of moving the agent to $S_2$, and $95\%$ chances to make the agent stay in $S_1$. On the other hand, we see that if the transition $(S_1, a_0, S_0)$ occurs, the agent receives $+5$ as a reward.

The goal of reinforcement learning is to find a policy $\pi : S \rightarrow Act$ that designates the optimal behaviour with respect to the reward function, that is, behaviour that maximizes overall rewards in the long run, with emphasis on the most immediate rewards. This policy is denoted as $\pi^*$.

\begin{figure}[ht]
    \centering
    \includegraphics[width=0.7\linewidth]{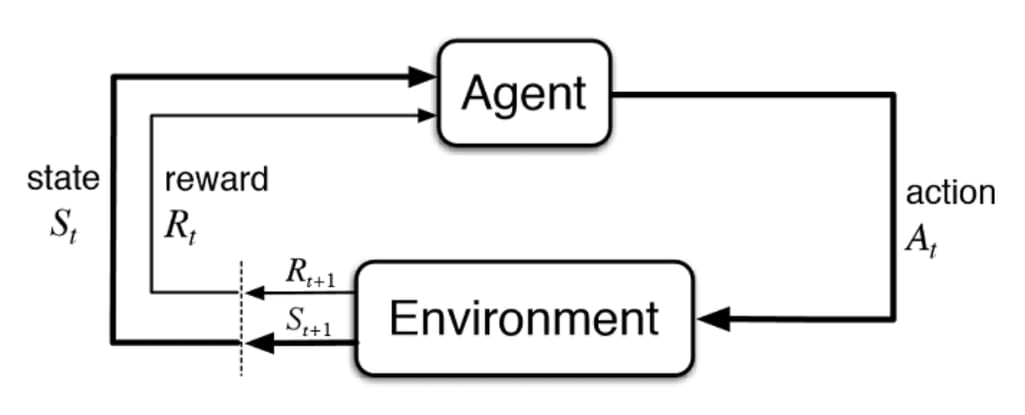}
    \caption{Reinforcement Learning training loop.}
    \label{fig:rl_training_bg}
\end{figure}

To learn $\pi^*$, we can use the $Q$-learning algorithm~\cite{watkins1992q} to learn the function $Q : S \times Act \rightarrow \mathbb{R}$. 
Specifically, the optimal \qf is defined as
\begin{equation}
    Q^{\pi^*}(s,a)=\mathbb{E}\left[\sum^{n}_{t=0}\gamma^{t} R(s_{t}, a_{t}) \Big{|} s_0=s, a_0=a\right]
    \label{eq:q}
\end{equation}
where $\gamma\in [0,1]$ is the discount factor, which determines how much emphasis is placed on the current reward versus future rewards, and $n$ is the number of steps in the agent's path. $Q^{\pi^*}$, then, gives the expected (discounted) sum of rewards over the agent's runtime, assuming that it takes action $a$ in state $s$, and thereafter takes a path, \emph{i.e.}, a sequence of state-action pairs, following $\pi^*$. We learn $Q^{\pi^*}$ by exploring the environment, typically through random actions that allow the agent to gather diverse experiences and to identify the most profitable actions based on the reward obtained and the resulting state. This exploration process is illustrated in the RL training loop shown in Fig.~\ref{fig:rl_training_bg}.

In $Q$-learning, for each transition $(s, a, s')$ the update
$$Q(s, a) \coloneqq Q(s, a) + \alpha(R(s, a) + \gamma \underset{a' \in A(s')}{max}Q(s', a') - Q(s, a))$$ 
is performed. In this expression, $\alpha \in [0, 1]$ is the learning rate, which determines how much emphasis is put on the former \qv in the update.
In other words, in $Q$-learning, we update $Q$ repeatedly until it converges to $Q^{\pi^*}$.
Then $\pi^*$ is: $$\pi^*(s) \in \argmax_{a \in A(s)}Q^{\pi^*}(s, a)$$

In the next section, we will refer to labelled MDPs. They can be defined as an extension of the definition of an MDP. This definition is as follows:

\begin{definition}\label{def:lmdp}
    A labelled MDP is a tuple $\langle S, A, P, R, L\rangle$\footnote{It is possible to remove the element $R$ corresponding to the reward and simply keep the tuple $\langle S, A, P, L\rangle$.} where $S, A, P, \text{ and } R$ are defined in the same way than in Definition~\ref{def:mdp}, and where $L : S \rightarrow 2^{AP}$ corresponds to a labelling function from states to subsets of a set of atomic propositions $AP$.
\end{definition}

\begin{figure}
    \centering
    \includegraphics[width=0.65\linewidth]{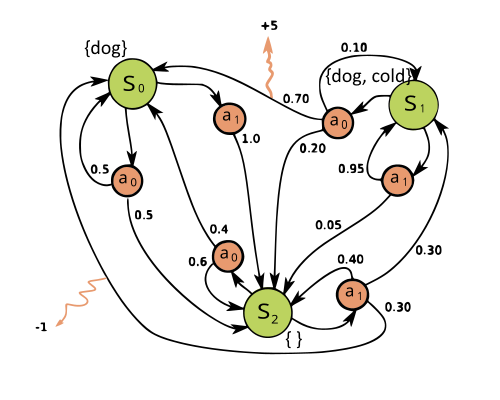}
    \caption{Example of a Labelled MDP.}
    \label{fig:lmdp}
\end{figure}

An example of labelled MDP can be seen in Fig.~\ref{fig:lmdp}. The state $S_0$ has a single label ``\texttt{dog}'', state $S_1$ has two labels ``\texttt{dog}'' and ``\texttt{cold}'', and state $S_2$ does not have any label.

While these labels are assumed to be mapped already to the set of states, they can in practice be computed at the time the agent enters the state, based either on the observations of the agent, or the properties of the state itself. This is particularly convenient when dealing with a non-finite of continuous state space.

\subsection{Norms}
\label{sec:norms}

In our society, norms serve as implicit or explicit guidelines that govern interactions, ensuring coordination, safety, and efficiency. These norms can be legal (\emph{e.g.}, traffic laws for autonomous vehicles), social (\emph{e.g.}, queuing behaviour in public spaces), or ethical (\emph{e.g.}, fairness in AI decision-making). These norms can sometimes be made explicit through text, but can also emerge dynamically through repeated interactions and may vary across cultural and situational contexts~\cite{mc2011linking}. They can also change and evolve over time~\cite{boella2009normative}.

Deontic logic is an area of logic that investigates normative concepts, \emph{i.e.}, deontic concepts~\cite{gabbay2021handbook}. It aims at developing tools and methods to represent these norms and reason about them. The deontic logic literature defines two types of norm: constitutive norms and regulative norms.

The former can be summarised as a way of describing how rules or norms bring certain social or institutional realities into existence by defining what actions, statuses, or roles count as within a given context. Searle characterises them as a way to build \emph{institutional} facts from \emph{brute} facts~\cite{searle1969derive}, but more simply, they can be seen as a \emph{counts-as} mechanism. For example, ``$X$ took an item in the store and left without paying'' counts as ``$X$ stole an item from the store''. Here, the brute fact is that $X$ left the store with an unpaid item, while the institutional fact is that $X$ stole this item. In short, they serve as incremental building blocks for determining what is and what is not from low-level observation. Note that the concept of constitutive norms has been further developed so that they can take into account a given context~\cite{searle1995construction}. For example, ``$X$ took an item in the store and left without paying'' counts as ``$X$ stole an item from the store'', under the condition that ``$X$ is not the manager of the store''. Such structure for a constitutional norm $C_i$ can be represented as $C_i(A, B\mid Y)$, which means that in context $Y$, $A$ counts as $B$ (for $A$ and $B$ being some events or properties). In this thesis, a constitutive norm having for context the tautology $\top$ will be rewritten $C_i(A, B)$.

On the other hand, regulative norms are closer to what we usually call ``norms'' in our society. They have a deontic content and indicate what is obligatory, permitted, and forbidden. In deontic logic, they are represented via the modalities \textbf{O} for the obligations, \textbf{P} for the permissions, and \textbf{F} for the prohibitions. They are often expressed with the following equivalences:

\vspace{0.2cm}
\hspace{1.1cm}
\begin{minipage}{0.45\textwidth}\centering
    \begin{enumerate}\tightlist
        \item[(i)] $\textbf{P}p \leftrightarrow \neg\textbf{O}\neg p$
        \item[(ii)] $\textbf{O}p \leftrightarrow \neg\textbf{P}\neg p$
    \end{enumerate}
\end{minipage}
\hfill
\begin{minipage}{0.45\textwidth}
    \begin{enumerate}\tightlist
        \item[(iii)] $\textbf{O}p \leftrightarrow \textbf{F}\neg p$
        \item[(iv)] $\textbf{F}p \leftrightarrow \textbf{O}\neg p$
    \end{enumerate}
\end{minipage}\\

For example, \textbf{O}$p$ signifies that one ought to do $p$, while \textbf{F}$p$ signifies that it is forbidden to do $p$. These notations can be extended to include conditionals. For $\textbf{X} \in \{\textbf{O, P, F}\}$, $\textbf{X}(p\mid q)$ indicates that one should comply with the norm \textbf{X}$p$ if it is the case that $q$. If a norm is simply noted as \textbf{X}$p$, we can assume $\textbf{X}(p\mid\top)$ where $\top$ is the logical tautology.

In this thesis, we make use of defeasible reasoning and, more specifically, defeasible norms. In the literature, normative conflicts may be exemplified by situations where we have \textbf{O}$p$ and \Pe{$\neg p\mid q$}, which means that if $q$ can be inferred from the knowledge base \kb, then we have an obligation that $\kb \models p$ and the permission that $\kb \not\models p$ at the same time. It can be interpreted in two different ways. If we follow the Prima Facie paradigm, the ideal world resolution is that $p$ holds, so there is no violation, while the suboptimal world resolution is that $p$ does not hold, so there is one violation. On the other hand, \citet{sergot1994contrary} see a world where $q$ is inferable as an exception to the norm \textbf{O}$p$. As such, not doing $p$ will not violate the obligation \textbf{O}$p$.
We can represent this type of situation with logic programming notation. Consequently, we can write $(\texttt{not } q) \rightarrow \textbf{O}p$, where ``$\texttt{not }q$'' means that $\kb \not\models q$, and the whole formula means that one does not ought to fulfil \textbf{O}$p$ if $\kb \not\models q$.

In order to make this dissertation clearer, we introduce Definitions~\ref{def:defeated}--\ref{def:violation}. Note that the terms used in these definitions may differ from the usual meaning given in the literature.

\begin{definition}[Defeated Norm]\label{def:defeated}
    A norm \textbf{O}$p$ is defeated when an exception to it is created by another norm \textbf{P}($\neg p\mid q$) and $q$ is inferable from the knowledge base. The norm can be rewritten as $(\texttt{not } q) \rightarrow \textbf{O}p$. For a prohibition \textbf{F}$p$, a defeat would be equivalent to having $\textbf{P}(p\mid q)$, similarly rewritten as $(\texttt{not } q) \rightarrow \textbf{F}p$.
\end{definition}

\begin{definition}[Activated Norm]\label{def:activated}
    A (potentially defeasible) norm $r \rightarrow\textbf{O}(p\mid q)$ is said to be activated when $q$ is inferable from the knowledge base.
\end{definition}

\begin{definition}[Compliance]\label{def:compliance}
    A norm \textbf{O}$p$ is not complied with if we have $q \rightarrow \textbf{O}p$ and $\kb \not\models p$, regardless of whether $\kb \models q$ or not.\footnote{Note that this differs from the more standard definition of non-compliance as a violation.} Similarly, a norm \textbf{F}$p$ is not complied with if we have $q \rightarrow \textbf{F}p$ and $\kb \models p$ regardless of whether $\kb \models q$ or not.
\end{definition}

\begin{definition}[Violation]\label{def:violation}
    A norm \textbf{O}$p$ is violated when $q \rightarrow \textbf{O}p$, $\kb \models q$, and $\kb \not\models p$. This is the same as non-compliance with a non-defeated norm. Alternatively, a norm \textbf{F}$p$ is violated if $\kb \models p$.
\end{definition}

\begin{remark}
    In this dissertation, we will use the term ``norm status'', or ``status of a norm''. The status can take four different values. Let a norm $r \rightarrow \textbf{O}(p\mid q)$ ``Non-activated'' (or ``deactivated''), correspond to when the condition of the norm cannot be inferred from the knowledge base (\emph{e.g.}, $\kb \not\models q$). It takes priority over the other status. Furthermore, a norm can be said to be ``activated'', which corresponds to the case where it is not ``deactivated''. However, this does not provide any additional information about whether or not the norm is defeated. ``Non-defeated'' is when the norm sees its condition fulfilled (\emph{e.g.}, $\kb \models q,r$). Then ``defeated'' refers to when a norm is activated but its extra condition is not (\emph{e.g.}, $\kb \models q$ but $\kb \not\models r$).
\end{remark}

\subsection{Formal Argumentation}
\label{sec:argumentation}

Formal argumentation consists of a compilation of techniques to model dialogues and enable reasoning through them. An interested reader may want to start with the first volume of \emph{Handbook of Formal Argumentation}~\cite{bochman2018argumentation}.

These techniques revolve around the use of logical formalism. Formal argumentation can be divided into three main branches~\cite{yu2023distributed}, namely, argumentation as dialogue~\cite{mcburney2004syntax,mcburney2004locutions,arisaka2022multi}, argumentation as balancing~\cite{gordon2007carneades,alcaraz2024estimating}, and argumentation as inference~\cite{dung1995acceptability,modgil2014aspic+,toni2014tutorial}. This section will focus on the introduction of the latter, as this is the one used through the approaches that serves as a basis to this work. More specifically, we will focus on abstract argumentation as introduced by Dung in his seminal paper~\cite{dung1995acceptability}.

Abstract argumentation consists of a framework for modelling arguments and conflicts. Rather than focusing on the internal structure of arguments, it treats them as abstract entities and defines a binary ``attack'' relation between them. The central idea is to reason about which sets of arguments can coherently stand together. Such a set is called an \emph{extension}. The intuition behind this concept, as Dung puts it, is that the way humans argue is based on a very simple principle which is summarised succinctly by an old saying: ``The one who has the last word laughs best''.

More formally, an abstract argumentation framework consists of the following.

\begin{definition}[Argumentation Framework]
    An argumentation framework is a tuple $F = \langle\args, \defs\rangle$, where $\args$ is a set of elements called arguments and $\defs \subseteq \args \times \args$ is a relation over the arguments referred to as attack.
\end{definition}

\begin{figure}
    \centering
    \includegraphics[width=0.5\linewidth]{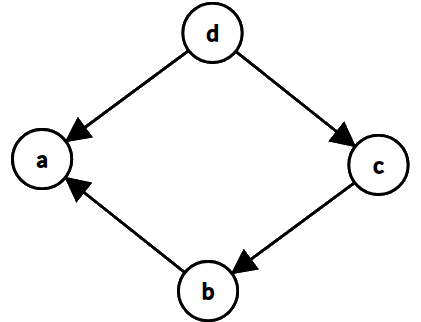}
    \caption{Representation as a directed graph of an argumentation framework.}
    \label{fig:af_example}
\end{figure}

It can be represented as a directed graph. For example, Fig.~\ref{fig:af_example} shows an argumentation graph corresponding to the argumentation framework $F = \langle\args, \defs\rangle$ where $\args = \{a, b, c, d\}$ and $\defs = \{(b, a), (c, b), (d, a), (d, c)\}$.

Given $\SA \subseteq \args$, we recall the notions of conflict-freeness and acceptability. A conflict-free set is one in which no argument in the set attacks another. Acceptability represents the constraint that an argument is only acceptable if all its attackers are themselves attacked by an argument in the set.    
\begin{itemize}\tightlist
    \item $\SA$ is a \emph{conflict-free} set of arguments \emph{w.r.t.} $\defs$ if and only if $\nexists a,b\in\SA$ \emph{s.t.} $a \defs b$, 
    \item For all $a \in \A$, $a$ is acceptable \emph{w.r.t.} $\SA$  if and only if $\forall b \in \A$, if $b \defs a$,  then $\exists c \in \SA$, such that $c \defs b$.
\end{itemize}

We recall the standard definitions for extensions:

\begin{itemize}\tightlist
    \item $\SA$ is an \emph{admissible extension} if and only if $\SA$ is conflict-free and all arguments $a \in \SA$ are acceptable \emph{w.r.t.} $\SA$. 
    \item $\SA$ is a \emph{complete extension} if and only if $\SA$ is admissible and contains all acceptable arguments \emph{w.r.t.} $\SA$. 
    \item $\SA$ is a \emph{grounded extension} if and only if $\SA$ is a minimal complete extension with respect to strict set inclusion \emph{i.e.} $\nexists \SA' \subseteq \args$ \emph{s.t.} $\SA' \subset \SA$, and $\SA'$ is a complete extension.
\end{itemize}

According to these definitions, the grounded extension that can be computed from the graph shown in Fig.~\ref{fig:af_example} is $Grd(F) = \{b, d\}$. It is fairly easy to compute it by hand, as we can quickly see that $d$ is not attacked and therefore has to be in the grounded extension. Then, $b$ is attacked only by $c$, and $c$ is defeated by $d$ (which we previously included in the extension). Consequently, $d$ is defending $b$, allowing the latter to be in the extension as it is not attacked by any argument which is also in the extension. Such an argument, made acceptable because of the defence from another argument, is said ``reinstantiated''. For larger graphs, an algorithm has been proposed by \citet{nofal2014algorithms} to compute the grounded extension with polynomial-time complexity.

\begin{remark}
    Although the use of grounded extension is not incompatible with graphs having symmetric attacks, reflexive attacks, or cycles, such properties may lead to undesired behaviours. Consequently, when dealing with graphs that may present one or more of these properties, extensions using for instance the preferred semantics should be privileged.
\end{remark}

Abstract argumentation has been extended in various ways. Some works integrated the handling of preferences or values over the arguments~\cite{bench2002value,coste2012weighted,amgoud2017acceptability,bistarelli2021weighted}. Some other works added another relation denoting a support from one argument to another~\cite{cayrol2005acceptability,nouioua2010bipolar,yu2023principle}. While each of these variants has its own set of advantages, we believe the approach proposed by Dung shines by its simplicity—similarly to propositional logic among other logics—making it particularly well-suited for integration into practical systems.

\section{Background}
\label{sec:background}

This thesis builds upon three key contributions from the literature. Section~\ref{sec:bg_ajar} discusses AJAR~\cite{alcaraz2023ajar,alcaraz2026combining}, a framework that we previously developed and which integrates symbolic reasoning with reinforcement learning. Then Section~\ref{sec:bg_jiminy} presents Jiminy~\cite{liao2023jiminy}, a normative reasoning architecture designed to handle deontic concepts. Finally, Section~\ref{sec:bg_neufeld} introduces a reinforcement learning architecture in which an agent learns to obey norms~\cite{neufeld2022reinforcement,neufeld2023norm,neufeld2024learning}.
While this thesis adapts and integrates these three works into a unified approach, the following section provides the necessary background to clarify the foundations upon which this thesis builds.

\subsection{The AJAR Framework}
\label{sec:bg_ajar}

The AJAR (A Judging Agents-based RL-framework) framework~\cite{alcaraz2023ajar,alcaraz2026combining} consists of the supervision of a reinforcement learning agent during its learning phase by a set of judges.

This framework was initially proposed for the field of \emph{Machine Ethics}. As such, in addition to its main task, the agent has to satisfy (or maximise) an ensemble of moral values. Each moral value is handled by a single judging agent. A judging agent consists essentially of a directed graph where nodes are arguments, and edges are considered as attacks.\footnote{Arguments and attacks in the sense of Dung's formal argumentation, although we cannot talk about an argumentation graph at this point as it may have inconsistencies among its arguments.} Then, a mapping function indicates whether an argument is supporting (\emph{i.e.}, \emph{pros} argument) the moral value, going against (\emph{i.e.}, \emph{cons} argument) it , or is neutral and instead serves to defeat some other arguments. For example, ``\textit{Low gas emissions}'' supports the ``\textit{Environment sustainability}'' moral value, while ``\textit{High gas emissions}'' goes against it. Then, ``\textit{Not often activated}'' may be neutral toward the ``\textit{Environment sustainability}'' moral value but at the same time defeat the ``\textit{High gas emissions}'' argument.

During the learning phase, after each agent's action within the environment, a preprocessed version of the state is forwarded to the judges. This preprocessed state serves to determine which argument is activated. Then, each judge computes a subgraph containing only the activated arguments. Finally, it computes an extension. From the arguments within this extension, and depending on whether they are \emph{pros} or \emph{cons}, each judge calculates a numerical reward for its given moral value. Then, an aggregation function scalarises each of these rewards with the reward from the environment (corresponding to the actual task of the agent, without any ethical consideration). Then, this final aggregated reward is returned to the reinforcement learning agent so that it can update its \qvs.

More formally, inspired by the AFDM (Argumentation Framework for Decision-Making)\footnote{We essentially remove the features that are not relevant for our technical specificities while keeping the notion of \emph{pros} and \emph{cons} arguments.} proposed by \citet{amgoud2009decision}, we define an AFJD (Argumentation Framework for Judging a Decision) as:

\begin{definition}
    An Argumentation Framework for Judging a Decision (AFJD) is a tuple $AF = \langle\args,$ $\defs,$ $\Ff,$ $\Fc\rangle $ where:
    \begin{itemize}\tightlist
        \item $\args$ is a non-empty set of arguments
        \item $\defs$ is a binary relation called \emph{attack relation}
        \item $\Ff \in 2^{\args}$ is the set of arguments which indicates that the RL-agent's last decision is moral \emph{w.r.t.} the moral value considered by the judging agent
        \item $\Fc \in 2^{\args}$ is the set of arguments which indicates that the RL-agent's last decision is immoral \emph{w.r.t.} the moral value considered by the judging agent.
    \end{itemize}
    
    For the sake of clarity and to alleviate notations in the sequel, for a given $AF = \langle\args,$ $\defs,$ $\Ff,$ $\Fc\rangle $, we note $AF_{[\args]}=\args$, $AF_{[\defs]}=\defs$, $AF_{[\Ff]}=\Ff$, $AF_{[\Fc]}=\Fc$. The set of all possible sub-AFJD for $AF$, \emph{i.e.}, all AFJD which arguments are a subset of $AF_{[\args]}$, is denoted as:
    \begin{align*}
        \mathcal{P}(AF)  := & \{\langle\args',\defs', \Ff', \Fc'\rangle: \args' \subseteq AF_{[\args]},  &\\  & \defs' \subseteq \args'^{2} \cap AF_{[\defs]}, & \\ &   \Ff' \subseteq \args' \cap AF_{[\Ff]}, \Fc' \subseteq \args' \cap AF_{[\Fc]} \} 
    \end{align*}
\end{definition}

Then, we accordingly define the AJAR framework as follows\footnote{Originally, AJAR was defined over a Decentralised Partially Observable MDP (DecPOMDP) and was handling several learning agents. For the sake of clarity, we simplify the definition so that it uses a standard MDP and judges a single learning agent.}:

\begin{definition}
    A Judging Agents-based RL-framework (AJAR) is a tuple $\mathcal{F}$ such that: $$\mathcal{F} = \langle\JAgts, \{AF_j\}_{j \in \JAgts}, S, A, P, R, \{\epsilon_j\}_{j \in \JAgts}, \{\Jj\}_{j \in \JAgts}, \gagr \rangle \text{ where}$$
    
    \begin{itemize}\tightlist
        \item $\JAgts $ is a set of judging agents,
        \item $\forall j \in \JAgts,$ $AF_j$ is a AFJD, 
        \item $\langle S, A, P, R \rangle$ is a MDP (see Definition~\ref{def:mdp})
        \item  $\forall j \in \JAgts $, $\epsilon_j: \States \to \mathcal{P}(AF_j) $ is a function that from a current state $s_t \in S$, assigns the sub-AFJD that the judging agent $j$ will use
        \item $\forall j \in \JAgts $, $\Jj: \mathcal{P}(AF_j)  \to \mathbb{R}$ is the \emph{judgment function} that returns from an argumentation graph the associated reward of the judgment,
        \item $\gagr: \mathbb{R}^{\left|\JAgts\right|} \to \mathbb{R}$ is an aggregation function 
        \item For all states $s_t \in S$ and the reward obtained from the MDP for the given transition $R_t = R(s_{t-1}, a_{t-1}, s_t)$, the reward $\mathcal{R}$ returned by AJAR is \emph{s.t.}: $$\mathcal{R}(s_t) = \gagr\Big(\{R_t\} \cup \varprod\limits_{j \in \JAgts}\Jj \big(\epsilon_j(s_t)\big)\Big)$$
    \end{itemize}
\end{definition}

\begin{figure}[ht]
    \centering
    \begin{minipage}[b]{0.45\linewidth}
        \centering
        \includegraphics[width=\textwidth]{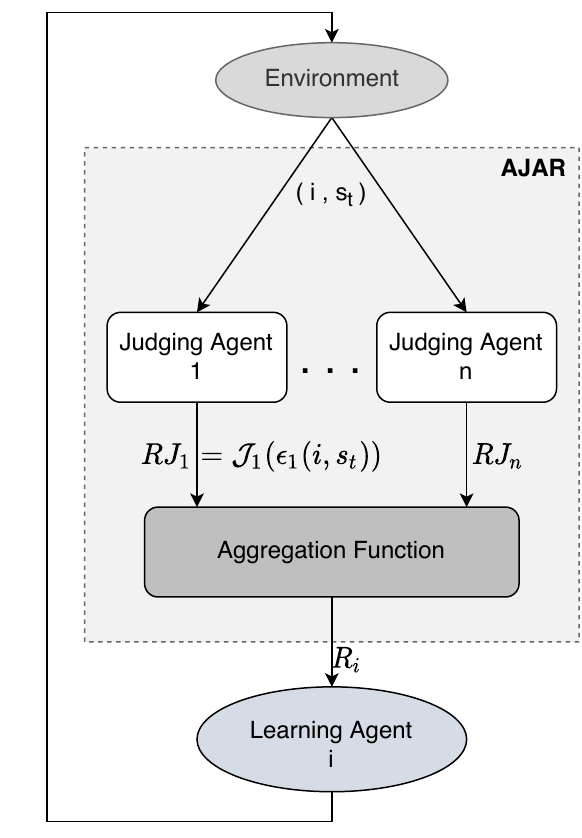}
        \caption*{(a) Overview of AJAR.}
        \label{subfig:ajar_overview}
    \end{minipage}
    \hfill
    \begin{minipage}[b]{0.45\linewidth}
        \centering
        \includegraphics[width=\textwidth]{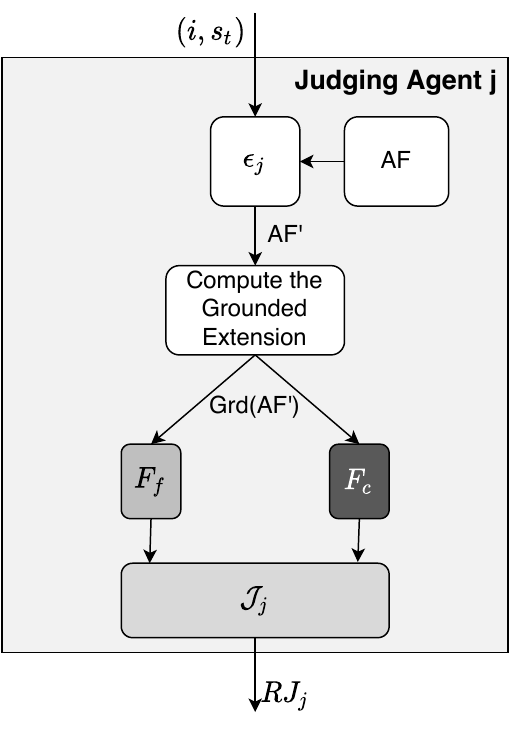}
        \caption*{(b) Detailed judgment.}
        \label{subfig:ajar_judgment}
    \end{minipage}
    \caption{Representation of the AJAR framework.}
    \label{fig:ajar}
\end{figure}

The judging process is illustrated, for a specific agent $i$, by Figs.~\ref{fig:ajar}a and~\ref{fig:ajar}b.

One way of defining the function $\Jj$ is (but is not limited to since the choice of the function ultimately falls to the designer) the following:

\begin{align*}
    pros &= \left| \Grd(\epsilon_j(s_t))\cap \epsilon_j(s_t)_{[\Ff]} \right| \quad\quad
    cons = \left| \Grd(\epsilon_j(s_t))\cap \epsilon_j(s_t)_{[\Fc]} \right|
\end{align*}

\begin{center}
\(
    \Jj\left(\epsilon_j(s_t)\right) =
    \begin{cases}
        \dfrac{pros}{pros + cons}, & \text{if } pros + cons \neq 0 \\
        \dfrac{1}{2}, & \text{otherwise}
    \end{cases}
\)
\end{center}

\subsection{The Jiminy Architecture}
\label{sec:bg_jiminy}

\begin{figure}[ht]
    \centering
    \includegraphics[width=0.8\linewidth]{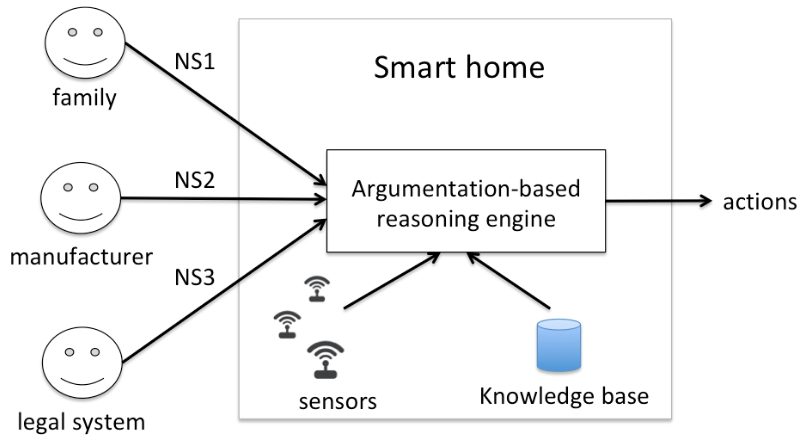}
    \caption{Jiminy's smart home example. Reused from~\citet{liao2019building}.}
    \label{fig:smarthome}
\end{figure}

The Jiminy architecture~\cite{liao2019building,liao2023jiminy}, whose name is inspired by Jiminy Cricket, is a moral advisor for an artificial agent. It aims at confronting the point of view of different stakeholders on a situation to determine, norm-wise, which action the agent has the right to perform. As this reasoning may require to be done in real-time for an artificial agent, stakeholders cannot simply express their views through a messaging application or any other communication process. An example of a situation where a smart home agent is supervised by three stakeholders, namely the family, the manufacturer, and the legal system, is shown in Fig.~\ref{fig:smarthome}.

\begin{figure}[ht]
    \centering
    \includegraphics[width=0.9\linewidth]{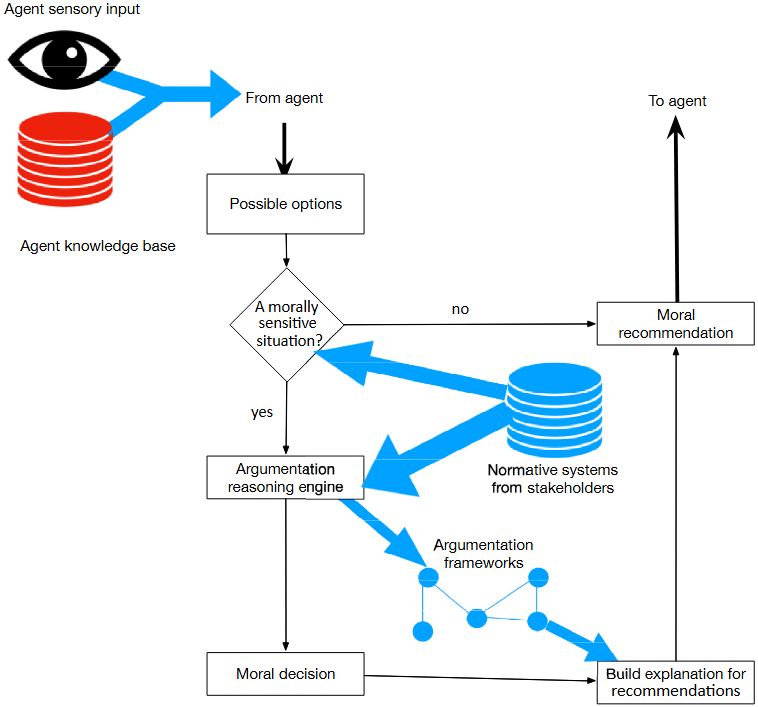}
    \caption{Representation of the Jiminy pipeline. Reused from~\citet{liao2023jiminy}.}
    \label{fig:jiminy}
\end{figure}

Jiminy, stakeholders' normative systems are modelled via a symbolic representation. This representation is called a stakeholder model, or stakeholder avatar. Through the use of structured argumentation~\cite{modgil2014aspic+}, these avatars are able to construct arguments from the symbolic observation of a state (the state here corresponding to the perceptions of the supervised agent). Then, arguments of the various avatars are organised based on their conflicts. Finally, using abstract argumentation, a set of actions that adhere to the result of the debate is extracted. The supervised agent can now select an action to do. This process is represented in Fig.~\ref{fig:jiminy}. Note that the strategy followed by Jiminy is the following:
\begin{enumerate}\tightlist
    \item Jiminy considers how the arguments of the stakeholders relate to one another, which may already resolve the dilemma.
    \item Jiminy combines the normative systems of the stakeholders such that the combined expertise of the stakeholders may resolve the dilemma.
    \item Only if these two other methods have failed, Jiminy uses context-sensitive rules to decide which of the stakeholders takes preference over the others.
\end{enumerate}

What interests us the most in this architecture is the separation of the stakeholders, which makes the normative system that the agent ought to follow more flexible and modular. Furthermore, its use of argumentation may simplify the process of explaining the decision of the agent to a human, as argumentation can be considered as a facilitator to understand logic-based reasoning~\cite{fan2014computing,rizzo2018qualitative} as it conveys a notion of causality~\cite{lewis2013counterfactuals,pearl2018bookofwhy}.

\subsection{Norm Guided Reinforcement Learning Agent}
\label{sec:bg_neufeld}

In his work, Neufeld introduced a Norm Guided Reinforcement Learning (NGRL) agent~\cite{neufeld2022reinforcement,neufeld2023norm,neufeld2024learning}. This framework extends classical model-free reinforcement learning so that the agent learns preset norms in addition to achieving the task for which it was designed.

In this framework, the RL-agent evolves within a labelled MDP (see Definition~\ref{def:lmdp}). When entering a state, the labels serve at triggering some constitutive norms to generate institutional facts or activate regulative norms. Then, if one or more regulative norms clash with the labels, the system considers that violations occurred.\footnote{Although this work uses a labelled MDP, it could be flattened to a Normative MDP (NMDP)~\cite{fagundes2016design} as the norms in each state can be modelled by applying the set of constitutive rules.} This is handled by returning to the agent, in addition to the reward signal from the environment, a value corresponding to the number of violations committed when performing the latest transition.

The NGRL agent is equipped with an additional \qf $Q_V$. It follows a similar functioning as the one of the \qf tracking the expected reward (renamed $Q_R$), except that this one tracks the expected violations when doing a certain action in a certain state. Then, after the learning, the agent is not only aware of the expected reward it will get when doing a given action, but also of the violations it will commit. Using a preference order and a thresholded lexicographic selection~\cite{vamplew2011empirical} (see Algorithm~\ref{algo:tlex_intro}), it selects the optimal action with respect to the task among the ones that do not exceed a maximal value for the expected violations.

\begin{algorithm}[ht]
\caption{Thresholded lexicographic selection}
\label{algo:tlex_intro}
\begin{algorithmic}[1]
    \Require $s$ (current state), $L$ (a list of \qfs, ordered by priority), $C$ a vector of thresholds for the \qfs in $L$
    \Ensure $A^*$ (the resulting set containing the optimal actions)
    \State $A^* \coloneqq A(s)$ \Comment{Gets the set of the possible actions in state $s$}
    \For{$Q_i \in L$}
        \State $T \coloneqq \{x \in A^* | Q_i(s, x) \geq C_i\}$ \Comment{Filters out actions below the threshold}
        \If{$T = \emptyset$}
            \State $A^* \coloneqq \arg\max\limits_{x \in A^*} Q_i(s, x)$ \Comment{Filters out suboptimal actions}
        \Else
            \State $A^* \coloneqq T$
        \EndIf
    \EndFor
    \State \Return $A^*$
\end{algorithmic}
\end{algorithm}

In his work, Neufeld assumes the ensemble of the constitutive and regulative norms to be given. Same goes for the thresholds used for the lexicographic selection. While this may seem already as being a significant amount of expert knowledge (in particular since the structure of these norms may be complex to setup and so not scalable), this is still significantly better than the model-based approaches relying on Linear Temporal Logic for safe RL~\cite{neufeld2021normative,makarova2025deontically} that require either to know the transitions or at least to learn those.

%% file: sections/2_normative_architecture.tex
In this chapter, we present the \pino architecture~\cite{alcaraz2026pinocchio}. This architecture consists of a reinforcement learning agent supervised by a normative advisor. The normative advisor itself is composed of multiple stakeholders' reasoning models, called avatars. We will see how such an architecture can learn norms represented in a symbolic manner and follow them within a stochastic environment. We will also see how these norms may be rendered irrelevant in some specific contexts, and how the architecture takes this into consideration.

\section{Introduction}

Normative reinforcement learning is a timely topic, as many systems deployed in the real-world are now using Artificial Intelligence (AI). As these systems now evolve among us, we need to ensure that they follow settled norms and conventions in order to avoid causing harm to people they interact with. In his work, Neufeld proposed the Norm Guided Reinforcement Learning (NGRL) architecture~\cite{neufeld2022reinforcement,neufeld2023norm,neufeld2024learning}. This architecture puts a strong emphasis onto the respect of the norms (and in an alternate variant, places emphasis on maintaining the frequency of violations below a certain threshold) by approximating not only the expected accumulated reward when doing a certain action in a certain state, but also the expect accumulated violations.

By its simplicity and modularity, this work is very suitable for extensions. In fact, its functioning is very close to a standard $Q$-learning algorithm. Furthermore, most of its components can be easily replaced. In the architecture proposed in this dissertation, we make the choice to substitute the normative supervisor component with the Jiminy architecture~\cite{liao2019building,liao2023jiminy} that has been designed to account not only for the context, but also for heterogeneous viewpoints.

Let an agent evolving within an environment presenting normative considerations, or a normative dilemma. In each state, a set of actions is available to the agent. There can be several stakeholders (\emph{e.g.}, the user, the law enforcement, or the manufacturer) who are interested or affected by the decisions of the agent, and each may share different ethical considerations. The role of Jiminy is to confront the points of view and arguments of the stakeholder to estimate which action(s) the agent has the right to perform. As sometimes the decision of the next action has to be taken in less than a second, it is not realistic to communicate with the stakeholders, who are humans, in real-time. For this reason, Jiminy represents the stakeholders through avatars, which are essentially knowledge-bases of inference rules that serves to model the mental state of the stakeholders. When entering a state, the agent forwards its observation to Jiminy. There can be a preprocessing step to convert raw values into meaningful symbols (referred to as \emph{brute} facts) that can be used by the inference rules. Conflicts among the arguments of the avatars are represented using structured argumentation. Then, following a three-fold process, Jiminy tries to resolve the dilemma. Firstly, it analyses the relation between the arguments by constructing a structured argumentation graph and then converting it to an abstract argumentation. Computing an extension gives the legitimate actions that the agent can do. This may already solve the dilemma. Secondly, if the first step was not sufficient, Jiminy combines the norms and expertise of the stakeholders to resolve the dilemma. Thirdly, and only if the two previous methods failed, Jiminy uses context-sensitive rules to determine which of the stakeholders should take preference over the others.

This architecture has several benefits. The first one is its transparency. Indeed, structured argumentation and symbolic representations make it possible to see how each stakeholder constructs its arguments, and why such argument is attacking or conflicting with such other argument. The second main benefit is that it does not require a unified and consistent ethical model among the stakeholders, as conflicts are resolved through the argumentation process. This is a crucial point as requiring a unified view would boil down to having someone in charge of preprocessing the information of each stakeholder, essentially moving the scalability bottleneck to an upstream task. An alternative would be to allow each stakeholder to evaluate the situation separately, and take the majority voting, or apply a preference order, but this would remove the possibility for a stakeholder having a very relevant argument to share this information with the other stakeholders, although they could have changed their stance because of it.

However, Jiminy also suffers from some drawbacks. First, the fact that it is transparent raises concerns about the privacy and agency of the stakeholders. For example, if one of the stakeholders is the manufacturer, it may be undesired for him to release sensitive information about, for instance, the inner working of the product, or agreements that he passed with some other parties and that alter his normative system. It is also a loss of agency as the stakeholders are now deprived of the strategic aspect that formal argumentation may have through dialogue games (\emph{e.g.}, Fatio's protocol~\cite{mcburney2004locutions,mcburney2004syntax}, discussion games~\cite{caminada2015discussion}) as they are forced to make public the entirety of their knowledge. Consequently, information that was initially hidden to the other stakeholders can now be turned against them. This may incentivise stakeholders to provide a very limited amount of information for the modelling of their avatar. Second, the current working of the Jiminy architecture requires quite an extensive amount of knowledge, through the provision of the inference rules required to instantiate the attacks among the arguments, and computation resources, as it not only requires to compute an extension from an abstract argumentation graph, but also needs to run an inference model to build the arguments from the raw inputs of the state. Finally, Jiminy acts proactively, meaning that it does not judge if the action of an agent was good afterward, but instead plans ahead what is legitimate. This is very unrealistic, in particular within a partially observable stochastic environment containing a large number of elements or factors in its input.

In order to combine the reasoning of Jiminy with the NGRL agent, we use a modified version of the AJAR framework~\cite{alcaraz2023ajar,alcaraz2026combining}. This framework consists of a reinforcement learning agent whom the reward is not directly obtained from the environment (\emph{i.e.}, MDP) but rather from a set of judging agents. Each of these judging agent has its decision model for a given moral value represented through an argumentation graph. At each time step, it evaluates the consequences of the actions of the judged agent and accordingly generates a reward. All the rewards of the judges are then aggregated and returned to the agent. This framework has several advantages. First, it is very modular, as each judge can be added or removed without having to do any other change. Second, it brings additional intelligibility compared to standard reinforcement learning. Not only is the model of each judge built expertly, using symbols that make it readable and understandable by non-experts, but also it is possible to look at the graph structure for a given state to determine why the learning agent considered that doing such action would improve its reward. However, this framework also suffers from some disadvantages. For instance, the aggregation method is somehow arbitrary. Furthermore, since there is a single judge per moral value, only one ethical principle is adopted, and views are not confronted. Last, AJAR was originally designed for Machine Ethics. This field shares similarities with the learning of norms, but also major differences. For example, ethics often balances the compliance to moral values with the sacrifice of the agent with respect to its main task. In normative systems, adherence to norms is usually not an option but an obligation. The agent is then not free to decide whether a reward is important enough so that undertaking an action that violates the norm would be considered legitimate. Another example is that in ethics, the agent must recognise situations that are considered unethical, whereas in normative systems, the agent simply follows the norms without reevaluating them.

The next sections will aim at proposing an approach, building upon the three aforementioned architectures, to normative reinforcement learning. This approach, which we call \pino~\cite{alcaraz2026pinocchio}, follows the principles of the NGRL agent as proposed by Neufeld, but replaces its normative guide by a more sophisticated version derived from the Jiminy architecture. Then, the two are connected through the AJAR framework.

\section{Model}
\label{sec:model_ch1}

\subsection{Context-Aware Normative Reasoning}
\label{sec:contextaware}

In this section, we introduce an architecture of normative supervisor inspired by the Jiminy architecture~\cite{liao2019building,liao2023jiminy}. This component will be referred to as the judge, as its role will be similar. In fact, it will have to evaluate the consequences of the actions taken by an agent and determine which norm is violated by the outcome of this action.

Unlike Jiminy, the normative conflicts are no longer handled within this component but will instead be managed through the reinforcement learning component.

If a norm is activated, that is, its condition is satisfied by the relevant considerations, \emph{i.e.}, a set of propositional atoms extracted from the preprocessed state, the system checks whether the agent complies with the norm or not. If not, then the argumentation graph built from the avatars is used to compute whether the norm is defeated, \emph{i.e.}, the current situation creates an exception to the norm or not. If the norm is not defeated, then because the agent does not comply with it, the agent is committing a violation with respect to this norm.

Another difference with Jiminy is that here the goal is not to decide which action the agent should or should have taken, but rather to decide whether the reached state adheres to the norms. As such, norms are no longer used as arguments that attack or support actions. Instead, they are the elements that are discussed within the argumentation process. Each norm is discussed separately, similar to the AJAR architecture~\cite{alcaraz2023ajar,alcaraz2026combining} where moral values (since it is originally a work on Machine Ethics) are debated individually.


We give a more formal definition of the judge in the immediately following Definitions~\ref{def:stakeholder}--\ref{def:advisor}.

\begin{definition}[Stakeholder Model]\label{def:stakeholder}
    Let a finite set of arguments \args and a set of regulative norms $N$ be given by the environment. A stakeholder model (or avatar) is a tuple $\mathcal{M}_i = \langle \mathcal{L}, C, \defs, M, getArgs\rangle$ where:

    \begin{itemize}\tightlist
        \item $\mathcal{L}$ is a language defined by a set of well-formed formulae (wff). One requirement is that $\args\subseteq\mathcal{L}$
        \item $C$ is a set of constitutive norms $C_i$ of the form $C_i(a, b)$ with $a, b \in \mathcal{L}$ where $b$ is either an argument (in the sense of abstract argumentation) or an institutional fact that can serve to trigger other constitutive norms. Note that, in addition to other possible constitutive norms, this set always contains the norms $C_r(q, r)$, with $ r \in N$ a regulative norm, which serves to denote the fact that each regulative norm discussed by the judges always forms one argument. Constitutive norms are useful for constructing abstractions from facts so that the application of a norm can be generalised~\cite{tomic2020learning}\footnote{For example, to make the norm ``\textbf{F}(\texttt{steal(item)}) usable, it is possible to abstract the objects ``milk'' and ``bread'' as ``item'' by using a constitutional norm.}
        \item \defs is an attack relation \emph{s.t.} $\defs \subseteq \args \times \args$
        \item $M : N \to 2^{C}$ a mapping function from a regulative norm in $r \in N$ to a subset of the constitutive norms of the stakeholder model. We then define, respectively,
        \begin{itemize}
            \item $\Pi_r C = \{C_i | \exists i \in I. C_i(a, b) \in M(r)\}$ the subset of constitutive norms considered relevant by the agent when arguing about $r \in N$, with $I$ the set of constitutive norm indices
            \item $\Pi_r \args = \{b | \exists x. C_i(x, b) \in M(r)\}$ the subset of arguments considered relevant by the agent when arguing about $r \in N$
            \item $\Pi_r \defs = \{ x\defs y \in \defs | x \in \Pi_r \args \lor y \in \Pi_r \args \}$, the subset of defeats considered relevant by the agent when arguing about $r \in N$. Note that in the latter case $a\defs b \in \Pi_r \defs \rightarrow a, b \in M(r)$.

        \end{itemize}
        \item $getArgs(x)$ is the function that returns the set of all the arguments that can be constructed from a given set of observations $x \in \mathcal{L}$. It is defined by the closure of the constitutive norms as follows:
        \begin{itemize}\tightlist
            \item $Cl(k) = \bigcup_n Cl_n(k)$
            \item $Cl_0(k) = k$ and $Cl_{n+1}(k) = \{b | \exists C_i(a, b). a \in Cl_n(k)\}$
        \end{itemize}
    \end{itemize}
\end{definition}

\begin{definition}[Advisor]\label{def:advisor}
    An advisor \adv is a tuple $\langle \sh, \facts, \mathcal{L}, N, \succ^\mathcal{M}, \epsilon,  j\rangle$ where:

    \begin{itemize}\tightlist
        \item \sh is a set of stakeholders' models, referred to as avatars
        \item \facts is a finite set of facts, \emph{i.e.}, propositional atoms used to describe the world
        \item $\mathcal{L}$ a logical language
        \item $N$ is a set of regulative norms of the form $\textbf{X}(p|q)$ with $\textbf{X} \in \{\textbf{O}, \textbf{P}, \textbf{F}\}$ and $p,q \in \mathcal{L}$
        \item $\succ^\mathcal{M}$ is a strict preference order over \sh
        \item $\epsilon : states \to 2^{\facts}$ a function that takes a state (usually a raw input of the environment) and returns a subset of \facts that corresponds to the propositional atoms that can be constructed from the input state
        \item $j : 2^{\facts} \times N \times 2^\sh \to \{-1; 0\}$ is a function that takes as input a set of facts obtained from $\epsilon(s)$ where $s$ is the current observation of the environment, a norm to discuss, and a set of stakeholders, and that returns whether or not the discussed norm has been violated. This function can be described by a four step pipeline:
        \begin{enumerate}\tightlist
            \item For a state $s$ and a regulative norm $r \in N$, and for each stakeholder $m \in \sh$, compute the set of arguments relevant to the given norm that can be constructed from the facts $S_0^m = \Pi_r^m\args \cap getArgs(\epsilon(s))$
            
            \item Reunify the sets of arguments returned by the stakeholders $S_1^{\args} = \bigcup_m^M S_0^m$. Do the same for their attack relation $S_1^{\defs} = \bigcup_m^M \Pi_r^m\defs$
            
            \item With the union of the arguments and the attacks, construct an argumentation framework $\langle S_1^{\args}, S_1^{\defs}\rangle$ and compute an extension $Ext$. We recommend the grounded extension for its polynomial-time complexity and uniqueness property.\footnote{Any extension can work in principle, as long as there is a way to decide if the argument represented by the discussed norm is part of it. For example, if one uses preferred semantics, he may decide to consider the norm as accepted within the extension if it is sceptically accepted. Stable extensions are also exhibiting interesting properties, but their existence is not guaranteed.} If the argument corresponding to the norm is considered as undecided by the extension semantics, the preference order $\succ^\mathcal{M}$ is applied (see Example~\ref{ex:preference}). If even after this the norm argument is still undecided, then it is considered as not part of the extension, \emph{i.e.}, sceptically rejected
            
            \item Return $-1$ if there is a non-compliance (for example, there is $\textbf{O}(p|q)$ with $p \notin\epsilon(s)$ and $q \in\epsilon(s)$) and if the norm discussed is part of the extension previously computed, \emph{i.e.}, $r \in Ext$. Return $0$ otherwise, which means that no violation occurred.
        \end{enumerate}
        
    \end{itemize}
\end{definition}

\begin{example}[Applying the preference order]\label{ex:preference}
    Let two arguments $a, b \in \args$ that have a symmetric attack (\emph{i.e.}, $a\defs b$ and $b \defs a$). For each of these arguments, we compute the set of avatars that find it relevant, \emph{i.e.}, for an argument $x \in \args$, the set of stakeholders $S_x = \{m \in \sh | x \in \Pi_r^m\args\}$. We then obtain $S_a$ and $S_b$. If there exist $s \in S_a$ \emph{s.t.} for all $p \in S_b$, $s \succ^\mathcal{M} p$, then we remove $b\defs a$ from the union of the attack relations $S_1^{\defs}$ obtained at step $2$ of the function $j$. On the other hand, if there exist $s \in S_b$ \emph{s.t.} for all $p \in S_a$, $s \succ^\mathcal{M} p$, then we remove $a\defs b$ from $S_1^{\defs}$.
\end{example}

\begin{remark}\label{rem:constitutive}
    While the architecture allows one to chain constitutional rules over and over, we believe that this should be avoided, as it would reintroduce complexity in designing arguments and their structure, which this architecture aims to reduce. Instead, one should favour the use of the $\epsilon$ function such that enough facts are generated from it so that the arguments can be directly triggered through the application of a single constitutional rule. In short, the designer should generate as many facts as possible without trying to anticipate their purpose or how they could be organised. Meanwhile, the avatar designer's task is to determine which arguments are relevant, under which conditions, and which arguments they defeat or are defeated by.
\end{remark}

We made the choice to use the language \prop, which is a propositional language. This choice is motivated by the fact that it is decidable, less expensive in terms of computations than most other logics, and its expressiveness is sufficient for the needs of the proposed architecture (assuming we stick to the advice given in Remark~\ref{rem:constitutive}). Nevertheless, we extend it with the \texttt{not} unary connective that means that for a given knowledge-base \kb, $\texttt{not } \phi$ is the same as $\kb \not\models \phi$. It is defined as the set of well-formed formulae (wff), with the following BNF grammar for any $p \in \facts$ by $\phi \coloneqq \bot \; |\;  p \; |\; \neg \phi\; |\;  \phi \vee \psi\; |\; \texttt{not } \phi$, as well as the following notation shortcuts:
\begin{itemize}\tightlist
    \item $\top \coloneqq \neg \bot $, 
    \item $\phi \wedge \psi \coloneqq \neg(\neg \phi \vee \neg \psi)$,
    \item $\phi \rightarrow \psi \coloneqq \neg \phi \vee  \psi$
\end{itemize}


Let the truth domain be $\{\top, \bot, \texttt{undef}\}$. Then:

An interpretation model $I$ over a valuation $V: \facts \to \{\top,\bot,\texttt{undef}\}$ is given by the function $I: \prop \to \{\bot,\top,\texttt{undef}\}$ \emph{s.t.} $\forall \phi \in \prop, I \models \phi \text{ iff } I(\phi) = \top$ where:

\begin{enumerate}
    \item $\forall p \in \facts$, $I(p) = V(p)$
    \item $I(\neg \phi) =$
        $\begin{cases}
            \top & \text{if } I(\phi) = \bot \\
            \bot & \text{if } I(\phi) = \top \\
            \texttt{undef}    & \text{if } I(\phi) = \texttt{undef}
        \end{cases}$
    \item $I(\phi \vee \psi) =$
        $\begin{cases}
            \top & \text{if } I(\phi) = \top \text{ or } I(\psi) = \top \\
            \bot & \text{if } I(\phi) = \bot \text{ and } I(\psi) = \bot \\
            \texttt{undef}    & \text{otherwise}
        \end{cases}$
    \item $I(\texttt{not } \phi) =$
        $\begin{cases}
            \top & \text{if } I(\phi) = \texttt{undef} \\
            \bot & \text{otherwise}
        \end{cases}$
\end{enumerate}

The definitions introduced above allow us to compute, for a given norm, a given set of observations (that potentially serves at building higher-level facts through the constitutive norms~\cite{aldewereld2010making}), and multiple avatars, whether or not the agent reached a state where it commits a violation, \emph{i.e.}, does not comply with a non-defeated norm. In this model, only the consequences of an action\footnote{The term ``action'' is used in a very broad meaning here, as for an artificial agent ``not doing any thing'' can still be considered as being an action.} are judged. Consequently, the avatars do not need to account for the probabilities which arguments such as ``There was a high probability not to harm anyone''. This simplifies the task of designing the arguments and the attack relation of the stakeholders.

\begin{figure}[ht]
    \centering
    \includegraphics[width=0.6\linewidth]{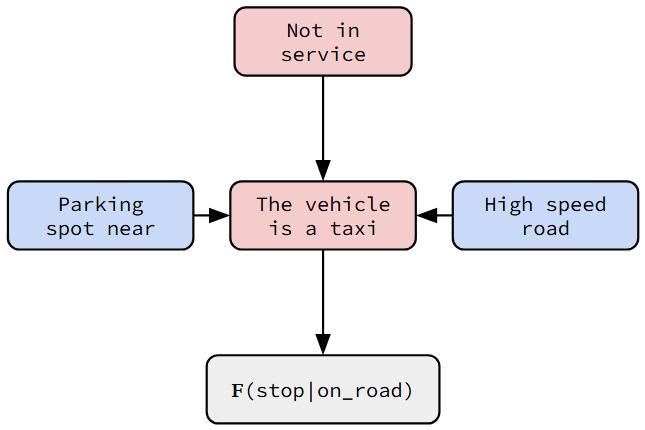}
    \caption{Example of graph generated by two avatars.}
    \label{fig:debate}
\end{figure}

To better illustrate the definitions, an example of a graph created from avatars debating the application of the norm $\textbf{F}(\texttt{stop}\mid\texttt{on\_road})$ is shown in Fig.~\ref{fig:debate}. In this example, there are two stakeholders, namely a taxi company, with the red arguments, and the law enforcement, with the blue arguments. We may add more, such as the customer, the local administration of the city, etc, but for the sake of this example we limit it to the aforementioned ones. The graph can be read as follows. First, the norm $\textbf{F}(\texttt{stop}\mid\texttt{on\_road})$ is activated only if the agent is on the road. If this is the case, then we see that there is an exception to this norm if the agent is a taxi, as we may wish to allow taxis to make short stops on the road to let customers in and out. This argument is supported by the stakeholder model of the taxi company. This same stakeholder defeats its own exception in the event that the agent is not currently in service. On the other hand, the law enforcement's model removes this exception made for taxis to stop on the road if there was a parking spot just near, or if the road in question is a high-speed road.

\subsection{A Normative Supervisor for a Normative Agent}

In this section, we detail how we use the judge component introduced in Section~\ref{sec:contextaware} to generate a reward signal to train a reinforcement learning agent to follow norms. The resulting architecture is called \pino ($\pi$: Reinforcement Learning, Normative CCHIO)~\cite{alcaraz2026pinocchio}.

\begin{figure}[ht]
    \centering
    \includegraphics[width=0.7\linewidth]{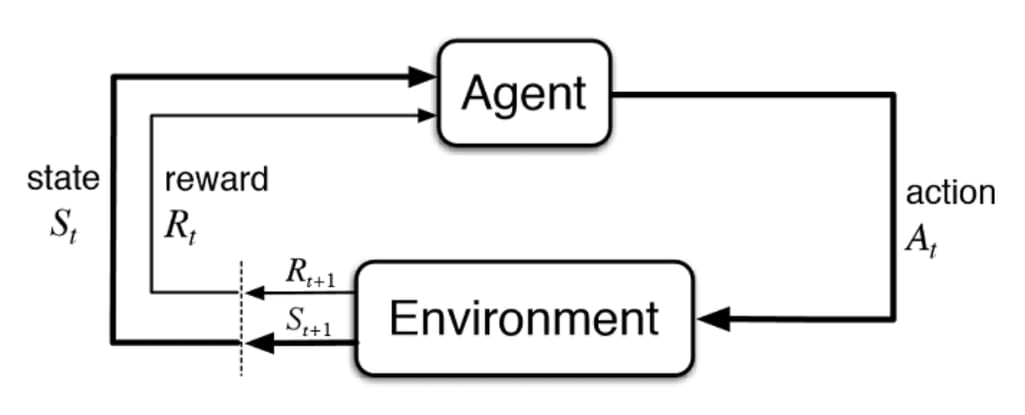}
    \caption{Diagram of a standard reinforcement learning training loop.}
    \label{fig:rl_training_loop}
\end{figure}

\begin{figure}[ht]
    \centering
    \includegraphics[width=0.6\linewidth]{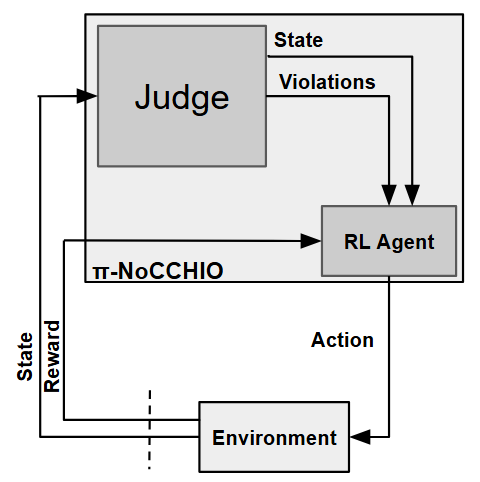}
    \caption{Diagram of the \pino training loop.}
    \label{fig:pino}
\end{figure}

Fig.~\ref{fig:rl_training_loop} is a diagram of a standard RL training loop. In comparison, a diagram of \pino is shown in Fig.~\ref{fig:pino}. Similarly to a reinforcement learning architecture, it can be divided into two main blocks. One is the environment, represented with an MDP (see Definition~\ref{def:mdp}). The second block is \pino. Inside this component, we can identify further subdivisions. Specifically, it consists of two subcomponents: The judge and the RL agent.

The first component is inspired by this work, \emph{i.e.}, the AJAR~\cite{alcaraz2023ajar,alcaraz2026combining} framework. In this work, some judging agents assess whether an RL agent acted ethically with respect to predefined moral values by altering the reward obtained from the environment. It is repurposed here for normative RL.

The judge receives the observations ``state'' for step $s_{t+1}$ resulting from the action(s) performed by the agent in step $s_t$. Then, two signals exit the judge component and are transmitted to the agent.

The first is the raw observation of the environment that is sent to the agent for the current step $s_{t+1}$. While it can remain unchanged, it can also be manually modified. The person doing such modification will be called ``Operator''. This opens up the possibility of indicating the norms that the agent will have to comply with in the next state. Indeed, by aggregating to the state, for each norm, one label among $\{\texttt{on}, \texttt{off}, \texttt{none}\}$ (more formally for a state $s$ and a set of norms $N$, we have $s \coloneqq s \cup \{\texttt{on}, \texttt{off}, \texttt{none}\}^{|N|}$), it is possible to communicate to the agent whether a given norm will be considered activated and non-defeated in the next state no matter what (\texttt{on}), or will be defeated (\texttt{off}), or let to the decision of the judges (\texttt{none}). This allows to manipulate the behaviour of the agent post-learning phase. The agent would then adapt its behaviour accordingly. During the learning phase, manually overriding a norm\footnote{This would be done randomly to simulate when a human may desire to take the control.}, which will aggregate the extra information to the state $s_t$, takes the priority over the avatars. Thus, instead of returning to the agent, for each norm, the signal $j(\epsilon(s_{t+1}))$, the value set in the previous state $s_t$ is used.

\begin{remark}
    Allowing for the manual editing of the status of the norms has several advantages, but also some drawbacks. On one hand, it gives more control to the final user, as it is possible to provide extra guidance to the agent whenever this is required, as a human user may better understand the situation or wish to deviate from the avatars' guidance. On the other hand, it potentially makes the system more vulnerable to cyberattacks, as one could potentially breach it to modify the status of the norms.\footnote{Note that modifying the norms themselves, or adding new norms, is not feasible, as the agent would not have been trained with it and would then not be able to understand what it ought to do according to these news norms.} However, this can be partially solved by splitting the norms into two sets. One set in which the norms can be manually edited (this set containing the least sensitive norms), and the other set for which it is not possible to override the system. Since the agent's state would not include these protected norms, their defeat status could not be altered.
\end{remark}

The second signal indicates to the agent whether or not it committed a violation by performing a certain action at time step $s_t$. This signal is similar to the reward signal in a standard MDP.

\begin{remark}
    Here, we present a simple version of \pino where all the norms are considered equal in terms of importance. The agent will then learn to minimise the number of violations committed. However, this violation signal can be channelised so that the agent knows exactly which norms were violated. This can then be used, in combination with a preference order over the norms, to favour the respect of some over the others, especially in the event of a normative conflict. On the other hand, if one prefers to use weights for the norms rather than a preference order, it is possible to keep the first approach. In the event of a violation, the function $j$ would not return $-1$ but instead return $-w_n$ which corresponds to the negative of the weight associated with the norm $n$. Doing this would also account for the accumulation of the violations of norms with low weights that end up being worse than the violation of a single norm with a high weight.
\end{remark}

\begin{figure}
    \centering
    \includegraphics[width=0.9\linewidth]{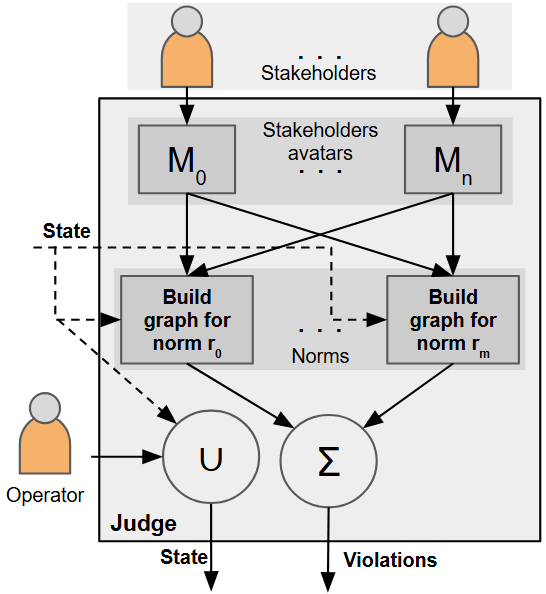}
    \caption{Diagram of the judge component.}
    \label{fig:judge}
\end{figure}

A diagram of the judge component can be seen in Fig.~\ref{fig:judge}. It features $n+1$ avatars ($\mathcal{M}_{0-n}$) and $m+1$ regulative norms ($r_{0-m}$). The block $\cup$ represents the aggregation of the (potential) edits of the operator with the state $s_{t+1}$. The block $\Sigma$ represents the summing of the judgements, $j(\epsilon(s_{t+1}), r_i, \{\mathcal{M}_0, \ldots, \mathcal{M}_n\})$, for each norm (note that, as mentioned earlier, this step can be skipped if one wants to channelise the norms within individual signals).

The second component of this architecture is the RL agent, inspired by the one proposed by Neufeld~\cite{neufeld2022reinforcement,neufeld2023norm,,neufeld2024learning}. It receives the signal corresponding to the reward obtained from the environment, as well as the extended state and the violation signal from the judge component. It differs slightly from a standard reinforcement learning agent as it does not possess a single \qf, but two of them, namely $Q_R$ and $Q_V$.\footnote{In finite MDPs---either due to terminal states or a fixed iteration limit---it is advisable to set the discount factor $\gamma = 1$ when updating $Q_V$. This choice ensures that future, possibly distant but inevitable violations are not artificially devalued, allowing the agent to correctly account for unavoidable consequences in its decision-making.}

\begin{remark}
    While it would be possible to treat violations as a penalty over the reward signal, or any other kind of trade-off~\cite{tomic2019robby,li2015reinforcement,panagiotidi2013towards}, this should be avoided. The main reason is that, in case of an unbounded reward, there is no guarantee that the agent will not find an interest in violating a norm to reach a very high reward.\footnote{This conclusion is similar to the one of Blaise Pascal's argument~\cite{pascal1670pensees}, in which he observes that if God exists and one believes, one gains infinite happiness (heaven), whereas if God exists and one does not believe, one loses infinitely (damnation). If God does not exist, the gains or losses of believing are finite. Therefore, it is rational to believe in God.}
    
    On the other hand, one may be tempted to multiply the reward signal by $0$ in the event of a violation, but this raises two other problems, which are the increase of the reward if this was a negative value (since it passes from some $-n, n\in \mathbb{N}$ to $0$), and the impossibility of making two or more violations count worse than a single one (since both $n \times 0$ and $n \times 0 \times 0$ are equal to $0$). If one tries to address this issue by multiplying by a number within $]0; 1[$ instead, it comes back to the first issue.
\end{remark}

The first tracks the expected reward, while the second tracks the expected amount of violations. To select its optimal action, this agent uses a preference order, defined as $\mathcal{P} \coloneqq \{V \succ R\}$, combined with a lexicographic selection, detailed in Algorithm~\ref{algo:lex}. The main idea is that, considering a set of doable actions $A(s)$ in state $s$, the agent first select a subset of it that maximises\footnote{As doing a violation returns the signal $-1$, the value $0$ corresponds to no violation.} the expected value of $Q_V$. This corresponds to the set of actions that limit as much as possible the commitment of violations.\footnote{This does not fully prevent violations from occurring. In a state where all the outcomes lead to at least one violation, the agent will still choose an action.} Then, from this subset, it takes the action that maximises the expected value of $Q_R$, and that corresponds to the most optimal action in relation to a given task among the actions that minimise violations.

\begin{algorithm}
\caption{Lexicographic selection}
\label{algo:lex}
\begin{algorithmic}[1]
    \Require $s$ (current state), $L$ (a list of Q-functions, ordered according to a preference order)
    \Ensure $A^*$ (the resulting set containing the optimal actions)
    \State $A^* \coloneqq A(s)$ \Comment{Gets the set of the possible actions in state $s$}
    \For{$Q_i \in L$}
        \State $A^* \coloneqq \arg\max\limits_{x \in A^*} Q_i(s, x)$ \Comment{Filters out suboptimal actions}
    \EndFor
    \State \Return $A^*$
\end{algorithmic}
\end{algorithm}

\subsection{Running Example}
\label{sec:runningexample}

Once the learning phase is over, it is no longer required to access the avatars. Consequently, they can be removed from the deployed system so that the private information of each stakeholder is kept secret. In the event of the agent causing an accident or harm\footnote{It is also possible to do this for a simple malfunction, but we believe that for the sake of privacy, just like law enforcement needs a mandate to search in the house of someone, they would require a mandate as well to access the avatars.}, it is possible to reconnect the avatars so that they can compute the status of the norms for the state corresponding to the incident, providing hints to the investigator about what could have led the agent to perform such action.

To better understand how the \pino architecture works, hereby we detail a running example to illustrate first the learning process, and second the choice of an optimal action in a given state. In this running example, we will consider an autonomous taxi, as well as two stakeholder which are the taxi company and the law enforcement. In Section~\ref{sec:example_judging}, we will show the process of judging the action of the agent goes. Then, in Section~\ref{sec:example_optimal}, we will show how the agent computes the optimal decision it ought to do once the learning phase has been performed.

\subsubsection{Judging an Action}
\label{sec:example_judging}

Let an agent be an autonomous driving taxi. It features two avatars such that $\sh = \{\sh_{taxi}, \sh_{law}\}$ that correspond to the taxi company and the law enforcement. There are two norms within the environment, which are the respect of the speed limit (\emph{i.e.}, $\textbf{F}(\texttt{speeding})$) and the parking regulations that prohibit stopping on the road (\emph{i.e.}, $\textbf{F}(\texttt{stop}\mid\texttt{on\_road})$).

Consider the following constitutive rules for $\mathcal{M}_{taxi}$

\begin{enumerate}\tightlist
    \item $C_1(\texttt{is\_taxi},\ \texttt{The\_vehicle\_is\_a\_taxi})$ - The agent is a taxi
    
    \item $C_2(\texttt{not in\_service},\ \texttt{Not\_in\_service})$ - The agent is currently not in service
    
    \item $C_3(\texttt{expected\_greater\_than\_target\_time},\ \texttt{Customer\_is\_late})$ - The destination cannot be reached before the time is exceeded
    
    \item $C_4(\texttt{no\_traffic},\ \texttt{No\_traffic})$ - There are very few cars on the road
    
    \item $C_{\textbf{F}(\texttt{speeding})}(\top,\ \textbf{F}(\texttt{speeding}))$ - In any context, it is forbidden to exceed the speed limit
    
    \item $C_{\textbf{F}(\texttt{stop}\mid\texttt{on\_road})}(\texttt{on\_road},\ \textbf{F}(\texttt{stop}\mid\texttt{on\_road}))$ - When the agent is on the road, it is forbidden for it to stop
\end{enumerate}

and for $\mathcal{M}_{law}$

\begin{enumerate}\tightlist
    \item $C_1(\texttt{closest\_parking<=40m} \land \texttt{nb\_free\_spots>=10},\ \texttt{Parking\_spot\_near})$ - There is a parking with places left near
    
    \item $C_2(\texttt{in\_city},\ \texttt{No\_exception\_in\_cities})$ - In a city, there is no exception that can be applied
    
    \item $C_3(\texttt{overtaking},\ \texttt{Overtaking})$ - The agent is endorsing an overtaking action
    
    \item $C_4(\texttt{speed\_excess>=30},\ \texttt{30+ kph Speeding})$ - The agent exceeds the speed limit by more than $30$ kph
    
    \item $C_5(\texttt{on\_highway},\ \texttt{High speed road})$ - Highways are considered as high speed roads
    
    \item $C_{\textbf{F}(\texttt{speeding})}(\top,\ \textbf{F}(\texttt{speeding}))$ - In any context, it is forbidden to exceed the speed limit
    
    \item $C_{\textbf{F}(\texttt{stop}\mid\texttt{on\_road})}(\texttt{on\_road},\ \textbf{F}(\texttt{stop}\mid\texttt{on\_road}))$ - When the agent is on the road, it is forbidden for it to stop
\end{enumerate}

Now, let a state $s$ where the autonomous taxi, \emph{i.e.}, the learning agent, is stopped on the road, within a city, next to an almost empty parking lot.

A view of how the state $s$ may look like for this running example is given in the ``State'' column of Table~\ref{tab:observations}.

First, \pino computes $\epsilon(s)$, which will provide a set of (true) atomic propositions, derived from the state $s$, that is a subset of $2^{\facts}$. There are several ways of doing this. The most natural and easy way is to associate some Boolean formulae with atomic propositions. For instance, it can be something similar to $$\text{``\texttt{IF agent.speed $>=$ environment.speed\_limit THEN facts.append(speeding)}''}$$ The second method requires the use of functions to generate the propositions dynamically. For example, we may create a function $nbCarNearby(s)$. This function could return a set of propositions, such as \{\texttt{car\_nearby=2}, \texttt{car\_nearby}$\leq 5$, \ldots\}. The last method consists of making use of another AI system, such as a classifier, a predictor, an oracle, or a Large Language Model (LLM), to extract the proposition from an unstructured state such as an image, a sound, a text, or a map. An example of extracted facts is given in the column ``Facts'' of Table~\ref{tab:observations}. However, this last method is more expensive in resources and more complex to setup.

Then, the constitutive norms are applied to the set of facts to generate more facts, and eventually, generate the arguments. The column ``Arguments'' of Table~\ref{tab:observations} shows the arguments that can be built from the facts previously derived from the constitutive norms of the avatars. Note that the norm arguments are built if their condition is within the facts. Since $\textbf{F}(\texttt{speeding})$ has $\top$ as a conditional, it is always part of the generated arguments.

\begin{table}
    \centering
    \caption{Example of a state ($s$), with the facts ($\epsilon(s)$) and arguments ($getArgs(\epsilon(s))$) that can be built from it.}
    \begin{tabular}{|l|l|l|}\hline
        \textbf{State} & \textbf{Facts} & \textbf{Arguments}\\\hline
        \texttt{speed=0kph} & \texttt{stop} & \texttt{The\_vehicle\_is\_a\_taxi}\\
        \texttt{area\_type=urban} & \texttt{in\_city} & \texttt{No\_exception\_in\_cities}\\
        \texttt{terrain=road} & \texttt{on\_road} & \texttt{Parking\_spot\_near}\\
        \texttt{vehicle\_type=taxi} & \texttt{is\_taxi} & \texttt{No\_traffic}\\
        \texttt{begin\_service\_time=8:00} & \texttt{in\_service} & $\textbf{F}(\texttt{speeding})$\\
        \texttt{end\_service\_time=19:00} & \texttt{nb\_car\_nearby=1} & $\textbf{F}(\texttt{stop} \mid \texttt{on\_road})$\\
        \texttt{time=16:23} & \texttt{no\_traffic} & \\
        \texttt{$\langle$Img. of the surroundings$\rangle$} & \texttt{closest\_parking<=40m} & \\
        \texttt{passenger\_count=0} & \texttt{nb\_free\_spots>=10} & \\
        \texttt{customer\_requested\_stop=true} &  & \\
        \texttt{$\langle$GPS information$\rangle$} &  & \\\hline
    \end{tabular}
    \label{tab:observations}
\end{table}

\begin{figure}[ht]
    \centering
    \begin{minipage}[t]{0.49\linewidth}
        \centering
        \includegraphics[width=\linewidth]{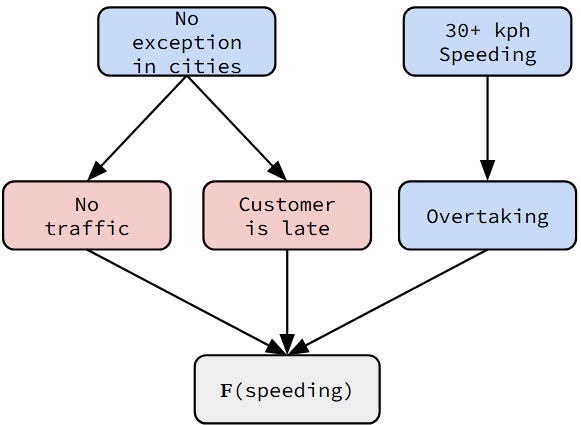}
        \caption{Argumentation graph for the norm $\textbf{F}(\texttt{speeding})$.}
        \label{fig:example_speeding}
    \end{minipage}%
    \hfill
    \begin{minipage}[t]{0.49\linewidth}
        \centering
        \includegraphics[width=\linewidth]{img/example_stopping.png}
        \caption{Argumentation graph for the norm $\textbf{F}(\texttt{stop} \mid\texttt{on\_road})$.}
        \label{fig:example_stopping}
    \end{minipage}
\end{figure}

\begin{figure}[ht]
    \centering
    \begin{minipage}[t]{0.49\linewidth}
        \centering
        \includegraphics[width=\linewidth]{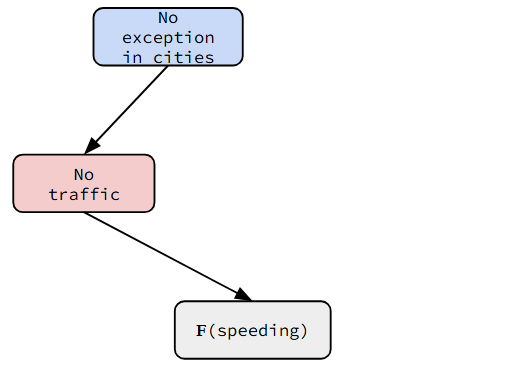}
        \caption{Argumentation graph in state $s$ for the norm $\textbf{F}(\texttt{speeding})$.}
        \label{fig:example_speeding_context}
    \end{minipage}%
    \hfill
    \begin{minipage}[t]{0.49\linewidth}
        \centering
        \includegraphics[width=\linewidth]{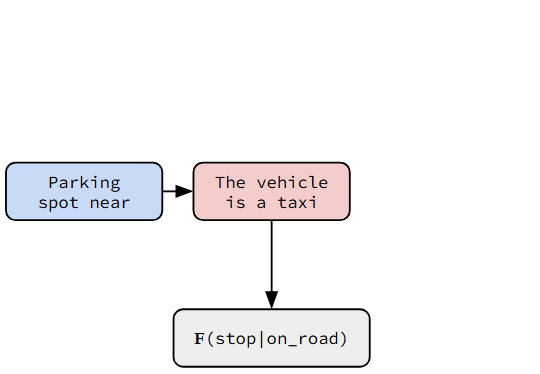}
        \caption{Argumentation graph in state $s$ for the norm $\textbf{F}(\texttt{stop}\mid\texttt{on\_road})$.}
        \label{fig:example_stopping_context}
    \end{minipage}
\end{figure}

Then, the arguments generated by each avatar can be combined into a single graph per norm. Fig.~\ref{fig:example_speeding}--\ref{fig:example_stopping} show how the graph for each norm would be constructed when combining all the arguments of the avatars, while Fig.~\ref{fig:example_speeding_context}--\ref{fig:example_stopping_context} show the same graphs, but only with the arguments retained in this example's state (see Table~\ref{tab:observations}). Arguments considered relevant by $\sh_{taxi}$ are in red, and the ones considered relevant by $\sh_{law}$ are in blue.\footnote{While this is not the case here, it is possible for stakeholders to share some arguments.} The norm argument is in grey.

Looking at Fig.~\ref{fig:example_speeding_context}, we can see that the norm $\textbf{F}(\texttt{speeding})$ is part of the grounded extension. As the agent is in a city, the argument of $\sh_{taxi}$ ``\texttt{No\_traffic}'' is defeated. The norm is then active and not defeated. Since the agent is stationary, it is not speeding, and so complies with this norm. As such, $j(\epsilon(s), \textbf{F}(\texttt{speeding}), \sh) = 0$ which means that the agent did not violate this norm.

Then, looking at Fig.~\ref{fig:example_stopping_context}, we can see that the norm $\textbf{F}(\texttt{stop}\mid\texttt{on\_road})$ is activated (since the agent is currently on the road) and not defeated as well. Albeit the agent is a taxi (and that this attacks the norm), the fact that there is a parking spot near defeats this argument, and thus reinstantiate the norm. The agent is not complying with the norm, as the norm says that it is prohibited to stop, and the agent is currently stopped. Consequently, $j(\epsilon(s), \textbf{F}(\texttt{stop} | \texttt{on\_road}), \sh) = -1$ which means that the agent violated this norm.

The agent then receives its reward signal from the environment, as well as the violation signal equal to $-1$. Using the $Q$-learning algorithm, it updates its \qvs for $Q_R$ and $Q_V$.

\subsubsection{Selection of the Optimal Action}
\label{sec:example_optimal}

Once the learning phase is over, the agent has to follow the optimal policy $\pi^*$ it learnt during the training. Consider an example scenario with a state $s$ and the norms presented earlier in Fig.~\ref{fig:example_speeding}--\ref{fig:example_stopping} in which it has to pick up a customer who is standing nearby. This would grant an immediate reward of $+100$. On the other hand, since the agent should keep the waiting time of the customer as short as possible, it is given a negative reward (for instance $-1$) for every time step in which the customer is waiting.

Let the action space for the agent in this state be $A(s) = \{$\texttt{stop}, \texttt{drive\_at\_30kph}, \texttt{drive\_at\_50kph}, \texttt{drive\_at\_70kph}$\}$. The information that can be derived from the state is that the speed limit is $50$ kph since the agent is within a city, and there is a parking spot nearby. \qvs $Q_R(s, a)$ and $Q_V$ for the given state $s$ and for each action $a \in A(s)$ are given in Table~\ref{tab:qvalues}.

\begin{table}
    \centering
    \caption{Expected \qvs for a fake scenario with a state $s$ and a list of actions $a \in A(s)$.}
    \begin{tabular}{c|c|c}\hline
        \textbf{Actions} & $Q_R(s, a)$ & $Q_V(s, a)$\\\hline
        \texttt{stop} & $+100$ & $-1$\\
        \texttt{drive\_at\_30kph} & $+40$ & $0$\\
        \texttt{drive\_at\_50kph} & $+60$ & $0$\\
        \texttt{drive\_at\_70kph} & $+80$ & $-1$\\\hline
    \end{tabular}
    \label{tab:qvalues}
\end{table}

As there is a parking spot nearby, even though the agent is a taxi, it is still prohibited to stop on the road. Consequently, the action \texttt{stop} would grant the immediate reward of $+100$ for letting the customer in, but would also count as a violation of the norm $\textbf{F}(\texttt{stop} | \texttt{on\_road})$. Thus, $Q_V(s, \texttt{stop})$ is equal to $-1$. Similarly, \texttt{drive\_at\_70kph} grants a reward of $+80$ (losing some due to the time penalty that it gets by the time it reaches the parking spot). As this speed exceeds the limit, it would also count as a violation of the norm $\textbf{F}(\texttt{speeding})$. On the other hand, both \texttt{drive\_at\_30kph} and \texttt{drive\_at\_50kph} are complying with all the norms. Yet, since driving at $30$ kph requires more time to reach the parking spot than if the agent were driving at $50$ kph, the total expected reward is lower, due to the extra time penalty it would be given.

If we run the lexicographic selection over this set of actions, we would then have the following subsets at each step:

\begin{enumerate}\tightlist
    \item Starting set $A(s)$: $\{\texttt{stop}, \texttt{drive\_at\_30kph}, \texttt{drive\_at\_50kph}, \texttt{drive\_at\_70kph}\}$
    
    \item Filtering suboptimal actions regarding $Q_V$: \{\texttt{drive\_at\_30kph}, \texttt{drive\_at\_50kph}\}

    \item Filtering suboptimal actions regarding $Q_R$: \{\texttt{drive\_at\_50kph}\}
\end{enumerate}

This set only contains one action, \texttt{drive\_at\_50kph}, which is then optimal with respect to the norms and the task of picking up the customer. If the final set was containing more than one action, the agent could select one randomly as they would all be considered equivalent in terms of violations and reward.


\subsection{$\delta$-Lexicographic Selection}
\label{sec:dlex}

One criticism that one may have against the proposed approach is that it is too strict in its respect of the norms. This section then proposes an alternative to the standard lexicographic selection method that may add more flexibility to the system.

Let the following example. An agent has to choose, in a state $s$, between three actions $a$, $b$, and $c$. Their \qvs for $Q_V$ and $Q_R$ are shown in Table~\ref{tab:dlex}. Using the standard lexicographic selection method (referred to as \textsc{lex} in this section), it will choose the action $a$, even though the risk of committing a violation when doing $b$ is very small and provides a greater reward in return. In some contexts of application, one may wish to make its agent more flexible.

\begin{table}[ht]
    \centering
    \caption{Example of \qvs in a state $s$.}
    \begin{tabular}{c|c|c}\hline
        \textbf{Action ($x$)} & $Q_V(s, x)$ & $Q_R(s, x)$\\\hline
        $a$ & $0$ & $10$\\
        $b$ & $-1$ & $20$\\
        $c$ & $-0.2$ & $15$\\\hline
    \end{tabular}
    \label{tab:dlex}
\end{table}

Consequently, we propose a variant of \textsc{lex} introduced earlier. This approach aims at adding a tolerance margin to the selection process of the optimal action(s). This method is named $\delta$-Lexicographic (\textsc{dlex}) selection. It requires to expertly set a value for $\delta \in \mathbb{R^+}$ which will serve as a tolerance margin.\footnote{Note that $\delta = 0$ is the same as applying \textsc{lex}.} Algorithm~\ref{algo:dlex} shows how the optimal set of actions is then computed using this method.\footnote{Note that the application of the margin can be removed for the last \qf.}

\begin{algorithm}
\caption{$\delta$-Lexicographic selection}
\label{algo:dlex}
\begin{algorithmic}[1]
    \Require $s$ (current state), $L$ (a list of Q-functions, ordered according to a preference order), $\delta$ (tolerance margin)
    \Ensure $A^*$ (the resulting set containing the optimal actions)
    \State $A^* \coloneqq A(s)$ \Comment{Gets the set of the possible actions in state $s$}
    \For{$Q_i \in L$}
        \If{$lastElement(Q_i, L)$} \Comment{If last element of the ordering, the tolerance is ignored}
            \State $\delta \coloneqq 0$
        \EndIf
        \State $t \coloneqq \max\limits_{x \in A^*} Q_i(s, x) - \delta$ \Comment{Computes the threshold}
        \State $A^* \coloneqq \{x \in A^*|Q_i(s, x) \geq t\}$ \Comment{Filters out suboptimal actions below the margin $\delta$}
    \EndFor
    \State \Return $A^*$
\end{algorithmic}
\end{algorithm}

Consider the example of Table~\ref{tab:dlex}, with $\delta$ set to $0.3$. Using the $\delta$-lexicographic selection, we compute the threshold for $Q_V$, which is $t = 0 - \delta = -0.3$. After the first iteration, we have $A^* = \{a, c\}$ as both $Q_V(s, a), Q_V(s, c) \geq t$. Then, after the second iteration, $A^* = \{c\}$ since the \qv for this action is the optimal one. It is possible to define different margins $\delta_i$ for each $Q_i \in L$.

One advantage of this method is that it mitigates issues arising from \qvs that have not fully converged---such as near-identical values (\emph{e.g.}, $0.999$ and $0.998$)---which would for instance lead \textsc{lex} to incorrectly discard a nearly equivalent alternative.

Although in the proposed algorithm, we assume $\delta$ to be given, it can possibly be computed dynamically to include, for example, only the actions $x \in A(s)$ for which the given $Q_i(s, x)$ is in a $10\%$ range of the linear interpolation between the maximal and minimal \qvs for $Q_i$. More formally, for a $\rho\%$ tolerance, the value of $t$ in Algorithm~\ref{algo:dlex} would be replaced by

$$t \coloneqq \max_{x \in A^*}Q_i(s, x) - \frac{\rho}{100} \times \left| \max_{x \in A^*}Q_i(s, x) - \min_{x \in A^*}Q_i(s, x) \right|$$

In the example of Table~\ref{tab:dlex}, this would give $t = 0.1$ for $Q_V$ and $t = 19$ for $Q_R$.

\section{Evaluation}
\label{sec:evaluation_ch1}

This section aims at assessing the well functioning of the proposed architecture, as well as the well-behaving of the agent trained through it. In order to do this, we implemented a grid-world environment inspired by the example proposed in Section~\ref{sec:runningexample}.

\subsection{Environments}
\label{sec:env_ch1}

\begin{figure}[ht]
    \centering
    \includegraphics[width=0.5\linewidth]{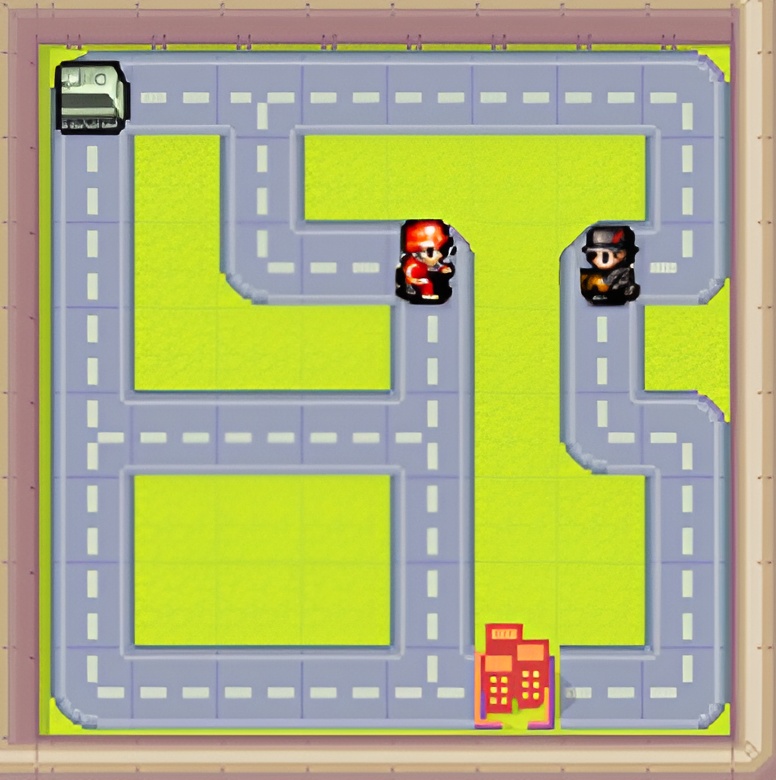}
    \caption{The \texttt{Taxi-A/B/C} environment.}
    \label{fig:env_taxi}
\end{figure}

The \texttt{Taxi-A} environment\footnote{The implementation is available on github: \hyperref{https://github.com/zaap38/Pinocchio-Architecture}.}, shown in Fig.~\ref{fig:env_taxi}, consists of a $10 \times 10$ grid with $3$ different types of tiles, namely, \texttt{road}, \texttt{pavement}, and \texttt{wall}. The agent can move freely onto the \texttt{road} and \texttt{pavement} tiles, but cannot move over the \texttt{wall} tiles. The upper left corner corresponds to the coordinates \xy{0}{0}, and the lower right corner corresponds to \xy{9}{9}. The agent spawns at the coordinates \xy{1}{1}.

The agent's state contains its coordinates, which items are currently on the map, its inventory, and the iteration number divided by $5$ (to give the agent a notion of time without explicitly giving the time step in which it is). Once $60$ iterations have been reached, the environment is reset.

The agent possesses a total of $8$ actions, each being a tuple of two parameters resulting from the combination of a direction $\{\texttt{up}, \texttt{down}, \texttt{right}, \texttt{left}\}$ and a speed $\{\texttt{slow}, \texttt{fast}\}$. For example, the action of the agent moving slowly upward is represented by the tuple $(\texttt{up}, \texttt{slow})$. If the agent chooses a direction that makes it collide with a wall, its position remains unchanged, but it receives a $-10$ penalty on its reward. Furthermore, at each time step, the agent receives a $-1$ penalty to account for the time. This penalty is reduced to $-0.5$ if the agent chooses the action parameter $\texttt{fast}$.

There are three ``items'' dispatched within the environment, which are named \texttt{Red-Passenger}, \texttt{Black-Passenger}, and \texttt{Building}.

When an agent goes over \texttt{Red-Passenger} or \texttt{Black-Passenger}, a ``passenger'' is added to its inventory. Items \texttt{Red-Passenger} and \texttt{Black-Passenger} are then removed from the environment. The difference between \texttt{Red-Passenger} and \texttt{Black-Passenger} is that when reaching \texttt{Black-Passenger}, the agent will be considered as being on a parking spot when collecting the passenger. However, it will receive a reward penalty of $-5$ to account for the extra time that the customer may have needed to move to this location.

If an agent goes on \texttt{Building}, and is having a passenger in its inventory, then the passenger is removed, the agent receives $100$ as a reward, and the environment resets. Otherwise, nothing happens.

Three regulative norms are in application within this environment:
\begin{enumerate}[label=$R_\arabic*$ -]\tightlist
    \item $\textbf{F}(\texttt{pavement})$: It is forbidden to go over the pavement tiles.
    \item $\textbf{F}(\texttt{speeding})$: It is forbidden to exceed the speed limit (\emph{i.e.}, use the parameter ``$fast$'').
    \item $\textbf{F}(\texttt{stop} \mid \texttt{road})$: It is forbidden to stop on the road (\emph{i.e.}, reach items \texttt{Red-Passenger}, \texttt{Black-Passenger}, or \texttt{Building}).
\end{enumerate}

Similarly, two stakeholders are involved, namely, the taxi company and the law enforcement.
The list of all the constitutive norms in application within the environment is:

\begin{itemize}\tightlist
    \item $C_1(\texttt{role(taxi)})$
    \item $C_2(\texttt{morning},\ \texttt{not\_service})$
    \item $C_3(\texttt{night},\ \texttt{not\_service})$
    \item $C_4(\texttt{evening} \land \texttt{has\_passenger},\ \texttt{late})$
    \item $C_5(\texttt{time=0{-}1},\ \texttt{morning})$
    \item $C_6(\texttt{time=2{-}2},\ \texttt{day})$
    \item $C_7(\texttt{time=3{-}7},\ \texttt{evening})$
    \item $C_8(\texttt{time=8{-}11},\ \texttt{night})$
    \item $C_{9}(\texttt{dist\_parking\_{<}\_4},\ \texttt{parking\_near})$
    \item $C_{10}(\texttt{in\_city},\ \texttt{no\_exception})$
\end{itemize}

Note that $C_{1-8}$ belongs to $\sh_{taxi}$ and $C_{9-10}$ belongs to $\sh_{law}$. The argumentation frameworks corresponding to the defeat (or application) of each norm are:

\begin{flalign*}
&\left[
\begin{array}{l}
R_1: \mathbf{F}(\texttt{pavement}) \\
\quad \args = \{\mathbf{F}(\texttt{pavement})\} \\
\quad \defs = \emptyset
\end{array}
\right.&
\end{flalign*}

\begin{flalign*}
&\left[
\begin{array}{l}
R_2: \mathbf{F}(\texttt{speeding}) \\
\quad \args = \{\mathbf{F}(\texttt{speeding}), \texttt{late}, \texttt{no\_exception}\} \\
\quad \defs = \{(\texttt{late}, \mathbf{F}(\texttt{speeding})),
(\texttt{no\_exception}, \texttt{late})\}
\end{array}
\right.&
\end{flalign*}

\begin{flalign*}
&\left[
\begin{array}{l}
R_3: \mathbf{F}(\texttt{stop} \mid \texttt{road}) \\
\quad \args = \{\mathbf{F}(\texttt{stop} \mid \texttt{road}), \texttt{not\_service}, 
\texttt{role(taxi)}, \texttt{parking\_near}\} \\
\quad \defs = \{(\texttt{role(taxi)}, \mathbf{F}(\texttt{stop} \mid \texttt{road})),
(\texttt{not\_service}, \texttt{role(taxi)}),\\
(\texttt{parking\_near}, \texttt{role(taxi)})\}
\end{array}
\right.&
\end{flalign*}

The optimal reward that the agent can achieve without committing any violation is $79$.

A variant of the \texttt{Taxi-A} environment, named \texttt{Taxi-B}, removes the time penalty occurring at each time step when the agent is not carrying a passenger. In order to help the agent to still converge to the desired behaviour, an intermediary reward of $50$ is added when picking up a passenger. This reward is set to $45$ when picking up \texttt{Black-Passenger} instead to account for the $-5$ penalty. In this variant, the optimal reward is $140.5$.

Then, in a third variant called \texttt{Taxi-C}, the status of the norm $R_2$ (speeding) can be randomly altered, with a probability of $10\%$, such that the norm is activated even if the condition for its defeat were reunited. This last environment serves at testing the human intervention feature of the architecture.

\begin{figure}
    \centering
    \includegraphics[width=0.5\linewidth]{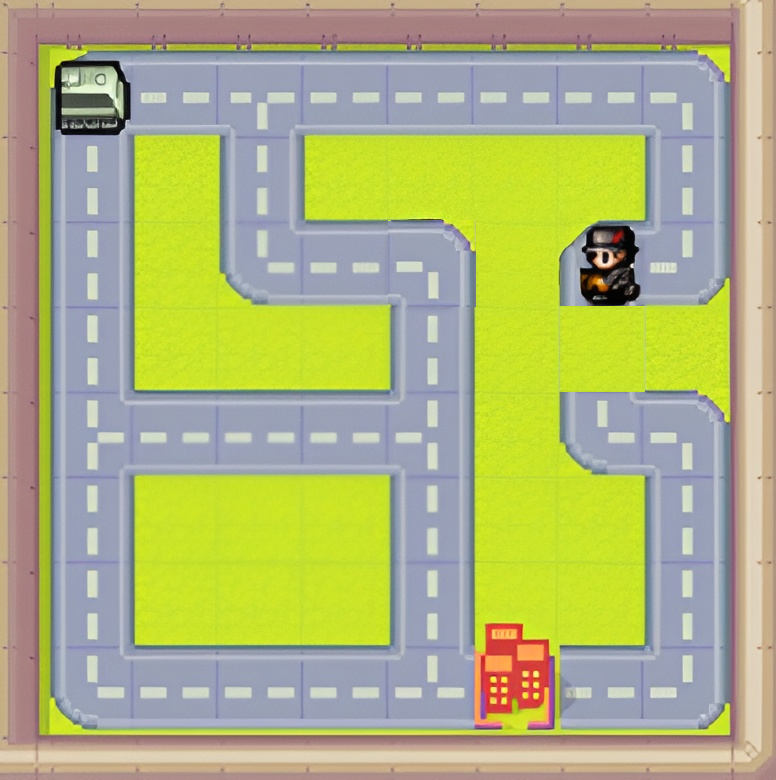}
    \caption{The \texttt{Taxi-D} environment.}
    \label{fig:env_blocked}
\end{figure}

Finally, the \texttt{Taxi-D} environment will aim at highlighting the difference of behaviour between the different agents tested. Fig.~\ref{fig:env_blocked} shows the new grid corresponding to this environment. We can see that the agent would greatly benefit from cutting by the small pavement tile at \xy{7}{4}.

\subsection{Results}

This section evaluates the proposed architecture by comparing $2$ agent configurations in $4$ different variants of the \texttt{Taxi-A} environment.

\subsubsection{Tested Agents}
\label{sec:tested}

We choose to compare $2$ different variants of agents in order to evaluate the proposed architecture. These variants are listed below. Note that they all receive the same representation of the state, as well as the same reward signal.

\begin{itemize}
    \item \textbf{\pino} $\times$ \textsc{lex} \textbf{(RL-Lex):} This agent uses the \pino architecture initialised with the two stakeholders and three norms introduced in Section~\ref{sec:env_ch1}. The selection of the best action is done using \textsc{lex}.
    
    \item \textbf{\pino} $\times$ \textsc{dlex} \textbf{(RL-DLex):} This agent is similar to the previous, except that it uses \textsc{dlex} to select its action.
\end{itemize}

Each of these variants shares the same learning rate $\alpha = 0.05$ and discount factor $\gamma = 0.99$.

\subsubsection{Experimental Results}

The experiments consist of $2,000,000$ learning steps in the \texttt{Taxi-A} and \texttt{Taxi-B} environments. The learning parameters were $\alpha = 0.05$ and the discount factor $\gamma = 0.99$. The learning phase takes roughly $10$ minutes with a Python3 implementation and an 11th Gen Intel Core i5-1145 2.60GHz. During the learning phase, the agent had a permanent minimum probability of $10\%$ to select a random action for the sake of exploration. The final behaviour was then assessed by observing a separate run in which the agent was following its optimal policy.

\begin{figure}[h!]
    \centering
    \begin{minipage}{0.49\linewidth}
        \centering
        \includegraphics[width=\linewidth]{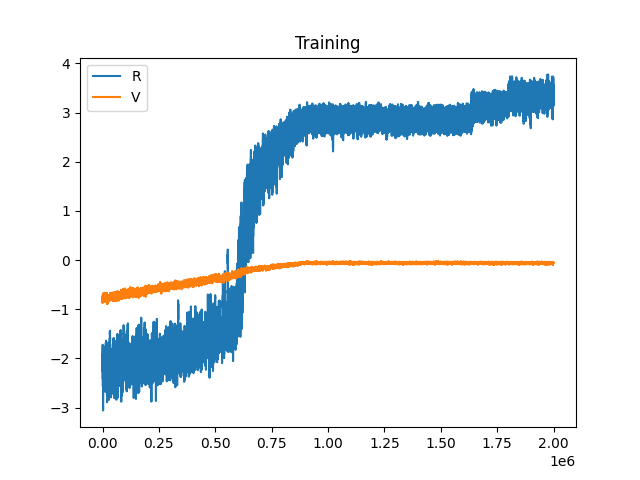}
        \caption{Evolution of the reward and violation count during the training phase in the \textbf{RL-Lex} in \texttt{Taxi-A} environment.}
        \label{fig:learning_taxi}
    \end{minipage}\hfill
    \begin{minipage}{0.49\linewidth}
        \centering
        \includegraphics[width=\linewidth]{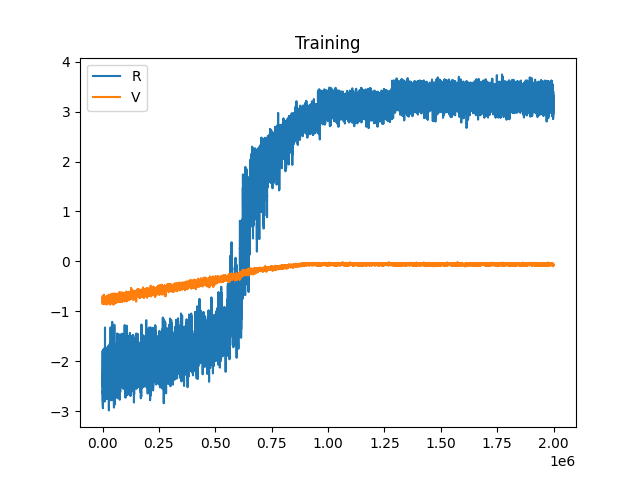}
        \caption{Evolution of the reward and violation count during the training phase in the \textbf{RL-DLex} in \texttt{Taxi-A} environment.}
        \label{fig:learning_taxi_d_a}
    \end{minipage}
\end{figure}

Fig.~\ref{fig:learning_taxi}--\ref{fig:learning_taxi_d_a} shows the evolution of the reward and violation count during the learning phase for the tested agents in \texttt{Taxi-A} environment. We can see that the agent quickly learns to avoid committing violations, while also managing to obtain a near optimal reward. The reward that it gets when performing its optimal policy is $77$. The $2$ points below the maximal theoretical reward are due to the agent being unable to know through its state representation when the day time is about to change from ``morning'' to ``day'', and thus does a back step instead of picking up its passenger.

\begin{figure}[h!]
    \centering
    \begin{minipage}{0.49\linewidth}
        \centering
        \includegraphics[width=\linewidth]{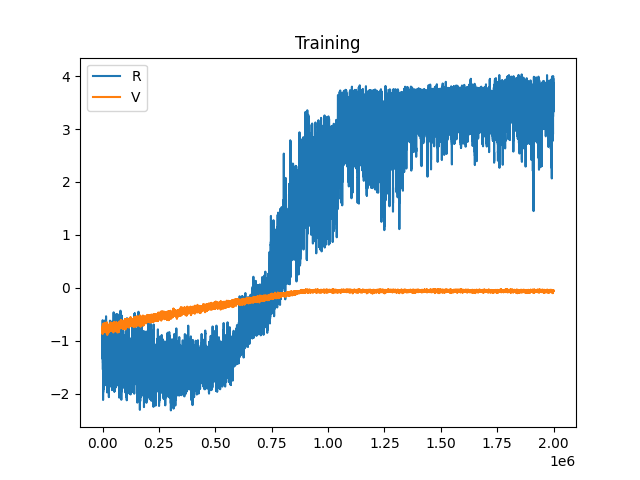}
        \caption{Evolution of the reward and violation count during the training phase in the \textbf{RL-Lex} in \texttt{Taxi-B} environment.}
        \label{fig:learning_taxi_b}
    \end{minipage}\hfill
    \begin{minipage}{0.49\linewidth}
        \centering
        \includegraphics[width=\linewidth]{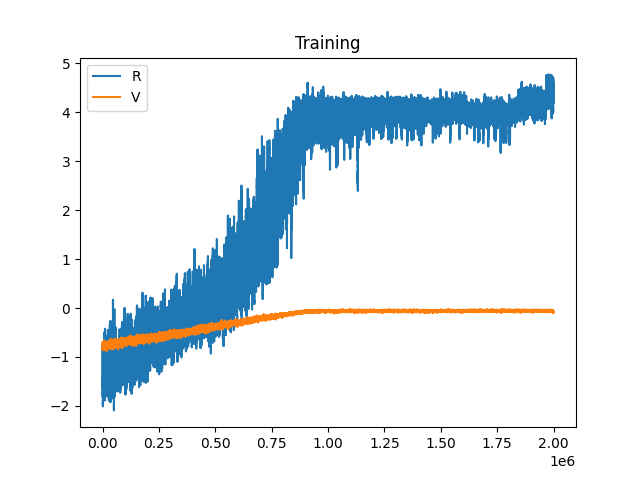}
        \caption{Evolution of the reward and violation count during the training phase in the \textbf{RL-DLex} in \texttt{Taxi-B} environment.}
        \label{fig:learning_taxi_d_b}
    \end{minipage}
\end{figure}

Fig.~\ref{fig:learning_taxi_b}--\ref{fig:learning_taxi_d_b} shows the same information for the variant \texttt{Taxi-B}. This time, the agent achieves the optimal reward of $140.5$ when following its optimal policy, while still avoiding to commit any violation. During the beginning of the learning phase, we can see that while the number of violations reduces, the average reward goes down. This is due to the fact that the agent discovers that picking \texttt{Red-Passenger} grants a bonus reward. However, as it also learns to avoid violations, it learns to avoid picking \texttt{Red-Passenger}. It then requires some more learning steps before finding \texttt{Black-Passenger}. After this, the reward goes up. Another interesting feature of the optimal behaviour is that instead of rushing for the passenger, the agent waits for the day time to be ``evening'' as it allows it to use the ``fast'' action parameter. This phenomenon is called \textit{Norm Avoidance}~\cite{alcaraz2025norm}. As it is not the focus of this chapter, no attempt has been made to mitigate it. However, approaches to mitigate it will be discussed later in this thesis.

\begin{figure}[h!]
    \centering
    \begin{minipage}{0.49\linewidth}
        \centering
        \includegraphics[width=\linewidth]{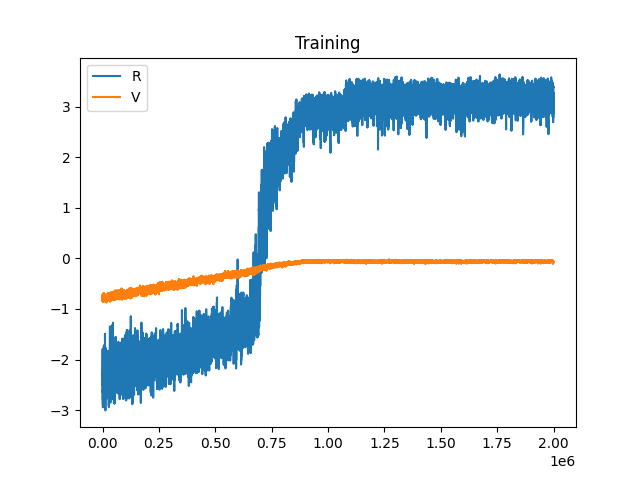}
        \caption{Evolution of the reward and violation count during the training phase in the \textbf{RL-Lex} in \texttt{Taxi-C} environment.}
        \label{fig:learning_taxi_c}
    \end{minipage}\hfill
    \begin{minipage}{0.49\linewidth}
        \centering
        \includegraphics[width=\linewidth]{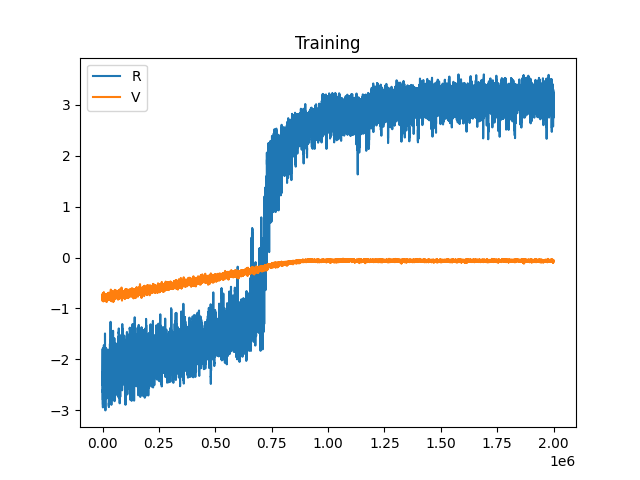}
        \caption{Evolution of the reward and violation count during the training phase in the \textbf{RL-DLex} in \texttt{Taxi-C} environment.}
        \label{fig:learning_taxi_d_c}
    \end{minipage}
\end{figure}

Then, Fig.~\ref{fig:learning_taxi_c}--\ref{fig:learning_taxi_d_c} shows, for the tested agents in the \texttt{Taxi-C} environment, the evolution of the reward and violation count when the agent is learning with random human interventions that may force it to respect the speed limit, disregarding the defeat conditions. Through observations of the agent trace, we could see that the agent was able to learn an optimal behaviour while still complying with the norms, in particular when the norm $\mathbf{F}(\texttt{speeding})$ is manually altered.

Finally, we compare the two agents in the \texttt{Taxi-D} environment. The \textbf{RL-DLex} agent was set with a fixed tolerance of $1$, which means that it could tolerate one violation. As expected, \textbf{RL-Lex} stayed on the pavement, while \textbf{RL-DLex} cut by the pavement, thus achieving a greater reward at the cost of one violation.

In conclusion, we can see that the proposed architecture allows to derive from the state of the environment more complex facts that can serve at determining whether a norm has to be complied with, with respect to the arguments of multiple stakeholders. We can also see that the agent is then able to learn a behaviour compliant with the norms while still achieving an optimal behaviour. Then, we assessed the capacity of the agent to comply with manual updates of the norm status. Finally, we have shown that the \textbf{RL-DLex} variant was able to achieve a greater reward by exploiting a tolerance margin.

\begin{remark}
    As the computation of the grounded extension may raise concerns about the time complexity when dealing with bigger graphs, we would like to provide the following insights: With Python3 (i5-1145 2.60GHz, 16Go RAM) and a graph of 2K nodes and 13K edges, computing the grounded extension takes in average 6 ms.
\end{remark}

\section{Related Work}

This section describes works related to the architecture proposed in this chapter. It first explores the work focusing on normative reinforcement learning. The discussion is then expanded by exploring works that make use of normative supervisors, and finally, to approaches related to normative agents in general.

\subsection{Normative Reinforcement Learning}
\label{sec:normativerl}

When put in perspective with the RL-related and Machine Ethics literature, not so many works have been done in the field of Normative RL. Furthermore, these works are fairly recent when compared to the first discussions around normative agents.

\citet{makarova2025deontically} introduce a method in which an RL agent learns a policy (achieving at least local optimality) to accomplish a ``mission'', \emph{i.e.}, reaching a state corresponding to the completion of the task of the agent, while avoiding states that would not respect the norms. In this work, the agent learns within an MDP (see Definition~\ref{def:mdp}) the probabilities associated with the transitions state-action-state. What the agent learnt can then be transformed into a Probabilistic Computation Tree Logic (PCTL) by unrolling the MDP and associating to each transition the probability observed during the learning. From each state (\emph{i.e.}, node of the tree) can be derived truth values for a set of STIT (see to it that) logic formulae. Some of these formulae describe norms, and consequently states to avoid or reach depending on the norm being a prohibition or an obligation. Some other formulae describe state-of-affairs in a given state, allowing for the assessment of the completion of the mission. This essentially transforming their MDP into a labelled MDP (see Definition~\ref{def:lmdp}). Then, these values are propagated back through the nodes of the tree if the probability of the transition stays above a fixed threshold\footnote{\citet{makarova2025deontically} claim that while one can see such threshold as being arbitrary, in computer engineering such a value would be derived from empirical studies and risk assessments.}, similarly to a constraint satisfaction problem.

The authors then compute from the obtained tree a policy that ensures the achievement of the mission while avoiding violating states. Although this approach provides many formal guarantees about the policy that the agent will learn, such as local optimality (at least), avoidance of the violating states, or achievement of the task. Also, it makes very clear why the agent favoured one action over another, granting it a good explainability potential. However, it also suffers from some strong assumptions that limit its applicability. For example, the size of a real-world environment often makes it unrealistic to assume the memorisation of the state resulting from another state-action pair to be possible, as the state space may be too big to be memorised with reasonable resources.\footnote{This issue is common among the model-based approaches, \emph{i.e.}, the reinforcement learning approaches trying to learn the transition probabilities from one state to another (in opposition to the model-free approaches that focuses on the learning of the expected reward) such as safe RL techniques using Linear Temporal Logic (LTL) to identify a safe policy~\cite{alshiekh2018safe,jansen2018safe,neufeld2022enforcing}. This renders model-based approaches hardly scalable.} Secondly, it is stated that this approach does not consider environments exhibiting normative conflicts (that here would take the form of the non-existence of a path satisfying the constraints), where satisfying one norm makes at least one other norm violated. This, once again, limits the applicability of the aforementioned approach.

In their work, \citet{scheutz2025using} propose to use Inverse Reinforcement Learning (IRL) to learn what action the agent ought to do from a dataset of observed behaviours. This allows them to derive what they call low-level obligations. Then, by extending their MDP to a labelled MDP, they can extract higher-level obligations. The main advantage of this method is that we are certain that there cannot be a mistake in the design of the norms, as there is no such phase. However, this work suffers from several problems. First, it assumes that the training data does not contain observation taken from malicious agents potentially disrespecting the norms. This is a very strong assumption, especially given that these data may never be humanly examined. Second, their approach does not distinguish between the respect of an actual norm, and simply a convenient action for an agent. For example, if an agent takes an item in a store, it is obligated that it pays. However, when an agent takes the elevator to go to the tenth floor, even though there were stairs, this is not what one would like to qualify as a norm.\footnote{If we do consider this as a norm, we may have to consider any action that is suboptimal in regard to the reward as prohibited.} Last, it requires the environment to possess symbols in order to extract norms which are meaningful to a human. If this is not the case, then most of the generated norms will look like ``If you visited state $x$, you ought to visit state $y$''. This, combined with the multitude of ``false positive'' norms which may be extracted, may render the identification and understanding of the actual norms challenging.

Finally, we would like to introduce the work of Neufeld~\cite{neufeld2022reinforcement,neufeld2023norm,neufeld2024learning}, where a model-free normative RL architecture has been developed to train an agent to follow the norms provided expertly through a knowledge base. Similarly to the ones presented above, this work uses a labelled MDP to represent the state-of-affairs in each state. Labels are used to detect whenever a norm is fulfilled or violated. For instance, let an environment where the constitutive norm $\textbf{C}(red\_fence,\neg white\_fence)$---which essentially means that a red fence is not white---and the regulative norm \Ob{$white\_fence|fence$}---which means that if there is a fence, it should be white---are in application.\footnote{This example is inspired by the Dutch cottage housing regulation as presented by \citet{prakken1996contrary}.} A state having the labels $\{fence, red\_fence\}$ would then trigger a violation. Instead of updating a single \qf, the agent updates one for the expected reward ($Q_R$), and one for the expected violations ($Q_V$). When in a state $s$, the agent knows that the expected reward it will receive when doing the action $a$ is $Q_R(s, a)$, and the expected number of violations is $Q_V(s, a)$.\footnote{A violation is usually expressed with a negative value. This means that $Q(s, a) = 0$ denotes no violation, while $Q(s, a) = -n$ denotes $n$ violations.} Then, when running its optimal policy, the agent chooses the action to take by performing a thresholded lexicographic selection over the possible actions. A lexicographic selection~\cite{gabor1998multi} consists of iterative selection of subsets of actions such that they are optimal with respect to a \qf. The order in which this selection is performed, called a lexicographic ordering, is fixed. In Neufeld's work, it is $\mathcal{P} = \{V \succ R\}$, meaning that the first subset will contain the actions that maximise the violation \qv (as maximising this value is the same as minimising the number of violations committed), and the second subset the actions maximising the violation \qv and maximising the reward. A thresholded lexicographic selection~\cite{vamplew2011empirical} extends this process by adding a minimal threshold value $C_i$ for each element $i$ in the lexicographic ordering. Instead of selecting only the optimal subset at each iteration, it selects all the actions that return a value above the corresponding threshold. If no action reaches the threshold, the most optimal is chosen. The threshold for the last value of the ordering is usually set to an infinite value. The process of selecting an optimal action given a set of \qfs is better shown in Algorithm~\ref{algo:tlex}.

\begin{algorithm}
\caption{Thresholded lexicographic selection}
\label{algo:tlex}
\begin{algorithmic}[1]
    \Require $s$ (current state), $L$ (a list of \qfs, ordered by priority), $C$ a vector of thresholds for the \qfs in $L$
    \Ensure $A^*$ (the resulting set containing the optimal actions)
    \State $A^* \coloneqq A(s)$ \Comment{Gets the set of the possible actions in state $s$}
    \For{$Q_i \in L$}
        \State $T \coloneqq \{x \in A^* | Q_i(s, x) \geq C_i\}$ \Comment{Filters out actions below the threshold}
        \If{$T = \emptyset$}
            \State $A^* \coloneqq \arg\max\limits_{x \in A^*} Q_i(s, x)$ \Comment{Filters out suboptimal actions}
        \Else
            \State $A^* \coloneqq T$
        \EndIf
    \EndFor
    \State \Return $A^*$
\end{algorithmic}
\end{algorithm}

If we restrict Neufeld's approach to a standard lexicographic selection and disregard the discount factor $\gamma$, it converges to an optimum that minimises the violations and maximises the reward among the remaining options. Furthermore, it does not require more expert knowledge than a standard RL environment (the only exception being the labels that help to know if a norm is violated or not). Since the norms are given explicitly in a knowledge base, and expressed with symbols, it is fairly easy for a human to understand what the agent tries to respect as a norm. We believe the version using the thresholded lexicographic selection to be more controversial, as the choice of the threshold value may be somewhat arbitrary. Another criticism one can make is that the management of the norms and the conflict remains fairly minimalist. The author addressed this issue in further works by proposing a theorem prover-based normative supervisor~\cite{neufeld2022reinforcement,neufeld2024learning}. However, since we plan to integrate a normative supervisor, this thesis will be based on the version previously described.

As this last work provides a reliable and flexible basis, this thesis will mostly use it as a base for the normative RL side.

\subsection{Normative Supervisors}
\label{sec:supervisor}

As reinforcement learning is not the only way to make an agent act normatively, here are some works that can be classified as normative supervisors, \emph{i.e.}, entities that govern an agent by deciding whether it has the right to take such action or not.

An example of normative supervisor is the Jiminy architecture~\cite{liao2019building,liao2023jiminy} previously introduced. Its integration to an agent was out of the scope of the papers where it was proposed, Consequently, several limitations arises when trying to integrate it to a reinforcement learning agent. This makes it inapplicable in its current state.

First, it is designed to be used in real time during the agent's deployment, as there is no learning phase involved. This means that the heavy computations will have to occur in real time. On onboard systems, it might be problematic to permanently allocate that much resources. To alleviate this issue, the Jiminy architecture proposes to skip the reasoning process if no normative conflict, or norm-relevant element, is detected in the current state, but it is unclear how this detection would be achieved without actually entering the reasoning process.
Second, it does not guarantee any privacy for the stakeholders. Indeed, as their model has to be stored within the system, potentially anyone can have a look inside to see the elements constituting each avatar. A subsequent problem is that anyone can potentially modify these models so that the supervised agent deviates from the originally intended behaviour.
Third, it lacks the ability to deal with uncertainty. Indeed, as Jiminy intervenes before the agent to select a set of actions that are considered to be adhering to the norms, it cannot learn from the consequences of the actions performed. Therefore, the agent may adopt a behaviour in which it does what seems to be good (with respect to the norms) rather than what is truly good. This incapacity to deal with uncertainty is also problematic for avoiding long-term violations. Indeed, sometimes it is preferable to commit a violation immediately, as not doing it may later lead to the unavoidable commitment of a worse violation.

\begin{figure}
    \centering
    \includegraphics[width=1.0\linewidth]{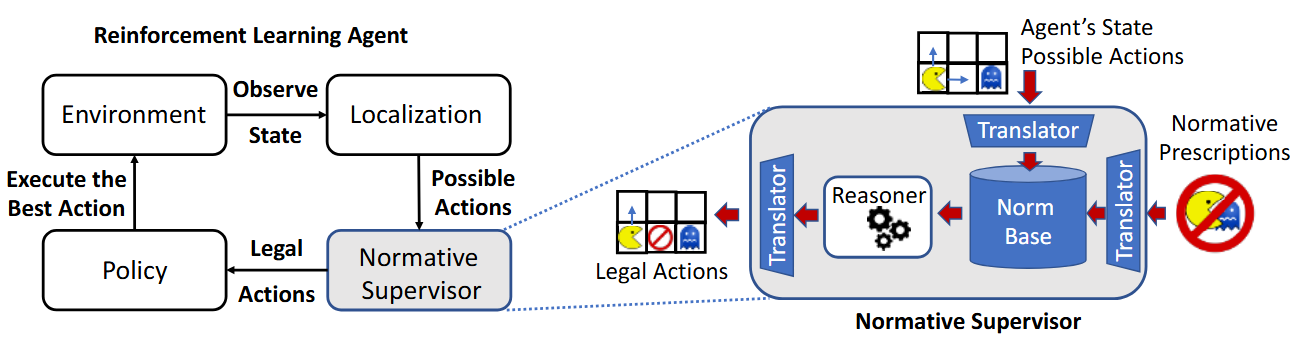}
    \caption{Normative architecture proposed by \citet{neufeld2021normative}.}
    \label{fig:neufeld}
\end{figure}

In the work of \citet{neufeld2021normative}, an architecture for normative supervision is proposed. It takes the form of a shielding architecture, shown in Fig.~\ref{fig:neufeld}, where a normative supervisor first receives the observations of the agent, and then computes a set of legal actions from a set of possible actions. These legal actions are then forwarded to the reinforcement learning agent, who can choose only between these.

It suffers from similar issues as Jiminy. The first is the inability to deal with stochasticity. Although the RL-agent is capable of optimising its reward (for the main task), it is dependent on the choice of the normative supervisor to choose a legal action. This means that depending on how the norms are structured within the normative supervisor, the agent may choose an action that is legal, but that will later lead to an unavoidable violation. The second issue is that, if no legal action is available, the agent will have to choose one illegal action. As the agent did not train at predicting the normative consequences of such illegal actions (as it tracks the reward, like a standard RL-agent, but not the expected future violations, neither preferences nor weights over them), it may not achieve optimality in case of a normative conflict.

\subsection{Other Approaches to Normative Agency}
\label{sec:normative}

This section discusses other approaches related to normative agent architectures that are not directly related to reinforcement learning. This includes approaches based on planning or BDI (Beliefs, Desires, Intentions) models~\cite{georgeff1991modeling}.

\citet{castelfranchi1999deliberative} introduces deliberative normative agents that are aware of the norms in a multi-agent environment and can choose whether to violate them in specific contexts. A proposed architecture for such agents allows them to recognise norms, adopt them, or deliberately violate them when appropriate. This framework refines earlier models by adding components for reasoning in norm-governed settings. Unlike earlier hard-coded approaches used in social simulations, it enables agents to adapt their behaviour over time.

In \citet{boman1999norms}, a method is proposed to enforce norms on supersoft agents, which process vague and imprecise information using a decision module. Norm compliance is achieved by manipulating utilities, removing harmful actions, or disqualifying behaviours that reduce global utility in the system. This approach does not define a full agent model, but rather enforces rules on utility-driven agents whose decisions are always obeyed. Although effective in ensuring compliance, it lacks flexibility as it prevents norm violation and sanctioning, limiting its usefulness in domains where contracts can be broken and outcomes are uncertain.

Then, in \citet{dignum2000towards}, an approach is proposed that integrates norms and obligations into the BDI model to support socially motivated reasoning. Agents explicitly represent norms, allowing them to decide whether to comply and adapt when norms become obsolete or differ between agents. The control algorithm extends the BDI process by including deontic events in the option generator, producing sets of compatible plans. Plan selection is guided by social benefits and severity of punishment, enabling agents to adopt norms for collaborative behaviour. However, the model only addresses conflicts between norms, resolving them through predefined preference orderings rather than dynamic utilities.

The Normative Agent Architecture (NoA) presented in \citet{kollingbaum2003noa} extends the BDI model with mechanisms to reason about norms. It consists of a specification language for beliefs, goals, plans, and norms, and an interpreter that executes them. Rather than excluding forbidden actions, NoA labels them as such, leaving agents norm-autonomous by allowing deliberate violations when conflicts arise. NoA uses predefined plans of deterministic actions. It does not model punishments for violations. Instead, agents aim to maximise norm respect, with violations occurring only when compliance is impossible.

In \citet{boella2006game}, a formal game-theoretic model is proposed for agents negotiating contracts, introducing a qualitative approach based on recursive modelling. Agents predict whether their actions violate norms and may be sanctioned. Conflicts among motivations are resolved through priority relations. The model assumes deterministic transitions.

\citet{andrighetto2007immergence} presents the EMIL-A architecture designed to handle the emergence of norms by recognising new norms, forming normative beliefs and goals, and generating intentions and plans to act accordingly. It includes a long-term memory of norms and a repertoire of compliant actions, with decisions guided by the salience of norms. Agents compare expected utility when deciding whether to comply, considering possible sanctions and incentives. However, the model lacks an explicit representation of sanctions and does not detail the normative action planner.

\cite{lopez2006normative} proposed a normative framework for agent-based systems, combining models of norms, normative multi-agent systems, and normative autonomous agents. Agents decide whether to comply with norms by evaluating goals that may be hindered or rewarded, and consider potential punishments when rejecting norms. The architecture incorporates agent profiles---such as social, rebellious, and opportunistic---that guide goal selection through priority functions.

\citet{meneguzzi2009norm} extend BDI languages to allow behaviour modification at run-time in response to newly accepted norms. Norms are represented with activation and expiration conditions and can be prohibitions or obligations referring to states or actions. The interpreter scans the agent's plan library to remove violating plans for prohibitions or generate new plans for obligations. Although the method focuses on adapting behaviour to accepted norms, enabling norm violation would require a decision process for accepting or rejecting norms, which restricts the agent's autonomy.

In \citet{cardoso2009adaptive} designs norm-aware utility-oriented agents that adjust sanctions in contracts. Agents differ in risk tolerance, affecting decisions to enter contracts, and social awareness, influencing compliance with obligations even when not personally advantageous. These agents do not construct plans, but instead compute expected utilities.

Finally, \citet{joseph2010deductive} proposes a coherence-driven agent architecture that extends the BDI model with a theory of deductive coherence, guiding decisions based on coherence maximisation rather than expected utility. Norms are represented with graded priorities, and norm violations are linked to sanctions using logical implications. Non-deterministic information is handled with grades, and sanctions are encoded as beliefs. Agents adopt norms and sign contracts that are coherent with their current mental state, rather than those that are more profitable. The architecture does not specify a planning context.

\section{Summary}

This chapter presents the \pino architecture~\cite{alcaraz2026pinocchio} and its subcomponents---which is a combination of the Jiminy architecture, the AJAR framework, and Neufeld's NGRL agent---by providing its formal model and explaining how it integrates within a reinforcement learning setup. In the proposed architecture, a reinforcement learning obtain, in addition to its reward signal, a violation signal. The latter comes from stakeholders' representations, called avatars, that debate whether the agent committed a violation for a given norm through the use of formal argumentation. Then, we proposed an alternative to the standard lexicographic selection method that gives more flexibility and freedom to the agent. We have seen through an empirical evaluation that the proposed architecture not only makes the agent able to follow the norms, but also allows for some flexibility by following dynamic norm revisions from a human operator or making compromises to achieve a better reward. Then, it compares the benefits of the proposed approach with other similar works in the literature related to normative reinforcement learning and normative supervisors by outlining their limitations and detailing how \pino addresses them.

However, this architecture can be further improved. In consequence, here are some challenges and research directions that one may find relevant to explore. First, the way in which the stakeholders' arguments are exchanged can be improved. In the proposed architecture, the method we use is pretty naive. However, there exist several works within the literature that allow agents to have strategic argumentative discussions. Using these works may allow stakeholders to regain agency. It would also improve the overall fairness since there would be more control on how the arguments would be put forward within the discussion. Second, a more complex structure than Dung's argumentation could be used. Indeed, decisions could be modelled using, for example, probabilistic argumentation frameworks developed specifically for reinforcement learning~\cite{riveret2019probabilistic}, or representations more philosophically grounded~\cite{alcaraz2024estimating}.

%% file: sections/3_dynamically_modelling_norms.tex
The goal of this chapter is to provide the necessary tools to reduce the burden of the designer when using the \pino architecture presented in Chapter~\ref{cha:pino}. Consequently, this chapter reviews the norm identification techniques from the literature to help the reader in selecting one. Then, it proposes ARIA, an argumentative rule induction algorithm that aims at extracting an argumentation that represents the model that governs the decisions of an agent.

\section{Introduction}

As already mentioned in the introduction of this thesis, the \pino architecture requires the norms of the stakeholders to be represented as argumentation graphs. Although a stakeholder such as the manufacturer may be able to allocate resources to model such a structure, this is not always feasible. For example, user stakeholders may lack the time and expertise to develop such a structure. Even if they are able to, insufficient supervision can lead them to overlook critical aspects that, if neglected, may be detrimental. For this reason, it is important to provide a tool to automatically extract the model of some of the stakeholders. Fortunately, all potential ``user'' stakeholders tend to share similar patterns among them. For this reason, it is possible to use an algorithm that would extract, from previously collected behavioural data, a model of these stakeholders.

Doing this requires a two-step process. First, it is necessary to extract the norms that are relevant to the user. Second, the conditions under which the user finds a norm relevant have to be refined and structured as an argumentation graph. In the literature, there exists a broad range of approaches for norm identification, also referred to as norm detection, or norm mining. For this reason, the choice of which algorithm to use to extract the norms is left to the designer, the ARIA algorithm proposed in Section~\ref{sec:aria} focusing solely on the extraction of the exceptions to a norm. Nevertheless, as the designer requires a clear view on literature on norm mining in order to make a sensible choice, Section~\ref{sec:normmining} presents the results of a comprehensive review of the field of norm mining, as well as key takeaways for the choice of the norm mining technique used for \pino. An interested reader may also consider reading the complete review by \citet{alcaraz2026mining} as it identifies the challenges and research directions for the future in the field of norm identification. However, within this thesis, the content will be limited to the analysis of the current approaches.

The second step consists of refining the defeat condition of these norms and representing this condition through an argumentation graph. There can be several ways to organise the arguments in favour or against a norm, such as using Large Language Models (LLMs) (\emph{e.g.}, \citet{alcaraz2024assessing}). In Section~\ref{sec:aria} of this thesis, we introduce the algorithm ARIA (\emph{i.e.}, Argumentative Rule Induction A-star). By learning over a dataset, this algorithm generates a graph that serves at predicting a label depending on an input state. Then, using some techniques from the literature, it can serve at generating post-hoc explanations of a decision.

\section{Review of the Norm Mining Techniques}
\label{sec:normmining}

Norm Mining, also referred to as Norm Identification or Norm Detection, aims to enable autonomous agents to recognise both explicit norms, such as a law, and implicit norms, such as not being noisy in public transportations, in order to reason about their applicability, and adjust their behaviour accordingly. Usually, the main goal of these approaches is to translate the extracted norms into a form matching the Deontic Logic formalism.

This section provides an overview of the techniques developed in the last $15$ years that can help the user of \pino make an informed choice about the norm mining method.

\subsection{Research Questions}
\label{sec:slr_rq}

To help identifying what method is and is not doing, several key research questions have been formulated. These questions aim to address critical aspects of normative systems, ranging from detection and identification to synthesis and adaptation within multi-agent systems (MAS). This section elaborates on these research questions, providing hints on their significance and the assumptions underlying them.

\begin{enumerate}[start=1, label={-- \bfseries RQ\arabic*:}, leftmargin=*]\tightlist
    \item How to identify the norms of a society by observing it? -- This question focuses on passive observation techniques that allow agents to detect prevailing norms without direct interaction.
    
    \item How to identify the norms of a society by interacting with it? -- While passive observation provides valuable insights, active interaction offers additional dimensions for norm identification.
    
    \item How to differentiate individual norms from societal norms in a multi-agent society? -- Differentiating between individual and societal norms is crucial for understanding the emergence and enforcement of normative behaviour. Sub-questions \textbf{RQ3.1} and \textbf{RQ3.2} further explore the feasibility of distinguishing personal norms ($p$-norms) from group norms ($g$-norms) and the methodological challenges involved.
    
    \begin{enumerate}[start=1, label={-- \bfseries RQ3.\arabic*:}, leftmargin=*]
        \item Is it possible to differentiate a $p$-norm from a $g$-norm?
        
        \item How to perform this differentiation with only one agent?
    \end{enumerate}
    
    \item How to detect prohibition norms without norm enforcement? -- Detecting prohibition norms in the absence of explicit enforcement mechanisms is a challenging task.
    
    \item How to detect norms in communications? -- Communication analysis offers a rich source of normative information. Natural language processing (NLP) techniques play a crucial role in extracting norms from textual interactions.
    
    \item Is it possible to adapt to drifting norms without restarting the whole learning phase? -- Adapting to norm drift without reinitialising the learning process is essential for maintaining normative coherence in evolving societies. Norm evolution presents significant challenges for adaptive agents.
    
    \item Is it possible to detect sub-communities of agents? -- The emergence of sub-communities within a multi-agent society can indicate norm fragmentation.
\end{enumerate}

\subsection{Analysis of the Research Questions}
\label{sec:analysis_ch2}

This section identifies among the methods in the literature the main trends. Then, it attempts to answer the research questions listed in Section~\ref{sec:slr_rq}.

This section first describes quantitatively how the research questions were answered by the identified approaches. Then, it goes more in depth into each research question, analysing qualitatively how they address it.

\subsubsection{Quantitative Analysis}

Although each approach addressed the challenges raised by the research questions in its own way, we could identify some major trends in the employed methods. Below are listed the keywords associated with each trend, as well as their description:

\begin{itemize}\tightlist
    \item \textbf{Threshold:} The approach observes a community of agents. If a behaviour is repeated a certain number of times exceeding a threshold value, it is then added to the potential norms.
    
    \item \textbf{Comparison:} The method detects the norms by exchanging information and comparing its set of beliefs or desires with other agents from the environment, or with an external source.
    
    \item \textbf{Reasoning:} The agent uses a reasoning mechanism, or a mathematical formula, to derive the norms from the data.
    
    \item \textbf{Elitism:} The approach focuses on the observation of a limited number of agents, usually having a higher trust value. Those agents can also act as helpers toward other agents to help them in identifying the system's norms.
    
    \item \textbf{Log:} The approach makes use of the trace of other agents, potentially tracking the signals such as sanctions.
    
    \item \textbf{Data Mining:} The approach uses pattern recognition techniques, or machine learning techniques, to extract the norms.
    
    \item \textbf{Natural Language Processing (NLP):} The method uses grammar and semantics to detect the norms.
    
    \item \textbf{Yes:} The paper answered positively to a closed research question.
    
    \item \textbf{Not Answered:} The paper did not address the given research question. For a closed question, it is not necessarily equivalent to a negative answer.
\end{itemize}

\begin{figure}
    \centering
    \includegraphics[width=0.9\linewidth]{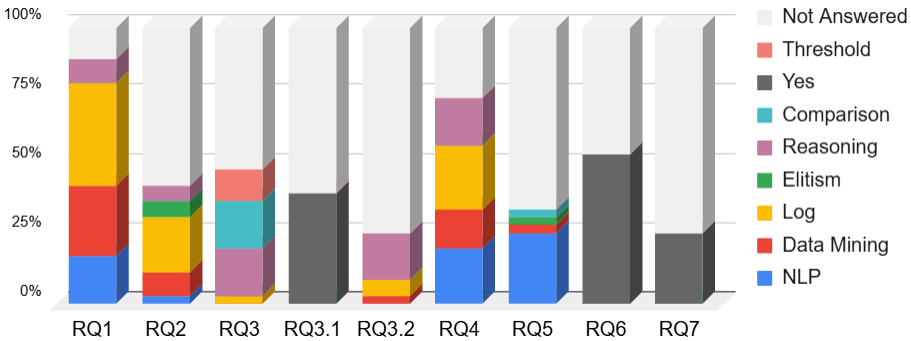}
    \caption{How each research question is addressed by each approach.}
    \label{fig:topic}
\end{figure}

Fig.~\ref{fig:topic} shows, for each research question, what are the major trends among the approaches answering it, as well as the proportion of approaches that do not address this question. Table~\ref{tab:rq_detailed} provides a detailed view of each of the techniques.

\begin{sidewaystable} 
    \centering
    \scriptsize
    \caption{Research Questions Breakdown}
    \begin{tabular}{|c|l|l|l|l|l|l|l|l|l|l|}\hline
            \textbf{N°} & \textbf{Reference} & \textbf{RQ1} & \textbf{RQ2} & 
            \textbf{RQ3} & \textbf{RQ3.1} & \textbf{RQ3.2} & \textbf{RQ4} &
            \textbf{RQ5} & \textbf{RQ6} & \textbf{RQ7} \\\hline
            1 & \cite{ferraro2021nlp} & NLP & NLP & & & & NLP & NLP & & \\\hline
            2 & \cite{mahmoud2012norms} & Log & Log & Threshold & Yes & Data Mining & Log & & Yes & \\\hline
            3 & \cite{mahmoud2016development} & Data Mining & & Comparison & & & & & & \\\hline
            4 & \cite{mahmoud2012norm} & Log & Elitism & Comparison & Yes & & Log & & Yes & Yes \\\hline
            5 & \cite{savarimuthu2010norm} & Log & Log & Reasoning & Yes & Reasoning & & & Yes & Yes \\\hline
            6 & \cite{mahmoud2016norm} & Reasoning & & Reasoning & Yes & Reasoning & Reasoning & & Yes & Yes \\\hline
            7 & \cite{mahmoud2013potential} & Log & & Comparison & & Reasoning & Reasoning & & Yes & \\\hline
            8 & \cite{savarimuthu2010data} & Log & Data Mining & Comparison & Yes & Log & Log & Comparison & Yes & \\\hline
            9 & \cite{riad2021run} & Data Mining & Elitism & & & & & & Yes & \\\hline
            10 & \cite{savarimuthu2013identifying} & Log & Log & Reasoning & Yes & Reasoning & Log & & Yes & \\\hline
            11 & \cite{gao2014extracting} & & & & & & NLP & NLP & & \\\hline
            12 & \cite{cranefield2016bayesian} & Data Mining & & Threshold & Yes & & & & Yes & \\\hline
            13 & \cite{campos2010case} & Log & Log & & & & & Elitism & Yes & Yes \\\hline
            14 & \cite{avery2016externalization} & Data Mining & Reasoning & & & & Reasoning & NLP & & \\\hline
            15 & \cite{sarathy2017learning} & Data Mining & & Threshold & & & Data Mining & & Yes & Yes \\\hline
            16 & \cite{dam2015mining} & & Log & & & & Log & Data Mining & & Yes \\\hline
            17 & \cite{aires2017norm} & NLP & & & & & NLP & & & \\\hline
            18 & \cite{oren2020norm} & & & & & & Reasoning & NLP & & \\\hline
            19 & \cite{mahmoud2012semantics} & Data Mining & Data Mining & Log & Yes & Log & Data Mining & & & \\\hline
            20 & \cite{murali2021mining} & Data Mining & & & & & Data Mining & NLP & & \\\hline
            21 & \cite{alechina2018incentive} & Log & & Comparison & Yes & & Log & NLP & & Yes \\\hline
            22 & \cite{dell2022complexity} & Log & & & Yes & & & & & \\\hline
            23 & \cite{christelis2010exploiting} & Log & & & & & & & Yes & \\\hline
            24 & \cite{morris2023agent} & Reasoning & Reasoning & Reasoning & Yes & Reasoning & & & Yes & Yes \\\hline
            25 & \cite{morales2013automated} & Log & Log & Comparison & & & Reasoning & & Yes & \\\hline
            26 & \cite{morales2015online} & Reasoning & Log & Reasoning & Yes & Reasoning & Reasoning & & Yes & \\\hline
            27 & \cite{liga2023fine} & NLP & & & & & NLP & & & \\\hline
            28 & \cite{liga2022transfer} & NLP & & & & & NLP & & & \\\hline
            29 & \cite{corapi2011normative} & Log & & Reasoning & Yes & & Log & & Yes &  \\\hline
            30 & \cite{tan2019s} & Data Mining & & Threshold & Yes & & Data Mining & & Yes &  \\\hline
            31 & \cite{cranefield2021identifying} & Log & & & & & Log & & &  \\\hline
            32 & \cite{pham2024multi} & & & & & & & NLP & &  \\\hline
            33 & \cite{oldenburg2024learning} & Data Mining & Data Mining & & & & Data Mining & & Yes &  \\\hline
            34 & \cite{fung2022normsage} & NLP & & & & & NLP & NLP & Yes & Yes \\\hline
            35 & \cite{moghimifar2023normmark} & NLP & & & & & NLP & NLP & &  \\\hline
        \end{tabular}%
    \label{tab:rq_detailed}
\end{sidewaystable}

\begin{figure}[ht]
    \centering
    \begin{minipage}[b]{0.49\linewidth}
        \centering
        \includegraphics[width=\linewidth]{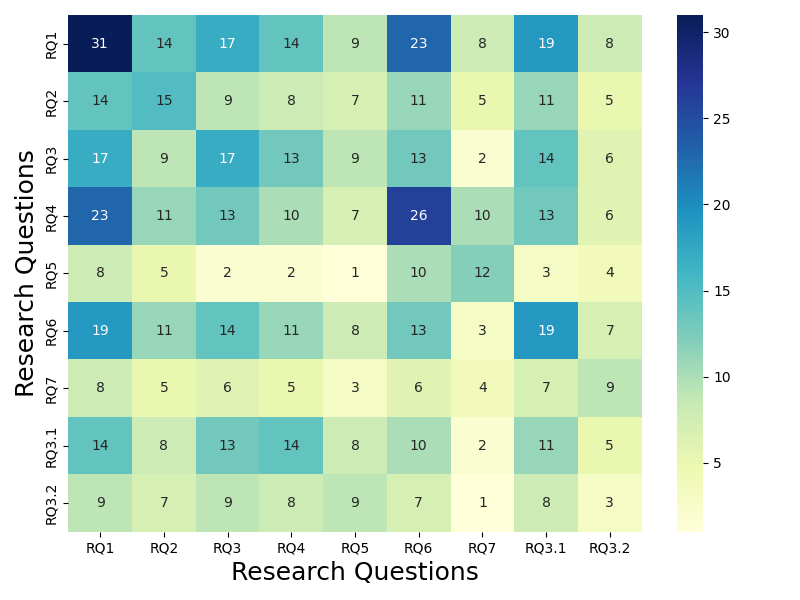}
        \caption{Comparison with the maximal value.}
        \label{fig:heatmap}
    \end{minipage}
    \hfill
    \begin{minipage}[b]{0.49\linewidth}
        \centering
        \includegraphics[width=\linewidth]{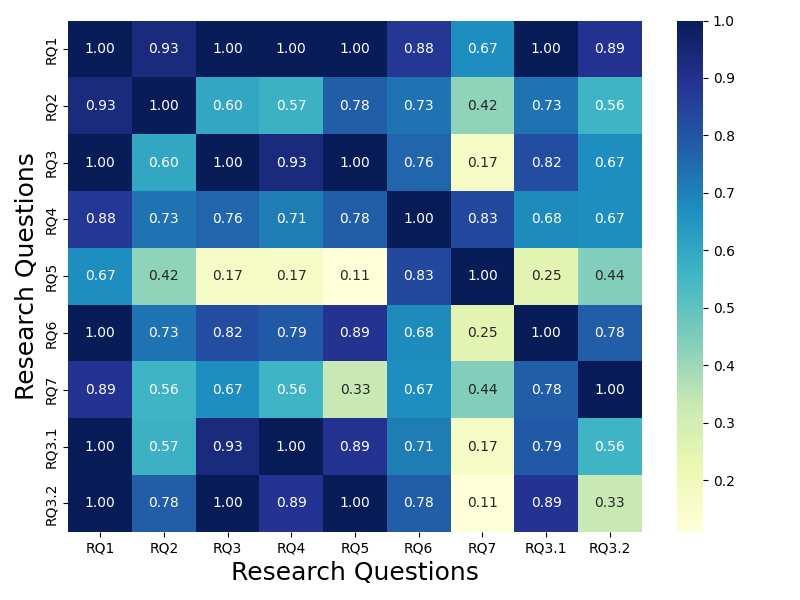}
        \caption{Comparison with the minimal value.}
        \label{fig:heatmap_normalized}
    \end{minipage}
\end{figure}

As an additional indicator, Fig.~\ref{fig:heatmap} shows the correlations among research questions by considering the most answered research question. On the other hand, Fig.~\ref{fig:heatmap_normalized} shows, given the minimum value of the number of papers answering the research question between two questions, the correlation.

It is important to note that the approaches not focusing on, or taking into account, MAS may be unable to answer a part of the research questions.

\subsubsection{Qualitative Analysis}

\paragraph{RQ1: How to identify the norms of a society by observing it?}
Various approaches have emerged in the literature, including:
Frequency-based detection methods such as the Potential Norms Mining Algorithm (PNMA) proposed by \citet{mahmoud2013potential}, which identifies norms through statistical analysis of observed behavioural patterns. Bayesian hypothesis testing as introduced by \citet{cranefield2016bayesian}, which calculates the likelihood of a candidate norm's existence. Plan recognition approaches by \citet{oren2020norm} that infer norms through observed action sequences. While deviating a bit from the research question as they do not properly observe agents, repository mining techniques~\cite{dam2015mining} and legal text analysis~\cite{ferraro2021nlp} provide insight into normative structures by extracting patterns from unstructured or semi-structured documents.

\paragraph{RQ2: How to identify the norms of a society by interacting with it?} The literature presents several interaction-based approaches: 
\citet{mahmoud2012norms} leverage event history analysis combined with ontology-driven inference, allowing agents to refine detected norms through iterative engagement.
Similarly, verification through peer interactions~\cite{savarimuthu2013identifying} to confirm candidate norms and query-based approaches~\cite{savarimuthu2010data} that actively test for norm existence are also addressed.

\paragraph{RQ3: How to differentiate individual norms from societal norms in a multi-agent society?}
\citet{mahmoud2013potential,mahmoud2016development} explore frequency-based differentiation, distinguishing descriptive norms (emerging from agent behaviours) from injunctive norms (those explicitly reinforced). Peer verification mechanisms~\cite{savarimuthu2013identifying} provide a means of validating societal norms versus personal behaviours through sanction-based differentiation. \textbf{RQ3.2} remains an open problem in the literature, as most approaches rely on multi-agent interactions. However, Bayesian updates~\cite{cranefield2016bayesian} and agent-centred norm evaluation~\cite{savarimuthu2010norm} suggest that individual agents could infer societal norms through probabilistic reasoning and historical observation.

\paragraph{RQ4: How to detect prohibition norms without norm enforcement?}
\citet{savarimuthu2013identifying,dam2015mining} present data-driven approaches using association rule mining and repository analysis to identify prohibition norms, while Bayesian event sequence analysis~\cite{murali2021mining} estimates prohibition likelihood based on historical compliance trends. The other approaches include analyses of infrequent patterns~\cite{mahmoud2013potential} that may indicate avoided behaviours, detection of absence patterns in expected action sequences~\cite{mahmoud2012norm,mahmoud2012semantics}, linguistic cues in communications~\cite{aires2017norm,ferraro2021nlp} that signal prohibited actions, and avoidance patterns in plan execution~\cite{oren2020norm}.

\paragraph{RQ5: How to detect norms in communications?}
Approaches include natural language processing of communications logs~\cite{avery2016externalization,dam2015mining}, extraction techniques specialised for formal documents like contracts~\cite{gao2014extracting}, analysis of modal verbs and deontic expressions~\cite{aires2017norm,ferraro2021nlp}, and event analysis from communication records~\cite{murali2021mining}.

\paragraph{RQ6: Is it possible to adapt to drifting norms without restarting the whole learning phase?}
\citet{riad2021run} propose mechanisms for online norm synthesis and utility-based adaptation. \citet{mahmoud2016norm} propose assimilation techniques that allow agents to incrementally adjust to shifting norms rather than resetting the learning process. The literature offers several other approaches, such as continuous monitoring and incremental updates~\cite{mahmoud2012norm,mahmoud2013potential,mahmoud2012norms}, Bayesian updating mechanisms~\cite{cranefield2016bayesian} that adjust confidence over time, case-based reasoning adaptation~\cite{campos2010case} for evolving normative systems, and online refinement techniques~\cite{morales2013automated,morales2015online} for dynamic norm synthesis.

\paragraph{RQ7: Is it possible to detect sub-communities of agents?}

\citet{campos2010case,morris2023agent} discuss case-based reasoning and agent-directed norm synthesis as potential solutions. These methods offer promising avenues for sub-community detection but also introduce challenges related to scalability and the granularity of norm differentiation within and across sub-communities. \citet{mahmoud2012norm} analyse behavioural clustering to identify societal subdivisions while other approaches include comparative analyses of repositories~\cite{dam2015mining} to identify community-specific norms, movement-based detection~\cite{savarimuthu2010norm} that analyses agent groupings, and analyses of heterogeneous groups~\cite{mahmoud2016norm} with distinct normative systems.

\subsection{Analysis of the Approaches}

This section presents a classification of the context in which each proposed approach is operating, as well as a comprehensive review of each if them.

\subsubsection{Classification of the Reviewed Approaches}

After reviewing the collected approaches, two major categories were identified based on their application context: \textit{Agent-Based} and \textit{Not Agent-Based} approaches. A method is considered Agent-Based if it identifies norms through the interactions of agents within an environment or through their communication with other individuals. The key characteristic of these approaches is the presence of actions (\emph{i.e.}, interactions) that facilitate the discovery of norms. In contrast, a method is classified as Not Agent-Based if it primarily relies on data analysis rather than interactive behaviours.
Each of these categories can be further divided into subcategories.

Agent-Based approaches can be grouped into three subcategories: \textit{Observatory}, \textit{Experiential}, and \textit{Communicative}. Observatory approaches rely on observing other agents (or traces of their actions) interacting in an environment. These methods are considered safer since they do not involve direct experimentation that could lead to norm violations. However, they may struggle to identify prohibitions, particularly when all observed agents comply with existing norms, leaving no violations to be detected. Experiential approaches operate on trial and error. This method is commonly found in behaviour learning techniques such as Reinforcement Learning. It is typically efficient and relatively simple to implement. However, unlike Observatory methods, it involves committing multiple norm violations before correctly identifying the normative behaviour. Communicative approaches rely on exchanging information with already integrated agents. Like Observatory methods, they are relatively safe. However, they tend to be the most complex to implement effectively, which limits their practical use.

\begin{figure}[ht]
    \centering
    \includegraphics[width=0.9\linewidth]{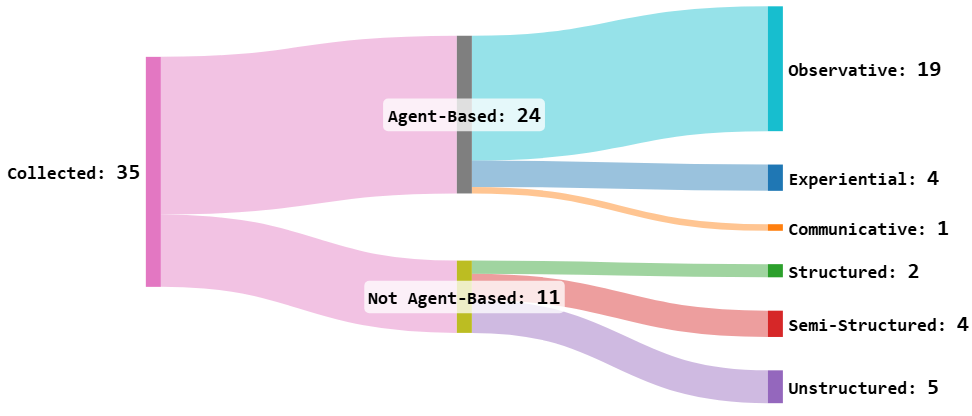}
    \caption{Taxonomy of the identified approaches.}
    \label{fig:taxonomy}
\end{figure}

Methods that do not fall under the Agent-Based category are distinguished by the type of data they process. We identified three subcategories: \textit{Structured}, \textit{Semi-Structured}, and \textit{Unstructured} data. Structured data consists of pre-encoded information, such as databases, where symbolic elements are already extracted and standardised. Approaches in this category typically apply pattern recognition techniques to identify norms. Semi-structured data includes documents that follow a standardised structure and contain recognisable keywords related to norms. Examples include legal texts and contracts. These approaches are more challenging than those using structured data, but remain more manageable than those using unstructured data. Unstructured data encompasses free-form content, such as forum discussions and natural language documents. Because these data sources are unprocessed, extracting meaningful symbols for norm identification is significantly more complex. Among the collected approaches, none addressed audio or video data, even though these are potential media for norm identification.
This classification is illustrated by Fig.~\ref{fig:taxonomy}.

\begin{figure}[ht]
    \centering
    \includegraphics[width=0.5\linewidth]{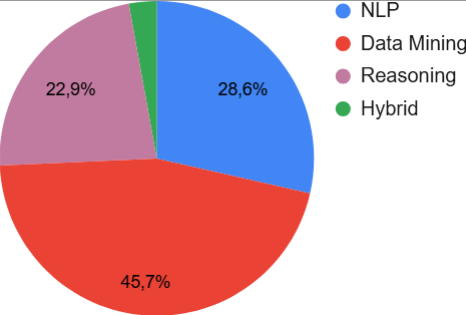}
    \caption{Main area of the proposed approaches.}
    \label{fig:category}
\end{figure}

In addition to the taxonomy based on methodology, we also classified the approaches according to their primary research focus, see Fig.~\ref{fig:category}. We identified three main areas: \textit{Natural Language Processing (NLP)}, \textit{Data Mining}, and \textit{Reasoning}. Additionally, we introduced a \textit{Hybrid} category for approaches combining at least two of these areas. NLP approaches focus on the semantic analysis of textual data to extract norms. Data Mining approaches analyse large datasets to identify patterns and infer norms. Reasoning approaches derive conclusions from limited data and refine their findings as more information becomes available. Hybrid approaches integrate elements from multiple research areas to enhance norm identification. Figures~\ref{fig:taxonomy} and~\ref{fig:category} illustrate the distribution of the methods across these categories.

\subsubsection{Norm Detection and Identification}
Several researchers have investigated techniques for detecting and identifying norms in MAS. \citet{mahmoud2013potential} propose a Potential Norms Mining Algorithm (PNMA) that enables agents to identify prevailing norms through observation of other agents' behaviours. Their approach allows an agent to revise its norms without requiring third-party enforcement mechanisms. The PNMA follows a structured process of data formatting, filtering, and extraction of potential norms from observed events. Building on this work, \citet{mahmoud2016development} present the Potential Norms Detection Technique (PNDT), which facilitates agents' adaptation to changing environments through self-enforcement. The PNDT framework comprises an agent's belief base, observation process, the PNMA algorithm, verification process, and updating process. Through simulations in an elevator scenario, they demonstrate how environmental variables affect norm detection success. \citet{cranefield2016bayesian} introduce a novel approach using Bayesian inference for norm identification. Their method effectively operates in scenarios where both compliance and violation occur regularly, calculating the odds of a candidate norm being established versus no norm existing. Empirical evaluation shows that norm-compliant behaviour can emerge after few observations. \citet{oren2020norm} develop a norm identification mechanism based on plan recognition, combining parsing-based plan recognition with Hierarchical Task Network planning to infer prevailing norms. Their approach handles norm violations through counting and thresholding, without relying on observations of explicit sanctions. \citet{sarathy2017learning} propose a norm representation scheme incorporating context-specificity and uncertainty using Dempster-Shafer theory. Their algorithm learns norms from observation while considering different contexts and the inherent uncertainty in the learning process, allowing agents to adapt to changing contexts.

\subsubsection{Norm Mining from Data}

Several researchers have explored data mining techniques for extracting norms from various sources. \citet{savarimuthu2010data} present an internal agent architecture for norm identification based on interaction observation. Their Obligation Norm Inference algorithm uses association rule mining to identify obligation norms. In related work, \citet{savarimuthu2013identifying} focus on identifying prohibition norms using a modified version of the WINEPI algorithm to generate candidate prohibition norms. Their framework considers social learning theory and distinguishes between candidate norms and identified norms. \citet{savarimuthu2010norm} further develop their architecture with the Candidate Norm Inference algorithm, which identifies sequences of events as candidate norms. Their approach enables agents to modify and remove norms if they change or no longer hold in the society, demonstrating the benefits of norm inference for utility maximisation. \citet{avery2016externalization} introduce Norms Miner, a tool for extracting norms from open source software development bug reports. Their automated approach discovers, extracts, and classifies norms from textual social interactions, making tacit knowledge explicit and accessible. The tool achieves solid performance with a recall of 0.74 and a precision of 0.73 in norm classification.

\citet{dam2015mining} explore mining software repositories for social norms, presenting results on coding convention violations across large open source projects. They propose a life-cycle model for norms within Open Source Software Development communities and demonstrate its applicability using data from the Python development community. \citet{ferraro2021nlp} apply Natural Language Processing techniques to normative mining from legal documents. They provide a comprehensive review of existing NLP techniques, particularly semantic parsing, and analyse their applicability to mining legal norms. The paper presents preliminary results on extracting normative rules using relation extraction and semantic parsing models. \citet{gao2014extracting} develop an approach for automatically extracting norms from contract text. Their prototype tool suite extracts norms and related concepts, evaluating the realism of normative models in MAS by assessing how effectively these concepts can be identified within contracts. \citet{murali2021mining} apply norm-mining techniques to a real-world dataset in international politics. They adapt a Bayesian norm mining mechanism to identify norms from bilateral sequences of inter-country events extracted from the GDELT database, demonstrating that a model combining probabilities and norms explains observed international events better than a purely probabilistic model.

\subsubsection{Norm Assimilation and Adaptation}

Several researchers have explored how agents can assimilate and adapt to norms in MAS. \citet{mahmoud2012norms} propose a technique for software agents to detect and assimilate norms to comply with local normative protocols. Their conceptual framework includes stages for a visitor agent to detect norms by analysing interaction patterns and matching them with a ``norms model base''. \citet{mahmoud2016norm} introduce a norm assimilation approach for MAS in heterogeneous communities. Their theoretical framework is based on an agent's internal belief about its ability to assimilate and its external belief about the assimilation cost associated with different social groups. They categorise assimilation decisions based on whether an agent ``can assimilate'', ``could assimilate'', or ``cannot assimilate''. \citet{mahmoud2012semantics} focus on defining the semantics of a proposed norms mining technique. They explicitly define the semantics of the entities and processes involved in norms mining, drawing inspiration from existing work in norms, normative systems, and data mining. \citet{mahmoud2012norm} outline a conceptual approach for norms detection and assimilation, focusing on discovering norm emergence based on interaction patterns between agents. Their approach utilises a norms mining technique and proposes using a norms learning technique to define the semantics of textual data.

\subsubsection{Norm Synthesis and Revision}

Several researchers have investigated techniques for synthesising and revising norms in MAS. \citet{morales2013automated} introduce IRON (Intelligent Robust On-line Norm synthesis mechanism), which synthesises conflict-free norms without over-regulation. IRON produces norms that characterise necessary conditions for coordination and are both effective and necessary, with the capability to generalise norms for concise normative systems. \citet{morales2015online} present an extended IRON mechanism designed for online synthesis of compact normative systems. Their enhanced approach incorporates improved evaluation methods, a generalisation operator requiring sufficient evidence, and a specialisation operator for refining underperforming generalisations. Empirical evaluation shows that IRON significantly outperforms BASE in terms of stability and compactness. \citet{riad2021run} propose a utility-based norm synthesis model for managing norms in complex MAS with multiple, potentially conflicting objectives. Their approach employs utility-based case-based reasoning for run-time norm synthesis, using a utility function derived from system and agent objectives to guide norm adoption. \citet{dell2022complexity} analyse the complexity of synthesising and revising conditional norms with deadlines. They demonstrate that synthesising a single conditional norm correctly classifying behavioural traces is NP-complete, as is synthesising sets of conditional norms and minimal norm revision. \citet{christelis2010exploiting} detail a first-order approach to norm synthesis, allowing for greater expressiveness through the use of variables. They propose optimisations to improve the performance of first-order norm synthesis, including a priori filtering, traversal pruning, repetitive operators, and duplicate runs. \citet{morris2023agent} propose an agent-directed norm synthesis framework that allows norms to be synthesised based on agent requests and interactions. Their approach involves individual agents in system governance, enabling revisions that benefit individual goals without conflicting with system-level objectives.

\subsubsection{Norm Conflict Detection and Resolution}

\citet{aires2017norm} focus on identifying potential conflicts between norms in contracts written in natural language. They develop a semi-automatic approach for identifying norms and their elements using information extraction techniques. Their tool helps prevent conflicts by comparing extracted norm information and classifying potential conflicts into types such as permission-prohibition, permission-obligation, and obligation-prohibition. \citet{alechina2018incentive} address the problem of detecting norm violations in open MAS. They demonstrate that perfect or near-perfect norm monitoring and enforcement can be achieved at no cost to the system, proposing incentive-compatible mechanisms for decentralised norm monitoring where agents themselves perform monitoring. \citet{campos2010case} propose adding an ``Assistance layer'' to MAS to handle norm adaptation. They use a Case-Based Reasoning approach within this layer, enabling the system to learn from past experiences and adapt norms to achieve organisational goals, illustrated through a Peer-to-Peer sharing network scenario.

\subsection{Takeaways}
\label{sec:takeaways}

To conclude this section, we provide some takeaways that one should consider when choosing an approach for norm mining, as not all share the same characteristics.

First, the selected approach may depend on the type of data that can be collected. Some approaches, such as those based on data mining, can perfectly learn from structured observations. On the other hand, when such data is not available, NLP-based approaches are suitable either by dealing with unstructured data or by exchanging with agents. Due to the way \pino is meant to function, it may be better to avoid techniques relying on communications with agents. Similarly, approaches that rely on interactions within the environment might not be suitable. Reasoning techniques might be useful when it comes to inferring more high-level norms abstracting from local and personal behaviours, or deriving context-specific norms from very general norms.\footnote{For example, ``One should not cause harm'' is very broad as there are many ways to cause harm.}

Then, the chosen method should also depend on the specificities of the observed normative system. One key point, in particular, is whether the agents have many personal norms. Some approaches cannot differentiate between those and the global norms, and for this reason are possibly not suitable for an application within the \pino architecture since the type of agent with which it deals (mainly humans) may have many personal norms in addition to the global ones.

Finally, some features that collected approaches may have are not relevant for \pino. For instance, being able to handle norm drifting is not, as in our case, the norm mining approach is meant to be used only once. Consequently, there cannot be norm drifting within the observation window. Similarly, mechanisms for conflict resolution are not relevant, as conflicts are handled anyway by \pino.

Building on the overview of existing techniques, the next section introduces the ARIA algorithm, which structures exceptions to norms as an argumentation framework to provide intelligible and explainable models.

\section{Argumentative Rule Induction}
\label{sec:aria}

This section introduces the ARIA (Argumentative Rule Induction A-star) algorithm~\cite{alcaraz2025star,alcaraz2025providing}, and indicates how it can be used to generate explanations. It describes how each component of the model works. First, it shows how the model would bring intelligibility to a black-box algorithm when deployed. Second, it details how it generates an explainable model with the help of some data, as well as a final explanation by using the algorithm proposed by \citet{fan2014computing,bex2021xaiargs}.

\subsection{Computing a Justification}

The goal of the approach is to justify the decision of a black-box algorithm (which will be referred to in the rest of the section as a BB or agent). More specifically, we want to observe one action and why it was, or was not, chosen. For example, why did the autonomous car apply the brakes or the accelerator. To do so, we introduce two core elements, as well as a running example with the Car dataset~\cite{misc_car_evaluation_19} to better understand how the approach should work.

\subsubsection{Universal Graph}
\label{sec:universal}

In order to be used, the approach requires a dataset that describes the behaviour of an agent in various situations or states. Such a dataset should provide as input the perceptions of the BB algorithm, such as raw sensor information or preprocessed and potentially high-level data, and as a label, if the tracked action has been performed or not (\emph{e.g.}, \texttt{turn\_right=True}). Once this dataset is available, the search process, described later in Section~\ref{sec:algorithm}, can start.

Once finished, this search process will return a graph, inspired by the structure of an argumentation framework, representing the overall relationship among the perceptions and how they can influence the final decision. We call this graph the Universal Graph, as it contains relations among all possible factors which will serve to construct an explanation later on. In the implementation, the nodes of the graph consist of attribute-value tuples from the dataset inputs. However, it is possible to use expert knowledge or automated feature engineering such as the one we developed~\cite{nourbakhsh2024feature} to generate more sophisticated attributes or values. For example, instead of having a node with a very arbitrary value ``\texttt{size=3.5cm}'', an expert could design values such as ``\texttt{size=small}'', which would include anything with a size less than 5 cm.

Additionally, there are two other distinguished nodes. The first is called the target and is written as $\target$. This argument represents the decision made by the agent. As will be detailed in the next section, it will denote the fact that the agent performed the tracked action if it is part of the extension, or otherwise that the agent did not. The second additional node is written as $\toparg$. Although this argument is not bound to any attribute-value tuple, it denotes a support relation to the target from the other nodes, as explained by \citet{boella2009meta}. As such, this node can only attack the target. Any other node attacking it will then be considered as supporting the target due to the defence relationship created between them.\footnote{Note that this differs from the support relation as it is defined in Bipolar Argumentation Frameworks~\cite{yu2020principle,yu2023principle}.}

\subsubsection{Contextual Graph}
\label{sec:contextual}

Once in possession of the Universal Graph, one can then compute what is called the Contextual Graph. This graph is a projection of the Universal Graph given a set of facts, \emph{i.e.}, a set of perceptions from the agent in a specific situation.\footnote{This may also correspond to the input of one entry of the BB's behavioural dataset.} As a consequence, it filters out all the nodes whose value is not part of the facts, except for $\target$ and $\toparg$ which are always in the Contextual Graph. The resulting graph is then a useful Argumentation Framework for representing the belief state of an agent.

It is now possible to compute an extension based on this Contextual Graph. While in theory any extension would be usable with some adjustments, we chose to use the grounded extensions as it features two advantages over the others. First, it is unique, and as such less ambiguous that a set of extensions when it comes to providing an explanation to the end user. Second, it computes in polynomial time\footnote{While the preferred extension also shares a polynomial complexity, it still requires a greater time to be computed.} which is an interesting feature to shorten the search phase.

Then, if the target $\target$ is part of the extension, and if this matches the decision that the BB made (\emph{i.e.}, doing the tracked action or not), it can serve to justify this decision by potentially providing an explanation based on this graph. This includes the cases where the BB took the wrong decisions, as long as the BB output matches the graph output. There exist several ways to provide an explanation from an argumentation framework~\cite{liao2021representation,fan2014computing,doutre2023visual}. In this case, we chose to use the one described by \citet{bex2021xaiargs,fan2014computing} described in the following subsection.

\subsubsection{Explanation Extraction}\label{sec:xaiextract}

The approach introduced by the work of \citet{fan2014computing} is characterised by finding all \textit{related admissible} extensions of an argumentation framework. That is:

\begin{definition}[Related admissibility]
\label{def:ra}
    If $\langle\args, \defs\rangle$ is an argumentation framework, then any subset $S \subseteq \args$ is a related admissible extension iff $S$ is an admissible extension and there exists $x \in S$ \emph{s.t.} $S$ defends $x$. Any such $x$ is called a topic of $S$.
\end{definition}

 Such related admissible extensions correspond to explanations of their topics: Given an accepted argument, the set of related admissible extensions having that accepted argument as their topic corresponds to the set of explanations, in the sense of being justifications~\cite{horridge2011justification}, for the acceptance of that argument. \citet{fan2014computing} provide a method for computing these related admissible extensions by procedurally constructing them as dispute trees that have the topic, \emph{i.e.}, the argument whose acceptance is to be explained, as the root. \citet{bex2021xaiargs} define a polynomial algorithm that may be used to compute the related admissible extensions of a given argumentation framework. Specifically, for any $\langle\args, \defs\rangle$, let $DefBy(A) = \{B \in \args \mid B \textup{ defends } A\}$ be such that $A, B \in \args$, and $DefBy(A, S) = DefBy(A) \cap S$ for any extension under some semantics $S \subseteq \args$. Then \citet{bex2021xaiargs} define a procedure to calculate these sets $DefBy(A)$, allowing the \citet{fan2014computing} approach to be computed by taking all admissible extensions $S^{adm}_i$ of an argumentation framework and calculating $DefBy(A, S^{adm}_i)$ to find all related admissible extensions for a topic $A$, \emph{i.e.}, the accepted argument that is to be explained.

In the context of this work, the base argumentation frameworks correspond to the contextual graphs described above. Depending on whether the $\target$ argument is accepted or not, the explanation can be calculated using the $DefBy(\cdot)$ procedure. Say that the $\target$ argument is accepted; then computing $DefBy(\target, S)$ for some extension $S$ of the contextual graph provides a justification for $\target$ under those semantics.

Alternatively, if the target is rejected, we may use the procedures defined by \citet{bex2021xaiargs} to compute the set $NotDef(A,S)$ which returns the set of all, direct and indirect, attackers of an argument $A$ for which there is no defence in the extension $S$. This may be used to provide a justification, \emph{i.e.}, the explanatory set, for the non-acceptance of $A$ under a semantics. In the case of the grounded extension, since it is unique then $S$ may be the unique grounded extension used to calculate the explanatory set $NotDef(A,S)$. For the general case where there may be multiple extensions, then under a semantics $sem$ providing the set of extensions, for some $0 < i \leq n$, $S_i$, the explanatory set \[NotAcc_{sem}(S) = \bigcup^{n}_{i = 1} NotDef(A,S_i)\] provides a full justification set for the non-acceptance of $A$ under some semantics.

\subsubsection{Example}
\label{sec:example}

\begin{figure*}
    \centering
    \includegraphics[width=0.98\textwidth]{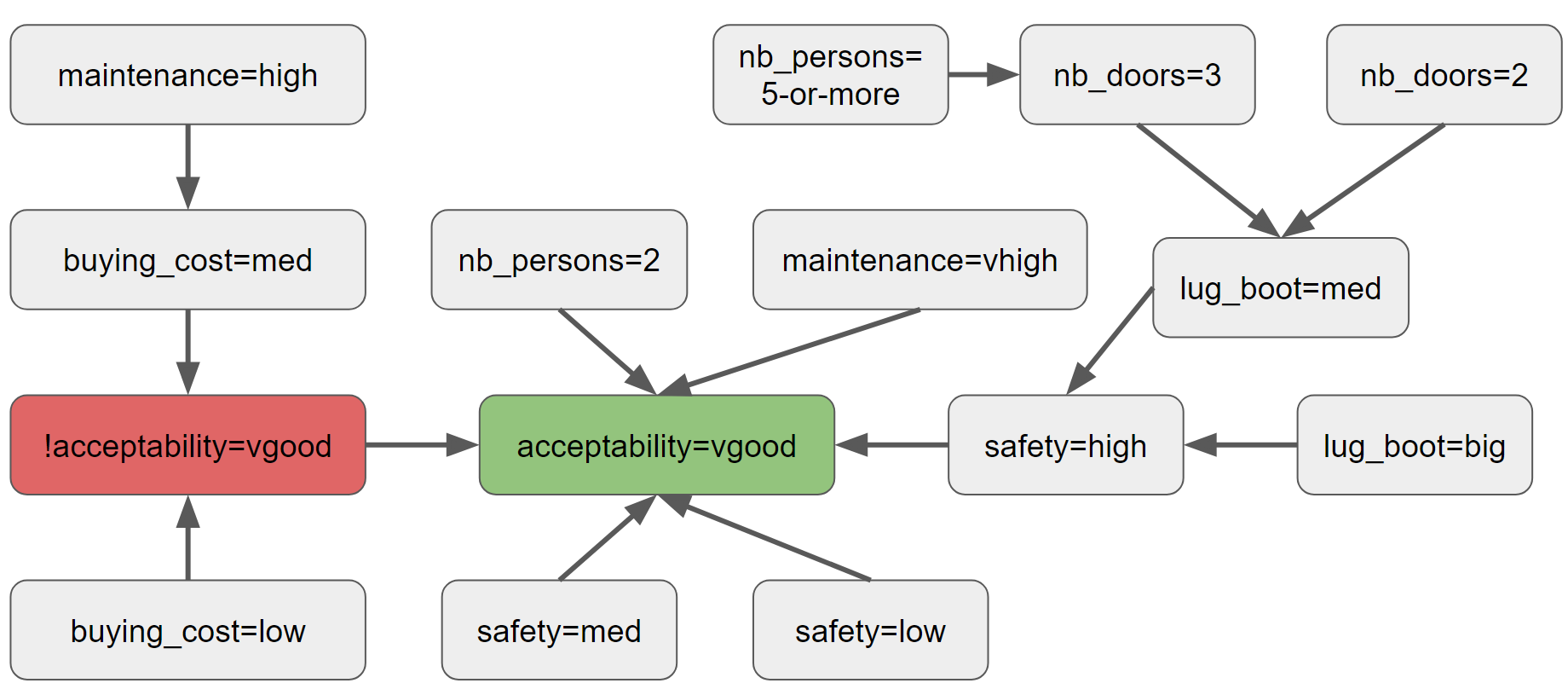}
    \caption{Universal graph for the Car dataset. The $\target$~argument is in green. The $\toparg$~argument is in red.}
    \label{fig:universal}
\end{figure*}

In order to make it clearer, we recall the example from \citet{alcaraz2025star} using the Car dataset~\cite{misc_car_evaluation_19}. Fig.~\ref{fig:universal} represents a graph that has been inferred with the approach from the aforementioned dataset.
The task is to evaluate the buying acceptability of a car based on several attributes and their values, such as the number of seats, the buying cost, the maintenance cost, etc. There are four classes, namely ``Unacceptable'', ``Acceptable'', ``Good'', and ``Very good''.
In this specific case, the target argument is associated with the class ``Very good'', \emph{i.e.}, ``The car has a very good buying acceptability''. As such, the target not being in the extension can be classified as ``Unacceptable'', ``Acceptable'', or ``Good''. Fig.~\ref{fig:universal} shows the Universal Graph and then gives an overview of how each argument (\emph{i.e.}, couple attribute-value) can influence the final decision.
However, it is not a justification yet.

\begin{table}
    \centering
    \caption{Summary of the facts, \emph{i.e.}, the values for the different attributes, for the example.}
    \begin{tabular}{|rcl|}\hline
        \textbf{Attribute} &   & \textbf{Value}\\\hline
        Buying cost & = & Medium\\
        Maintenance cost & = & Low\\
        Number of doors & = & 3\\
        Number of seats & = & 5-or-more\\
        Luggage boot & = & Medium\\
        Safety & = & High\\
        Acceptability (Label) & = & Very-good\\\hline
    \end{tabular}
    \label{tab:facts}
\end{table}

\begin{figure*}
    \centering
    \includegraphics[width=0.95\textwidth]{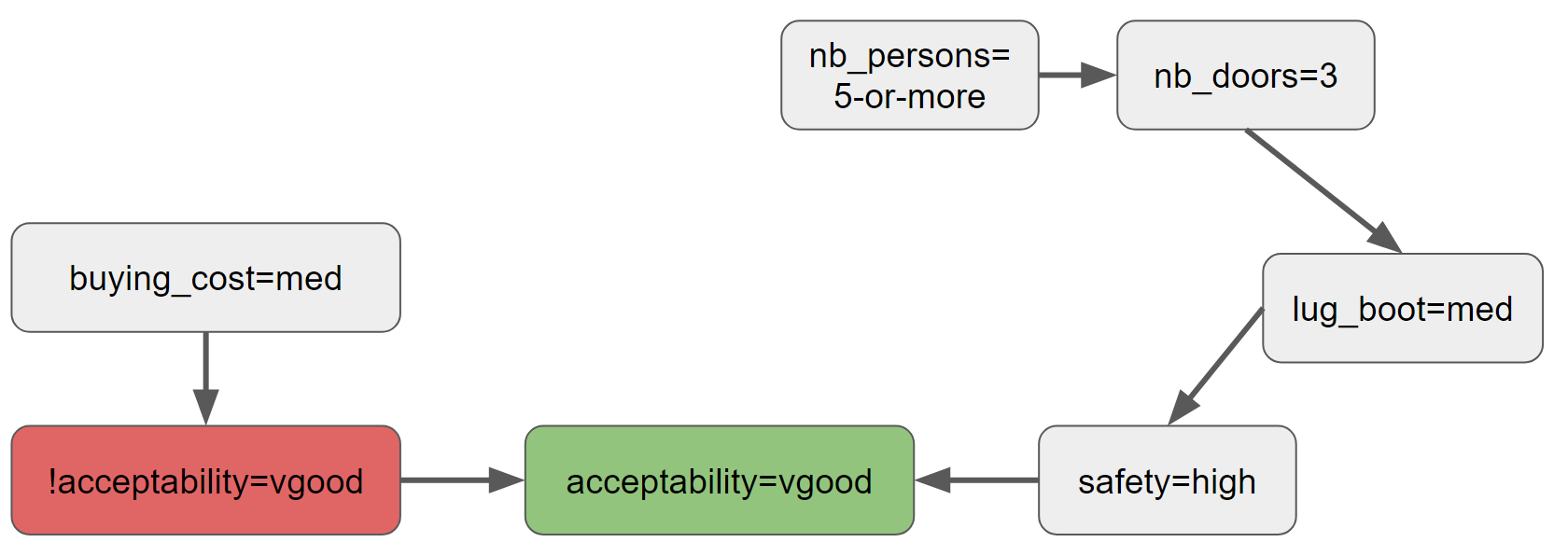}
    \caption{Contextual graph for the Car dataset and a specific input. The target argument is in green. The $\toparg$~argument is in red.}
    \label{fig:contextual}
\end{figure*}

Fig.~\ref{fig:contextual} shows the Contextual Graph derived from the Universal Graph shown in Fig.~\ref{fig:universal}. The latter corresponds to the set of facts presented in Table~\ref{tab:facts}, \emph{i.e.}, the values assigned to the different attributes.
In this example, the car has been classified as ``Very good'' in terms of buying acceptability. From the Contextual Graph, we can see the elements which lead to this decision. For example, because high safety was not sufficient, it was also required that it has a medium luggage boot. On the other hand, if it did not have at least 5 seats, the car would not have been classified as ``Very good'' since the number of doors was 3. Similarly, the medium buying cost supports the classification by attacking the $\toparg$~argument.

Applying the explanation procedure outlined in Section~\ref{sec:xaiextract}, using the procedure from \citet{bex2021xaiargs}, we obtain the set:\\ $DefBy(\texttt{acceptability=vgood})$ $=$ $\{ \texttt{acceptability=vgood}$, $\texttt{buying\_cost=med}$, \\$\texttt{lug\_boot=med}$, $\texttt{nb\_persons=5-or-more} \} $

An admissible extension $S_{adm}$ of the contextual graph above is:\\
$S_{adm} = $ $\{\texttt{acceptability=vgood}$, $\texttt{buying\_cost=med}$, $\texttt{lug\_boot=med}$, \\$\texttt{nb\_persons=5-or-more}\} $

So $DefBy(\texttt{acceptability=vgood} \cap S_{adm}) = DefBy(\texttt{acceptability=vgood}) = S_{adm}$. This set of arguments then provides a justification set for the topic corresponding to the target \texttt{acceptability=vgood}. Therefore, this explanation may be interpreted as follows: Whether a car is classified as very good is primarily influenced by its price (\emph{i.e.}, buying cost), and the luggage space if the number of seats is five or more.

\subsection{$n$-arguments}
\label{sec:narg}

Let the set of arguments corresponding to the presence of a couple attribute-value in the facts be called ``Positive arguments'' (or $p$-arguments). It is then possible to define ``Negative arguments'' (or $n$-arguments). In opposition to $p$-arguments, which appear in the contextual graph only if the facts match their condition, $n$-arguments are present in the contextual graph if their condition is not present in the facts. They denote a missing fact. While using only the set of $p$-arguments makes the representation of certain logical formulae impossible, the addition of the $n$-arguments allows one to represent a broader range of formulae. However, adding these extra arguments increases the number of neighbours for each node and, as such, increases the total computation time.

More formally, let $\atms$ be a non-empty set of propositional atoms. In this application, each propositional atom represents the possible valuation of an attribute, \emph{e.g.}, $q \coloneqq$ \\``\texttt{buying\_cost=Medium}'' $\in \atms$. We define the language  $\mathcal{L}_{\atms}$ as the set of well-formed formulas (wff), with the following BNF grammar, for any $p \in \atms$ :  
$$\phi \coloneqq \bot \; |\;  p \; |\; \neg \phi\; |\;  \phi \vee \psi\;   $$

As usual, we use the following notation shortcuts:
\begin{itemize}\tightlist
    \item $\top \coloneqq \neg \bot $, 
    \item $\phi \wedge \psi \coloneqq \neg(\neg \phi \vee \neg \psi)$,
    \item $\phi \Rightarrow \psi \coloneqq \neg \phi \vee  \psi$
\end{itemize}

An interpretation model $I$ over a valuation $V: \atms \to \{\top,\bot\}$ is given by the function $I: \mathcal{L}_{\atms} \to \{\bot,\top\}$ \emph{s.t.} 
 $\forall \phi \in \mathcal{L}_{\atms}, I \models \phi \text{ iff } I(\phi) = \top$ where:
\begin{enumerate}\tightlist
	\item $\forall p \in \atms, I \models p$ iff $V(p) = \top $
	\item $\forall \phi, \psi \in \mathcal{L}_{\atms}, I \models \phi \vee \psi$ iff $I \models \phi  $ or $I \models \psi$
	\item $\forall \phi \in \mathcal{L}_{\atms},I \models \neg \phi$ iff $I \not\models \phi$

\end{enumerate}

Additionally, we will denote the set of symbols $\bar{\atms} \coloneqq \{\bar{p}: p \in \atms\}$ that corresponds to the negation of propositional atoms. Henceforth, we call a class of interpretation models based on the set of valuations $\Omega \subseteq \{\top,\bot\}^{\atms}$ a \emph{propositional dataset}. We note that $\forall \phi \in \mathcal{L}_{\atms}, \Omega \models \phi  $ iff for all $V \in \Omega$, the interpretation model $I_V$ over $V$ is \emph{s.t.} $I_V \models \phi$.

\begin{example}
    The following example aims at showing a situation that cannot be represented solely using $p$-arguments and would require the addition of the $n$-arguments.
    We represent an argument as a couple, composed of a name $A$, and a condition, represented as a propositional formula of $\mathcal{L}_{\atms}$. 
Let $\args = \{(\target, \top), (\toparg, \top), (A, a), (B, b)\}$ with $a, b \in \mathcal{P}$, be the set of arguments and $\mathcal{F} \in \Omega$ be the set of facts.
    We describe a situation in which $\target$ should be part of the grounded extension, denoted $In(\target)$, under the following condition: $$In(\target) \rightarrow ((a \in \mathcal{F} \wedge b \notin \mathcal{F}) \vee (a \notin \mathcal{F} \wedge b \in \mathcal{F}))$$
    Without duplicating the arguments, it is not possible to represent such condition.
    Now, let us add to the current set of arguments the $n$-arguments $\bar{a}, \bar{b} \in \bar{\mathcal{P}}$, such that: $\args = \{(\target, \top), (\toparg, \top), (A, a), (B, b), (C, \bar{a}), (D, \bar{b})\}$.
    It is now possible to construct an argumentation framework with the set of arguments $\args = \{\target, \toparg, A, B, C\}$ and the attack relation $\atts = \{(\toparg, \target), (A, \toparg), (B, A), (C, B)\}$. 

    This argumentation framework can correctly represent the fact that $\target$ should be part of the grounded extension if $a$ is in the facts but not $b$, or if $b$ is in the facts but not $a$.
\end{example}

\subsection{Bipolar Argumentation}
\label{sec:baria}

In addition to the two variants proposed in \citet{alcaraz2025star}, we present a variant making use of bipolar argumentation proposed in \citet{alcaraz2025providing}. 

\subsubsection{The framework}

Within classical abstract argumentation frameworks~\cite{dung1995acceptability}, arguments are considered as abstract entities that can attack each other. However, in real-world debates and reasoning, we often find not only attacks but also supports between arguments. The following definition introduces the Bipolar Argumentation Framework (BAF) based on the definitions of~\citet{cayrol2005acceptability,yu2020principle}.

\begin{definition}

    We call \emph{Bipolar Argumentation Framework} (BAF) a tuple \\$\langle\args,\R, \Su\rangle$ where:
    \begin{itemize}\tightlist
        \item $\args$ is a non-empty set of arguments
        \item $\R \subseteq \args \times \args$ is a binary relation on $\args$ called \emph{attack relation}
        \item $\Su \subseteq \args \times \args$ is a binary relation on $\args$ called \emph{support relation}
    \end{itemize}

\end{definition}

Let $\langle\args,\R, \Su\rangle$ be a BAF. We define the notion of supported attack and indirect attack.

A supported attack occurs when an argument $A$ supports another argument $B$, and $B$ attacks a third argument $C$.
Even though $A$ does not directly attack $C$, it indirectly contributes to that attack by reinforcing $B$.
So, $A$ is considered to supported-attack $C$.

\begin{definition}[Supported attack]
$A \in \args$ \underline{supports the attack} of an argument $B\in \args$ iff there exists $(A_1,\ldots,A_n) \in \args^{n}$ \emph{s.t.} $A_1 = A$, $A_n = B$, $ A \Su A_2, \ldots, A_{n-2} \Su A_{n-1}$ and $ A_{n-1} \R B $.
\end{definition}

An indirect attack occurs when, for arguments, $A$, $B$, and $C$, $A$ attacks $B$, and $B$ supports $C$.
Here, $A$ undermines the support to $C$, and thus indirectly attacks $C$.

\begin{definition}[Indirect attack]
$A \in \args$ \underline{indirectly attacks} $B\in \args$ iff there exists $(A_1,\ldots,A_n) \in \args^{n}$ \emph{s.t.} $A_1 = A$, $A_n = B$, $ A \R A_2$ and $A_{2} \Su A_{3} \ldots, A_{n-1} \Su A_{n}$. 

\end{definition}

In everyday reasoning and debate, arguments not only attack one another (\emph{e.g.}, by contradiction), but also support each other (\emph{e.g.}, by reinforcing a shared conclusion). Bipolar Argumentation Frameworks (BAFs) capture this dual nature by allowing two types of relationships between arguments: Attack and support.
To reason about groups of arguments and their interactions, we introduce the notion of set-based relations, that is, how a set of arguments can attack, support, or defend another argument. This extends classical argumentation where only pairwise attacks are considered.
A set of arguments can set-attack another argument not only through direct attack but also via indirect or support-based pathways.
A set can also defend an argument by countering all its attackers (in this extended sense).
These set-based notions express a way to model how coalitions of arguments interact, reinforce, and counteract each other in structured debates.
What follows are the formal definitions of these concepts: Set-attack, set-support, and defence (also called acceptability).
\begin{definition}[Sets of Arguments and relations]
\vspace{-.05cm}
\begin{itemize}
    \item$\SA\subseteq\args$ \underline{set-attacks} $B \in \args$ \\iff $\exists  A \in \SA$, $A$ supports the defeat of $B$, or $A$ indirectly attacks $B$, or $A \R B$
    \item $\SA\subseteq\args$ \underline{defends} $A\in\args$ (or $A$ is acceptable with respect to $\SA$) if, and only if,  \\$\forall B\in\args$, if $\{B\}$ set-attacks $A$, then $\exists C\in\SA$, $\{C\}$ set-attacks $B$
\end{itemize}

\end{definition}

In BAF, the notion of conflict-freeness stays the same, \emph{i.e.}:
\begin{definition}[Conflict-free] 
$\SA\subseteq\args$ is a \underline{conflict-free} set of arguments if, and only if $\not\exists A,B\in\SA$ \emph{s.t.} $\{A\}$ set-attacks $B$.
\end{definition}

To evaluate which arguments are justified, we extend the idea of grounded semantics from Dung's framework to this richer setting by considering the definitions provided in \citet{yu2020principle}.

Grounded semantics identifies the most cautious (sceptical) set of arguments that can be accepted based on their ability to defend themselves against attacks while considering both direct and derived (\emph{e.g.}, supported or indirect) attacks.
The \emph{grounded extension} is the smallest (least fixed point) set of arguments that:
\begin{itemize}\tightlist
    \item Do not attack each other
    \item Defend themselves against all set-attacks from outside the set
    \item Are supported (possibly indirectly) by arguments within the set.
\end{itemize}
Based on the definition provided in \citet{yu2020principle}, we define the grounded extension for bipolar argumentation as the following :
\begin{definition}[Grounded extension]
Let \( \langle \args, \atts, \Su \rangle \) be a Bipolar Argumentation Framework.

We define a characteristic function \( F : 2^{\args} \to 2^{\args} \), where:
\[
F(S) = \left\{ A \in \args \mid \text{every set that set-attacks } A \text{ is itself set-attacked by some } C \in S \right\}
\]

The \emph{grounded extension} is the least fixed point of \( F \), \emph{i.e.}, the smallest set \( S \subseteq \args \) \emph{s.t.}:
\[
F(S) = S
\]

\end{definition}

This set contains the arguments that are initially acceptable (not attacked), and those that become acceptable step by step, as they are defended by already accepted ones. 

Taking into account these previous definitions, in the next section, we provide an algorithm which adapts the approach in \citet{alcaraz2025star} to BAF.

\subsubsection{Algorithm for bipolar argumentation}

\begin{figure}
   \centering
   \includegraphics[width=0.8\linewidth]{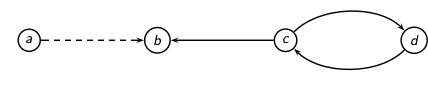}
   \caption{Example of bipolar argumentation framework. Solid edges denote attacks, and dashed edges supports.}
   \label{fig:baf-case}
\end{figure}

In this variant, a new ``Support'' relation is added. Furthermore, we adapt the grounded extension to incorporate the support relation in Algorithm~\ref{algo:bipolar}. 
In this algorithm, an argument is part of the extension if all its attackers are labelled \textsc{out}, or if it has at least one direct supporter which is \textsc{in} or \textsc{sup} (\emph{i.e.}, supported). For example, in Fig.~\ref{fig:baf-case}, the grounded labelling according to \citet{yu2023principle} would be $a$ labelled \textsc{in}, $b$ labelled \textsc{sup}, and $c$ and $d$ labelled \textsc{undec}. If the facts contain only $\{a,b,c\}$, we would have $a$ and $c$ \textsc{in}, $b$ \textsc{sup}, and $d$ \textsc{out}. All the arguments labelled \textsc{in} or \textsc{sup} are then considered part of the extension.

In order to adapt the justifications obtained by the methodology outlined in Section~\ref{sec:xaiextract}, the definition and associated procedure for computing the set $DefBy(\cdot)$ must be adapted to incorporate the support relation in bipolar argumentation frameworks.

More specifically, we use the notion of $defended_2$ introduced by \citet{yu2023principle}.

\begin{definition}[$Defended_{2-3}$]
    Let $F = \langle\args,\atts,\Su\rangle$ be a BAF and $Ext \subseteq \args$ be an extension.
    \begin{itemize}\tightlist
        \item The set of arguments \underline{$defended_2$} by the extension $Ext$, written as $d_2(F, Ext)$, is the set of arguments $A \in \args$ \emph{s.t.} for all $B\atts A$ with $B \in \args$, there exists an argument $C \in Ext$ \emph{s.t.} $C\atts B$, and there is an argument $D \in Ext$ \emph{s.t.} $D \Su C$ (supported-defence).
        \item The set of arguments \underline{$defended_3$} by the extension $Ext$, written as $d_3(F, Ext)$, is the set of arguments $A \in \args$ \emph{s.t.} for all $B\atts A$ with $B \in \args$, there exists an argument $C \in Ext$ \emph{s.t.} $C\atts B$, and for all arguments $D \in \args$ \emph{s.t.} $D \Su B$, there exists an argument $E \in Ext$ \emph{s.t.} $E\atts D$ (attacking-defence).
    \end{itemize}
\end{definition}

Using these definitions, the supports can now be integrated into Definition~\ref{def:ra}, \emph{i.e.}, \emph{Related Admissibility}, and can as such be part of an explanation.

\begin{algorithm}[ht]
\caption{Computes the Bipolar Argumentation Framework's extension}
\label{algo:bipolar}
\begin{algorithmic}[1]
    \Require $Facts$, $Args$
    \Ensure $Extension \subseteq Args$
    \State $Lab \coloneqq Args \to \{\textsc{in}, \textsc{out}, \textsc{undec}, \textsc{sup}, \textsc{must\_sup}\}$
    
    \Comment{Initialize all the arguments}
    \For{$x \in Args$}
        \If{$x \notin Facts$}
            \State $Lab(x) \coloneqq \textsc{out}$
        \Else
            \State $Lab(x) \coloneqq \textsc{undec}$
        \EndIf
    \EndFor

    \While{$\exists x \in Args$ \emph{s.t.} $(Lab(x) = \textsc{undec} \land \forall a \in \{x\}^-. Lab(a) = \textsc{out}) \lor Lab(x) = \textsc{must\_sup}$}
        \If{$Lab(x) = \textsc{must\_sup}$}
            \State $Lab(x) \coloneqq \textsc{sup}$
        \Else
            \State $Lab(x) \coloneqq \textsc{in}$
        \EndIf

        \For{$y \in \{x\}^+$} \Comment{Arguments attacked by $x$}
            \If{$Lab(y) \notin \{\textsc{sup}, \textsc{must\_sup}\}$}
                \State $Lab(y) \coloneqq \textsc{out}$
            \EndIf
        \EndFor

        \For{$y \in \{x\}_S^+$} \Comment{Arguments supported by $x$}
            \State $Lab(y) \coloneqq \textsc{must\_sup}$
        \EndFor
    \EndWhile

    \State \Return $Extension \coloneqq \{x \in Args \mid Lab(x) \in \{\textsc{in}, \textsc{sup}\}\}$
\end{algorithmic}
\end{algorithm}

\subsection{Search Method}
\label{sec:algorithm}

As previously said, the algorithm is performing an \astar~search to generate the final argumentation framework. This algorithm is described in Algorithm~\ref{algo:astar}. The main principle is to explore a search space by walking from neighbour to neighbour, following a heuristic that maximises the prediction accuracy.
In the implementation, given a set of attributes $Att$ and a set of available values $Val_i$ for an attribute $i$, we define the set of all the arguments which can be represented by the dataset attributes and values as: $$\args \coloneqq \{\target, \toparg\} \cup \{q : i \in Att,  j \in Val_i, q \coloneqq ``Att_i = Val_{ij}''\}$$

However, doing this with attributes featuring continuous value could generate too many, meaningless, arguments. To avoid this, we segment them into intervals. As such, an attribute $\phi$ having a value ranging from $0$ to $10$ would generate the arguments \texttt{$\phi$=0-2}, \texttt{$\phi$=2.1-4}, \ldots, with an interval size depending on the number of segments we wish to generate. Although this is convenient to quickly parse a dataset, we recommend the use of expert knowledge to design arguments based on these numerical values, such as \texttt{$\phi=$Above\_the\_average}.

We can then encode the attack relation as a matrix of size $|\atts| = |\args| \times |\args|$, where the element $\atts_{ij}$ is equal to $1$ if for $A_i,A_j \in \args$ there is an attack from $A_i$ to $A_j$, $0$ if not, and $-1$ if the attack is disallowed (\emph{i.e.}, would create a reflexive attack, a symmetrical attack\footnote{As the choice of the grounded extension---to guarantee the uniqueness of the explanation---is an undesirable semantics for graphs containing cycles, we disallow them. However, it is possible to allow them if more sophisticated extensions are used.}, or the attacker would be the $\target$~or $\toparg$~argument).
If the bipolar version of the algorithm is used, the value $2$ denotes a support. We forbid attacks between two arguments instantiated from the same attribute, as they would not be able to be part of the set of facts at the same time. However, if the dataset used is multivalued, this feature can be removed.
As such, we define two nodes (\emph{i.e.}, solutions) as neighbours if they differ by one value in this matrix, that is, whether their associated graphs differ by one attack (or support).

\begin{algorithm}[ht]
\caption{A* Search Algorithm}
\label{algo:astar}
\begin{algorithmic}[1]
    \Require \texttt{MaxIterations}
    
    \State \texttt{iteration} $\gets 0$
    \State \texttt{queue} $\gets \{\}$
    \State \texttt{bestNode} $\gets$ \texttt{getStartingNode()}
    \State \texttt{node} $\gets$ \texttt{getStartingNode()}
    \State \texttt{queue.add(node)}
    
    \While{$\neg$ \texttt{Empty(queue)} \textbf{and} \texttt{GetAcc(node)} $< 100$ \textbf{and} \texttt{iteration} $< \texttt{MaxIterations}$}
        \State \texttt{iteration} $\gets$ \texttt{iteration} $+ 1$
        \State \texttt{neighboursList} $\gets$ \texttt{getNeighbours(node)}
        
        \For{\texttt{neighbour} \textbf{in} \texttt{neighboursList}}
            \If{$\neg$ \texttt{Visited(neighbour)}}
                \State \texttt{queue.add(neighbour)}
                \State \texttt{Visited(neighbour)} $\gets \top$
            \EndIf
        \EndFor
        
        \State \texttt{node} $\gets$ \texttt{getNextPrioritaryNode(queue)}
        \State \texttt{queue.remove(node)}
        
        \If{\texttt{GetAcc(node)} $>$ \texttt{GetAcc(bestNode)}}
            \State \texttt{bestNode} $\gets$ \texttt{node}
        \EndIf
    \EndWhile

    \State \Return \texttt{bestNode}
\end{algorithmic}
\end{algorithm}

Additionally, we define a heuristic $h(x)$ given a solution $x$ which is equal to the sum of the incorrect predictions over the training data, plus a small fraction corresponding to the number of attacks in the graph, noted $x_\mathcal{R}$, divided by $|\args|^2$, where $\args$ is the set of all the arguments present in the dataset, or more formally: $$h(x) = \frac{|x_\mathcal{R}|}{|\args|^2} + \sum_{i \in data} \begin{cases}
      1, & \text{if } pred_i(x) \neq label_i \\
      0, & \text{otherwise}
\end{cases}$$
    
Solutions should minimise the value of this function. That way, a solution getting fewer errors than another one is systematically preferred, but in case of an equal error count, the one presenting the fewest attacks (and supports in case of bipolar argumentation) is preferred.

In order to build the graph, we follow a bottom-up tactic. The starting node has a graph composed of a single argument which is the target argument. An attack can only be added if there exists a path from the attacking argument to the target using this new attack. That way, when the search progresses, arguments and attacks are incrementally added and remain connected to the target. In order to reduce the exploration space as much as possible, we ensure that a solution is not visited twice by computing an associated hash and saving it. We then compare the candidate neighbours' hash to the ones already saved and add them to the queue only if they have not already been visited. Furthermore, if the addition of an attack had no effect on the correctness of the predictions compared to the solution without this attack, we terminate exploration of that branch. This greatly reduces the time required to explore the solution space.

\section{Evaluation}
\label{sec:evaluation_ch2}

In this section, we evaluate the algorithm proposed in Section~\ref{sec:aria}, as well as its variants. We then discuss the results obtained.

\subsection{Datasets}

In order to evaluate the approach both quantitatively and qualitatively, we used several datasets from the literature. Table~\ref{tab:datasets} summarises the size of the dataset and attributes' type and provides a short description of the content for each dataset.
The Voting, Breast Cancer Wisconsin (BCW), Heart Disease Cleveland (HDC), Iris, Wine, and Thyroid datasets can be found online at the UCI Machine Learning Repository~\cite{murphy1994uci}.\footnote{UCI Machine Learning Repository: \hyperref{https://archive.ics.uci.edu/}.} We make the classification task for the Iris dataset a binary choice by grouping the three classes as follows: The label is true if the class is ``Iris-virginica'', and false otherwise. We did a similar operation for the Wine dataset with the label being true if the class is ``1'', and false otherwise.

Furthermore, we used data from the Moral Machine experiment~\cite{awad2018moral}\footnote{Moral Machine experiment website: \hyperref{https://www.moralmachine.net/}.} to create a custom dataset. The original data contains the decision of a human annotator in an autonomous car accident scenario. The user has to choose between two outcomes, each of them causing harm to different individuals. The individual number and type (adult, kid, animal, etc) as well as their legal status (legal crossing, illegal crossing, or no legality involved) may vary.
The dataset, named MM-complete, is a pre-processed subset of the original Moral Machine data, where some attributes are either removed or grouped together (for instance, the category ``person'' grouping all the human individuals). The attributes express---using the values ``same'', ``less'', and ``more''---the difference for each attribute between the two outcomes, using the first outcome as the reference. For example, if the outcome n°2 hurts more animals than the outcome n°1, then the attribute ``animal'' will have for value ``more''. It also indicates the legality of the crossing using the values ``yes'', ``no'', and ``unspecified'' for the individuals saved in the outcome n°1 (via the attribute ``legal'') and the ones saved in the outcome n°2 (via the attribute ``legal\_alt'').

\begin{table}
    \centering
    \caption{Dataset name, size, and characteristic. Column ``Att. Types'' indicates how many attributes are continuous (c) or nominal (n). Column ``Arg.'' indicates the number of couples attribute-value (\emph{i.e.}, arguments) constructed with a segmentation into 6 intervals for the continuous attributes.}
    \begin{tabular}{|l|c|c|c|l|}\hline
        \textbf{Name} & \textbf{Size} & \textbf{Att. Types} & \textbf{Arg.} & \textbf{Description}\\\hline
        Voting~\cite{voting_105} & 435 & 16n & 34 & congressional voting records \\
        BCW~\cite{bcw_17} & 699 & 9c & 56 & medical diagnosis \\
        HDC~\cite{hdc_45} & 303 & 13c & 80 & medical diagnosis \\
        Iris~\cite{iris_53} & 150 & 4c & 26 & flower classification \\
        Wine~\cite{wine_109} & 178 & 13c & 80 & wine type classification \\
        Thyroid~\cite{thyroid_915} & 383 & 1c 15n & 65 & medical diagnosis \\
        MM-delta~\cite{alcaraz2025providing} & 200 & 5n & 17 & ethical decision making \\\hline
    \end{tabular}
    \label{tab:datasets}
\end{table}

\subsection{Quantitative Evaluation}

Table~\ref{tab:eval} shows the accuracies obtained, averaged over 10 runs, by the comparison baseline---the Support Vector Machine (SVM) classifier from the Python Scikit library~\cite{cortes1995support}---as well as several variants of the algorithm (namely the ARIA, $n$-ARIA, and Bipolar variants).
The results were collected after learning for $100$ iterations on the train data ($70\%$ of the data) and getting the accuracy over the test data (the remaining $30\%$). The dataset was reshuffled between each run of the same algorithm. The attributes with continuous values were segmented into $6$ intervals. The experiments were run with an 14th Gen Intel$^{(R)}$ Core$^{(TM)}$ i7-14700HX 2.10 GHz processor running at a 2.60GHz clock speed, with 8 physical cores and 16 logical cores, and with 32GB of memory.

\begin{table}\centering
    \caption{Accuracy of the baseline (the SVM classifier from the Python Scikit library) and the variants of ARIA on several datasets. The best value for each dataset is bolded, and the runner-up underlined.}
    \begin{tabular}{|r|c|c|c|c|}
    \hline
    \textbf{Dataset} & \textbf{SVM} & \textbf{ARIA} & \textbf{$n$-ARIA} & \textbf{Bipolar}\\ \hline
    Voting~\cite{voting_105} & 95.6 $\pm$ 1.5 & \underline{95.7 $\pm$ 1.9} & - & \textbf{95.9 $\pm$ 3.2} \\ 
    BCW~\cite{bcw_17} & \underline{95.1 $\pm$ 0.9} & 94.4 $\pm$ 1.2 & \textbf{95.6 $\pm$ 1.4} & 94.3 $\pm$ 1.4 \\
    HDC~\cite{hdc_45} & 57.8 $\pm$ 2.4 & \textbf{79.3 $\pm$ 3.0} & \underline{78.7 $\pm$ 2.4} & 70.7 $\pm$ 12.5 \\
    Iris~\cite{iris_53} & \textbf{96.0 $\pm$ 2.6} & 93.3 $\pm$ 1.5 & 90.9 $\pm$ 3.6 & \underline{93.5 $\pm$ 2.6} \\
    Wine~\cite{wine_109} & \textbf{99.1 $\pm$ 1.2} & 91.7 $\pm$ 6.4 & 92.4 $\pm$ 5.7 & \underline{96.7 $\pm$ 7.1} \\
    Thyroid~\cite{thyroid_915} & 93.5 $\pm$ 1.8 & \underline{94.3 $\pm$ 0.5} & 93.1 $\pm$ 1.7 & \textbf{97.0 $\pm$ 3.3} \\ \hline
    \end{tabular}
    \label{tab:eval}
\end{table}

As we can see, ARIA and its variants are competitive on most of the datasets, reaching accuracies close to or greater than the ones obtained by the baseline. The average of the accuracies is $91.5\%$ for ARIA, $91.1\%$ for $n$-ARIA\footnote{For the Voting dataset, the accuracy of ARIA is used.}, and $91.4\%$ for Bipolar. As we can see, those values are very close. As the standard deviation of the accuracies consistently overlaps for each of the datasets, it is not possible to confidently determine which of the approaches is the most reliable. Interestingly, Bipolar seems to perform much better than the other approaches on the Thyroid dataset, and poorly on the Heart Disease Cleveland dataset. Although we cannot explain the former, the latter seems to be mainly due to overfitting, as we could observe a drop in the accuracy of the test set after a certain number of iterations.
On the other hand, while the average runtime for $n$-ARIA was already significantly increased compared to ARIA, it seems that this is getting even longer for Bipolar, as it sometimes reaches thrice the time needed by $n$-ARIA to reach the iteration limit.

\subsection{Qualitative Evaluation of the Generated Explanations}
\label{sec:qualitative}

\begin{figure}
    \centering
    \includegraphics[width=0.8\linewidth]{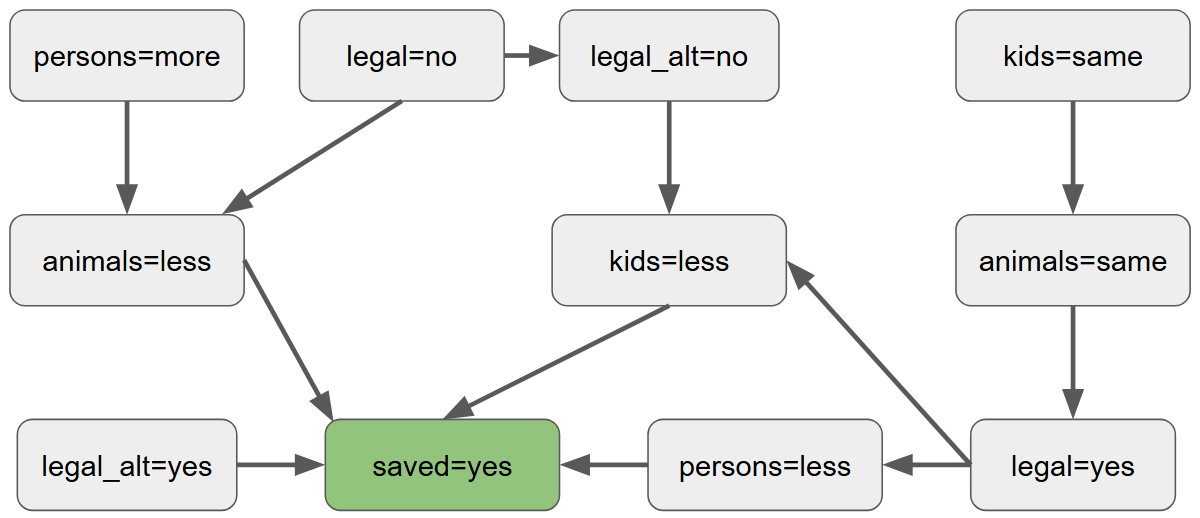}
    \caption{Universal graph for MM-delta.}
    \label{fig:qual}
\end{figure}

\begin{figure}
    \centering
    \includegraphics[width=0.7\linewidth]{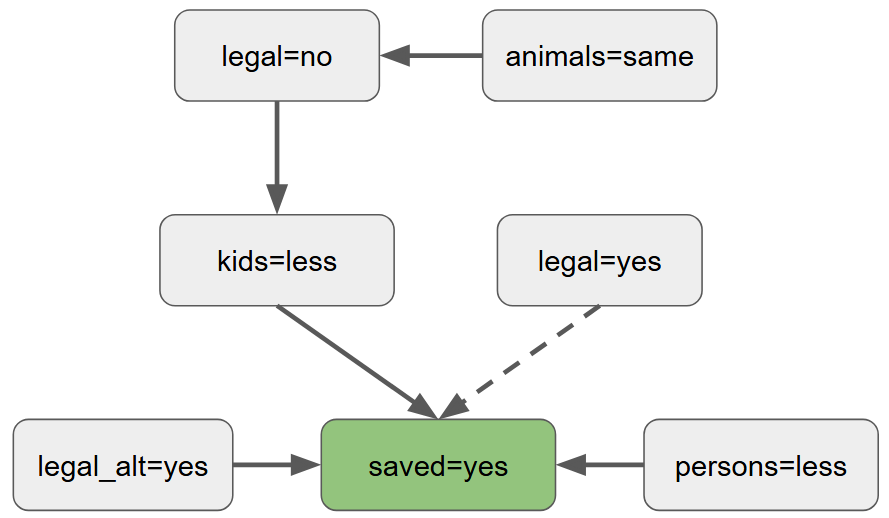}
    \caption{Bipolar universal graph for MM-delta.}
    \label{fig:qual_baf}
\end{figure}

This section aims at evaluating the relevance of the explanations provided for a given scenario when using the algorithm. More specifically, we focus on the base variant of ARIA, and the Bipolar variant. Fig.~\ref{fig:qual} shows the universal graph obtained after learning the MM-delta data set with ARIA for choosing the outcome n°1 or not. The $\target$ argument is in green. Fig.~\ref{fig:qual_baf} shows the one obtained by the Bipolar variant. Dashed arrows represent a support from one argument to another. Both graphs achieve the same accuracy of $98.4\%$ on the same test dataset. For both, the target argument is ``\texttt{saved=yes}'', which means that outcome n°1 was chosen. Note that choosing the outcome n°1 means not choosing the outcome n°2, and conversely, not choosing the outcome n°1 means choosing the outcome n°2.

\subsubsection{Scenario N°1}

\begin{figure}[ht]
    \centering
    \includegraphics[width=0.75\linewidth]{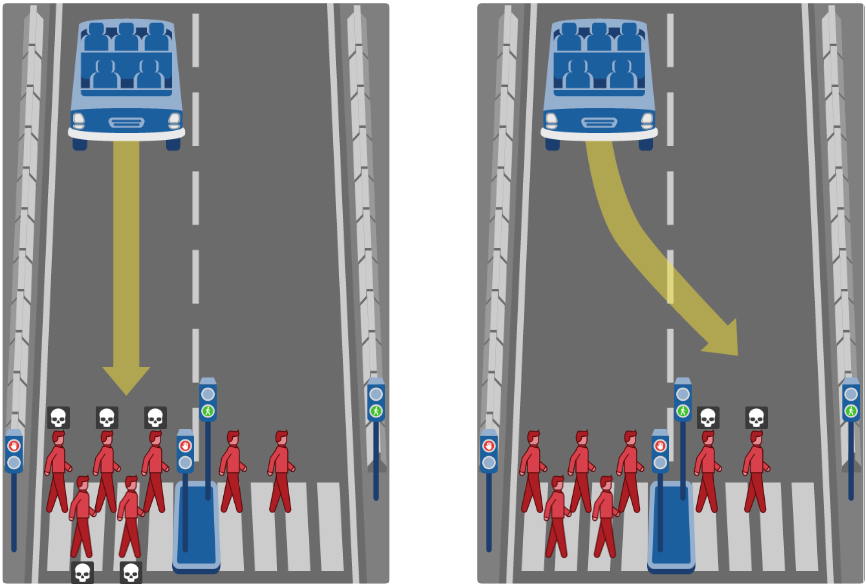}
    \caption{First scenario. Outcome n°1 is on the left, and outcome n°2 on the right.}
    \label{fig:scenario1}
\end{figure}

In the scenario represented in Fig.~\ref{fig:scenario1}, the agent has the choice between the outcome n°1, saving two persons but where the individuals saved were crossing legally, and the outcome n°2 saving five persons who were crossing illegally.

Given the set of facts represented in Table~\ref{tab:scenario1}, both graphs have in their extension the argument ``\texttt{saved=yes}'', meaning that they justify the choice of the outcome n°1. When applying the explanation generation algorithm defined in Section~\ref{sec:xaiextract} to the graph generated by the base variant of ARIA (see Fig.~\ref{fig:qual}), one possible explanation generated is $DefBy(\texttt{saved=yes}) = \{ \texttt{saved=yes}, \texttt{legal=yes}, \texttt{kids=same}\}$. This explanation may be interpreted as despite the fact that some elements were favourable to the choice of the outcome n°2 (such as the number of people hurt), the legal aspect of the situation is in favour of the people spared in the outcome n°1, and there is no difference in the number of children involved to alter this decision.

\begin{table}[ht]
    \centering
    \caption{Summary of the facts for the scenario n°1.}
    \begin{tabular}{|rcl|}\hline
        \textbf{Attribute} &   & \textbf{Value}\\\hline
        Persons & = & less\\
        Kids & = & same\\
        Animals & = & same\\
        Legal & = & yes\\
        Legal\_alt & = & no\\\hline
    \end{tabular}
    \label{tab:scenario1}
\end{table}

On the other hand, when applying the explanation generation to the graph generated using bipolar argumentation (see Fig.~\ref{fig:qual_baf}), $DefBy(\texttt{saved=yes}) = \{ \texttt{saved=yes}, \texttt{legal=yes}\}$. The explanatory set contains one less element. However, as one can see, the structures of the graphs are different, and in the case of the Bipolar graph, nothing can override the fact that the individuals spared in the outcome n°1 were crossing legally. We believe this is a limitation of the explanation generation process with bipolar argumentation. Indeed, it seems not to communicate any difference in the meaning that can be conveyed by a defence through an attack and a defence through a support.

\subsubsection{Scenario N°2}

\begin{figure}[ht]
    \centering
    \includegraphics[width=0.75\linewidth]{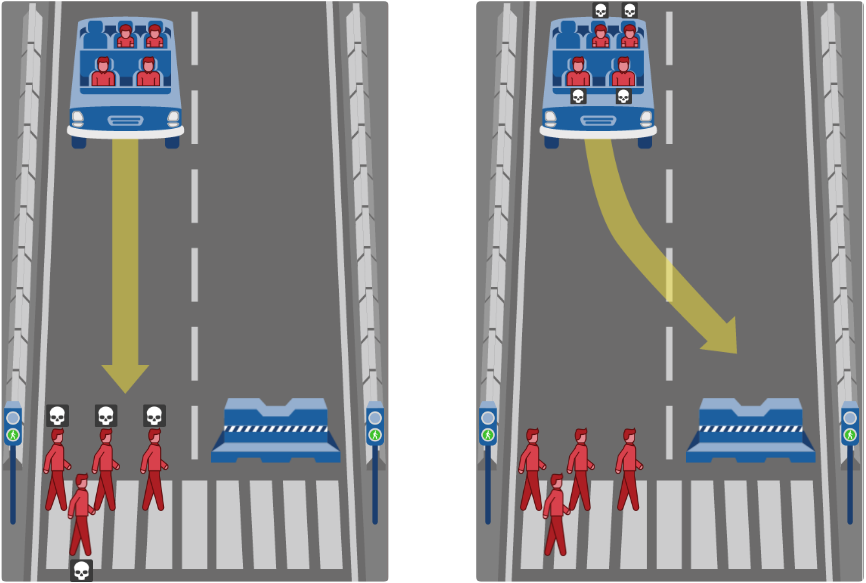}
    \caption{Second scenario. Outcome n°1 is on the left, and outcome n°2 on the right.}
    \label{fig:scenario2}
\end{figure}

In this scenario, represented by Fig.~\ref{fig:scenario2}, the agent has the choice between the outcome n°1, saving two less kids, and the outcome n°2 saving two more kids but where individuals who perished were crossing legally. The facts for this scenario are summarised in Table~\ref{tab:scenario2}.

\begin{table}[ht]
    \centering
    \caption{Summary of the facts for the scenario n°2.}
    \begin{tabular}{|rcl|}\hline
        \textbf{Attribute} &   & \textbf{Value}\\\hline
        Persons & = & same\\
        Kids & = & more\\
        Animals & = & same\\
        Legal & = & unspecified\\
        Legal\_alt & = & yes\\\hline
    \end{tabular}
    \label{tab:scenario2}
\end{table}

In this scenario, none of the graphs has in its extension the argument ``\texttt{saved=yes}'', meaning that they justify the choice of not choosing the outcome n°1 (hence choosing the outcome n°2). When applying the explanation generation algorithm to the graph generated by both the base and bipolar variants of ARIA, we obtain the explanation $NotDef(\texttt{saved=yes}) = \{ \texttt{saved=yes}, \texttt{legal\_alt=yes} \}$. This justification set represents that \texttt{legal\_alt=yes} is an argument in the (grounded) extension attacking the target, for which there is no defence within the extension. This explanatory set suggests that the fact that the individuals who would have perished if the first outcome had crossed legally is a sufficient reason to spare them (and so choose the outcome n°2).

\section{Related Work}
\label{sec:sota_ch2}

As Section~\ref{sec:normmining} already provides a detailed overview of the field of norm identification, this related work section will focus solely on rule induction approaches.

In recent years, deep neural networks have been shown to be very successful in solving classification problems. However, these algorithms suffer from a lack of explainability~\cite{szegedy2013intriguing}.

A solution to make a system understandable is to apply decision tree classification approaches. The algorithm C4.5~\cite{quinlan2014c4}, based on ID3~\cite{quinlan1986induction}, has been developed following this paradigm. However, these approaches have been shown to tend to be outperformed in many domains by rule induction algorithms~\cite{bagallo1990boolean,quinlan1987generating,weiss1991reduced}.

Rule induction algorithms are a category of approaches that usually tries to generate a set of rules for each decision or class. Then, if an input triggers one of the rules for a given class, it is classified as such. There exists a variant of C4.5, called C4.5-$rules$, on which many approaches were based. For example, IREP~\cite{furnkranz1994incremental} has been introduced to accommodate the issues of C4.5 related to its computation time by making the pruning more efficient. However, it usually produces more errors than C4.5 with domain-specific datasets. For this reason, Cohen developed an improved version called RIPPER$k$~\cite{cohen1995fast} which is simultaneously more efficient, more accurate, and more resilient to noisy data.
A rule induction algorithm based on genetic programming, SIA~\cite{venturini1993sia}, generates a population of rules and compares the predictions performed with an actual dataset. The algorithm has to maximise the correct predictions.
The algorithm ESIA~\cite{liu2000extended} (Extended SIA) is an extension of SIA. While the principle remains the same, ESIA contains modifications to the operators of SIA.

In parallel, much work has been done in explainability, especially with respect to new Deep Reinforcement Learning (DRL) algorithms presenting high performance but poor interpretability~\cite{adadi2018peeking}. In the next years, such a capability could become an obligation more than an option with upcoming legislation regarding the right to explanation~\cite{selbst2018meaningful}.
We divide these works into two categories. The first concerns algorithms that have intrinsic intelligibility, which means that the algorithm itself provides information for its understanding. The second one concerns algorithms producing post-hoc explanations~\cite{puiutta2020explainable}, meaning that we can apply these algorithms to various AI models to extract information.

\citet{verma2018programmatically} propose an intrinsic explainability method by presenting a reinforcement learning agent that produces interpretable policy rules. By representing the states in a high-level manner, it can express the rules determining the action to perform, \emph{i.e.}, the policy to follow.
Even though this work is competitive with DRL algorithms when working with symbolic inputs, it cannot handle perceptual inputs, such as pixels of an image, or stochastic policies, useful within game-like environment.
Moreover, policies generated in this work remain hard to grasp especially for a non-expert user, due to the large amount of numerical values present in the rules, which greatly decreases intelligibility.
Additionally, the replacement of black-box models by newer intrinsically explainable methods might be either not applicable or too expensive for a company. This might limit the spread of such a method.
Lastly, it is often the case that those explainable methods show slightly lower performance than black-box models. Thus, it forces a trade-off between performance and explainability.

On the other hand, post-hoc methods for explainability~\cite{puiutta2020explainable,hein2018interpretable,adadi2018peeking} aim at generating from reinforcement learning policies a set of equations that describes trajectories of the analysed policy. These outputs are presented with a certain amount of complexity in terms of explainability. Yet, authors admit that, even if pretty low, complexity equation sets allow good performance in addition to showing some explainability relief when compared to Neural Network approaches, they still under-perform it in terms of pure performance. Moreover, the equation system may start to become difficult to understand because of the abstract nature of some thresholds. Also, because this algorithm is computing rules of trajectories, it may struggle in highly discretised environments such as the ones with categorical inputs. Furthermore, this work and others presented in \citet{puiutta2020explainable} such as \citet{liu2018toward} or \citet{zahavy2016graying} are not agnostic to the learning algorithm for which they provide explanations and need access to the agent's policy.
Another post-hoc approach is the counterfactual explanation which consists of giving bits to the end user to help him understand what the machine is doing by presenting slight input variations to obtain different outputs~\cite{wachter2017counterfactual}. The problem of such a method is that it leaves the responsibility to the end user to make suppositions on what impacts the model's decision and what does not. Furthermore, it is not a scalable methodology at all.
\citet{tan2018distill} presented a model distillation. This method transfers knowledge from a large and complex model (teacher) to a faster and simpler model (student). However, even if it can successfully extract data from black-box algorithms, the problem of the explainability of the extracted data remains.

Last, the algorithm PSyKE~\cite{sabbatini2021design} is the closest to our approach in terms of application. In this work, the authors use a rule induction algorithm to generate Prolog-like rules based on the classifications of a given model, such as CART, GridEx, or k-nn. Although it tends to underperform the initial model, this is not a major issue as it only aims at providing a justification for the decisions taken by the classification model by giving the set of rules leading to this aforementioned decision. As such, they define a value called black-box fidelity, which shows how accurately the generated rules mimic the classification of the black-box model.

\section{Summary}
\label{sec:summary_ch2}

The goal of this chapter was to enable the possibility to automatically obtain the norms and their potential exceptions in the form of an argumentation graph. This can be broken down into a two-step process: ($1$) Detecting the norms active within an environment, and ($2$) Identifying the exceptions to these norms.

In Section~\ref{sec:normmining}, we provide a comprehensive review of the norm identification approaches developed over the past fifteen years. Through the analysis of each approach with respect to a list of established research questions, we identified some key takeaways, allowing the exclusion of several of the collected approaches, so that a user of the \pino architecture can make an informed decision and select a suitable norm mining technique.

Then, in Section~\ref{sec:aria}, we propose the ARIA algorithm that aims at generating an argumentation graph that represents the exceptions to a classification. Then, we conducted a quantitative study to evaluate its classification performance and a qualitative study to ensure that the generated graph was coherent and enabled meaningful explanation generation. Combining this algorithm with a norm mining method would allow for the gathering of norms and their exceptions in the form of argumentation graphs.

%% file: sections/4_norm_avoidance.tex
In this chapter, we introduce the concept of \emph{Norm Avoidance}. This phenomenon has been observed through the empirical evaluation of the \pino architecture~\cite{alcaraz2026pinocchio} introduced in Chapter~\ref{cha:pino}. Here, we first define this phenomenon and provide several examples to illustrate it. Then, we present a method which aim to mitigate this issue for the specific case of reinforcement learning.

\section{Introduction}

Recent progress in the field of artificial intelligence (AI) has opened the door to the integration of a variety of intelligent systems in our day-to-day lives.
The integration of these technologies creates the need for reliable normative agents (agents that not only perform a task autonomously but also operate under a set of norms to ensure compliance with social or organisational expectations) since deviant behaviour may result in malfunctions, sub-optimality, or even harm to people sharing their environment. This need introduces the challenge of accurately modelling norms and designing agents capable of taking these norms into account---normative agents---but from the inclusion of this capability new problems arise. For instance, these norms may come into conflict, either with one another or with the goals of the agent. One example is an autonomous vehicle that must not cross a solid line, but encounters an obstacle in its path. Additionally, these norms may be defeasible. For instance, an autonomous vehicle should avoid driving over the pavement, but it may need to do so if a fire truck behind it needs to pass.

In this chapter, we are interested in the issue of \textit{norm avoidance}~\cite{alcaraz2025norm}, the phenomenon where an agent formally adheres to a norm but circumvents its intended purpose by exploiting loopholes.  For instance, the agent might deliberately trigger an exception to a norm as a way of achieving its goals without violating the norm directly. This can potentially affect norm alignment, as the agent may perform actions that deviate from the expected behaviour. This phenomenon was discussed and analysed in theoretical studies in the field of deontic logic in the 1990s~\cite{sergot1994contrary,prakken1996contrary} (as in Sergot and Prakken's Cottage House example), and has frequently been discussed ever since~\cite{van1997reasoning,van1997cancelling,van1998deliberate}, most notably with the introduction of the notion of controllability of some states by the agent.

\textit{Norm avoidance} is frequently observed in real life. For example, the COVID-19 lockdown saw an increase in dog adoptions worldwide, attributed in part to the desire to bypass restrictions on going outside~\cite{ho2021did,morgan2020human}.
Though technically compliant, such behaviour does not respect the spirit of the initial norm. On the other hand, if a norm is rendered inapplicable by an external cause (\emph{i.e.}, a cause which is independent of the actions of the agent) or for reasons other than a desire to circumvent the norm, it may be preferable that the agent bypasses the norm after all. For instance, if the agent adopted the dog before the lockdown, or if their desire to adopt a dog was independent of their desire to go outside, there would be no issue with them having the dog and taking it for walks.

Examples of \textit{norm avoidance} have been observed recently in the field of normative or ethical reinforcement learning~\cite{neufeld2022enforcing}. Reinforcement learning (RL) is a technique wherein an autonomous agent explores a given environment and learns behaviour that maximises the rewards it gets from that environment over a period of time; it has been identified as a promising technology for teaching autonomous agents how to behave in a way that complies with norms. Since \citet{AML2016} work, there have been many approaches proposed to facilitate the learning of behaviour(s) that comply with a given set of norms or values. However, normative RL agents are prone to \textit{norm avoidance}, similar to how regular RL agents are prone to reward hacking (though these are two separate phenomena, and could appear independently or simultaneously). Reward hacking occurs when optimisation of an imperfect proxy reward function leads to poor performance compared to an unknown true reward function that would successfully incentivise the desired behaviour~\cite{skalse2022defining}.
Reward hacking is hard to predict, evidenced mainly by the appearance of unexpected behaviours during the experimental phase~\cite{amodei2016concrete,aslund2018virtuously,orseau2016safely}. However, because work on designing normative RL agents is still at a relatively early stage, \textit{norm avoidance} has remained mostly undetected and unaddressed by the literature, even though it has been observed in the work of \citet{neufeld2024learning}, where the agent committed violations when facing normative deadlocks that could have been avoided. Analogously to reward hacking, we believe it could lead to undesirable situations that could potentially cause harm to humans.

One of the examples in \citet{neufeld2022enforcing} consists of an agent (\emph{i.e.}, a vegan pac-man) evolving among ghosts. It is forbidden to eat the ghosts, but one exception exists, which is that if the agent is next to an orange ghost, it may eat it in order to protect itself. In such a state, eating an orange ghost is no longer treated as a violation. In the meantime, eating ghosts grants an additional reward to the agent. It is clear in this scenario that the agent has an interest reward-wise in going next to the orange ghosts. However, one can see that this goes against the intent of the norm which is to avoid eating ghosts, and only do it as a last resort.

\textit{Norm avoidance} is difficult, if not impossible, to prevent in human agents. For one thing, human agents can lie about their intentions---they could claim they wanted to have a dog despite only wanting to go outside. Moreover, it may not be possible or fair to try to legislate away the possibility of \textit{norm avoidance}; e.g., such a law would likely prevent genuine dog lovers from adopting dogs. In any case, human agents have a right to adopt a dog if they want to, even if it is merely in the interest of allowing them to go outside. On the other hand, artificial agents cannot have insincere or instrumental intentions. What is more, they have no rights---they should behave not only permissibly, but also in an ideal way.

The goal of this paper is to propose a preliminary computational definition of \textit{norm avoidance} in RL agents, and various means of mitigating it, based on previous approaches to normative reinforcement learning.

\section{Norm Avoidance: Concept and Definitions}
\label{sec:definitions_ch3}

In this section, we provide a formalised presentation of \textit{norm avoidance}---note that we have constructed only a preliminary definition here, tailored to our RL setting.

\begin{definition}[Norm Avoidance]
    We call a transition $(s,a,s')$ \textit{norm avoidant} if a norm $n$ is defeated, the transition is non-compliant, and at least one of the following conditions relating to an agent's \textit{goal states} (that is, a state in which a crucial goal is achieved; more on these below) are not met:
    
    \begin{itemize}
        \item[1.] $n$ has been defeated by external causes\footnote{A concrete notion of external cause is defined later in Def~\ref{def:external}.}
        and there is no alternative path to one of the agent's goal states which complies with $n$.
        \item[2.] The norm has been defeated on a fully compliant path to one of the agent's goal states and there is no alternative compliant path that does not defeat the norm.
    \end{itemize}
\end{definition}

The main idea behind this definition is that the agent should not be the one responsible for the defeat of a norm which is later not complied with. And in the event this occurs, it should be due to legitimate intentions, such as achieving another task that could not be achieved in another way than defeating the norm. Last, even if the agent is not the one responsible for the defeat of the norm, it should do its best to comply with the norm as far as possible. Consequently, the existence of an alternative fully compliant path to the goal should be favoured.

The question then becomes one of how to ensure that both these conditions are met by an RL agent. Our approach entails strategically equipping the agent with additional information as to whether it defeats/violates/complies with applicable norms, and which transitions reach goal states. We will discuss how to learn functions conveying this information in following subsections.

\section{Mitigation Strategies for Norm Avoidance in Reinforcement Learning}

In this section, we introduce several approaches that aim to limit the risks of \textit{norm avoidance} occurrence in the specific context of normative reinforcement learning. Then, we evaluate these approaches over multiple environments covering the various cases of \textit{norm avoidance}.

\subsection{Preliminary Definitions}
\label{sec:definitions}

This section contains definitions for elements commonly used in the proposed methods. First, we enrich our MDP with additional functions.\footnote{It can be considered to some extents as an extension of a NMDP as proposed by \citet{fagundes2016design}.}

\begin{definition}
    Let an enriched version of a Markov decision process be $\langle S, A, P, R, N, C, D, G\rangle$ where:
    \begin{itemize}\tightlist
        \item $N$ is a set of norms
        \item $C: S \times Act \times S \times N\rightarrow \{0, -1\}$ evaluates compliance with a given norm when the agent is transitioning from state $s$ to state $s'$ with action $a$; that is, it outputs $0$ in the case of compliance, and $-1$ otherwise
        \item $D: S \times Act \times S \times N\rightarrow \{0, -1\}$ indicates the defeat of a given norm when the agent is transitioning from state $s$ to state $s'$ under action $a$, outputting $0$ when the norm is not defeated, and $-1$ otherwise
        \item $G: S \times Act \times S \times Obj \rightarrow \{0, 1\}$ indicates whether the goal $o \in Obj$ (where $Obj$ is simply a set of goal labels) will be reached when the agent is transitioning from state $s$ to state $s'$ with action $a$; we refer to state $s'$ as a \textit{goal state}. A state can be labelled as a goal in relation to various objectives (\emph{i.e.}, tasks that the programme designer expects the agent to accomplish). Note that if a goal can be achieved multiple times, it should be divided into multiple objectives (\emph{i.e.}, if we can achieve a particular $goal$ three times, there should be $goal_1,goal_2,goal_3\in Obj$).
    \end{itemize}
    
\end{definition}

As with a standard MDP, it is possible to compute $Q$-functions for each of the above reward functions. We then define updates for the functions $Q_R$, $Q_D$, $Q_C$, $Q_G$ for a given state $s \in S$, an action $a \in A(s)$, an optimal (off-policy) action $a' \in A(s')$ in the resulting state $s'$, and a norm $n \in N$. Using functions $C$ and $D$, it is also possible to construct the function $$V(s, a, s', n) = min(0, C(s, a, s', n) - D(s, a, s', n))$$ in order to compute the expected violations $Q_V(s, a)$ later on. 
$Q_R$ is updated as usual with $U_1$ during the $Q$-learning algorithm. $Q_G$---which has a separate value for each objective $o \in Obj$---uses the update rule $U_2$, which takes a different set of learning parameters (namely $\gamma = 1$, made possible by our finite horizon assumption). Finally, for $x\in \{D,C,V\}$, the function $Q_x$ is updated using a modified version ($U_3$). $U_3$ does not accumulate the expected value over the path (\emph{i.e.}, it will not converge to Eq.~\ref{eq:q}) but instead transmits the minimal value (since our task is to maximise the values of the relevant $Q$-functions).
We do this as the signal for a defeated norm is returned every time step as long as the norm remains defeated. As a result, a path reaching a final state of the MDP, thus ending the simulation epoch, would give an expected value of the defeat of the norm proportional to the length of the path. As such, two paths of distinct length would have different expected value, rendering the agent ``responsible'' if it chooses the longer path, even though this is not the case.
These three updates are described below, where the learning rate is $\alpha\in [0, 1]$ and the next state is $s'$:

\begin{itemize}\tightlist
    \item $U_1$: $Q_R(s, a) \coloneqq Q_R(s, a) + \alpha(R(s, a) + \gamma \max_{a' \in A(s')} Q_R(s', a') - Q_R(s, a))$

    \item $U_2$: $Q_G(s, a, o) \coloneqq Q_G(s, a, o) + \alpha(G(s, a, s', o) + \max_{a' \in A(s')} Q_G(s', a', o) - Q_G(s, a, o))$
    
    \item $U_3: Q_x(s, a, n)$:=$Q_x(s, a, n)$+$ \alpha(\min(x(s, a, s', n), \max_{a' \in A(s')} Q_x(s', a', n)) - Q_x(s, a, n))$ 
\end{itemize}

\begin{remark}
    $Q_V$ can be updated using, as opposed to $U_3$, a variant of $U_2$ which modifies the function signature in order to integrate the norm $n$, provided that the violations are sequential or have a duration that we want to minimise.
\end{remark}

We then define $Q_G^\Sigma$ (representing how many goal states in total are expected to be reached) as well as $Q_x^\Sigma$ for  $x \in \{V, C, D\}$\footnote{If one wishes to preserve priorities among the norms, it is possible to weight them or, alternatively, keep the $Q$-functions separate in the lexicographic selection (introduced in Section~\ref{sec:definitions}).}:
$$Q_G^\Sigma \coloneqq \sum_{o \in Obj}Q_G(s, a, o) \text{   and   } Q_x^\Sigma(s, a) \coloneqq \sum_{n \in N}Q_x(s, a, n)$$
\label{sigma}
We also want to represent the notion of an agent's responsibility for a norm's defeat (\emph{i.e.}, a notion of whether the norm was defeated by external causes or by the agent).

\begin{definition}\label{def:external}
     We say that an agent is responsible for the defeat of a norm if it performs an action that does not maximise $Q_D$ (\emph{e.g.}, the action increases the probability that the norm will be defeated in the future when compared to other possible options available to the agent). Formally, for a given norm $n$ and an action $a$ performed by the agent in state $s$, if it is true that \begin{equation}
         \exists x \in A(s) \emph{   s.t. } Q_D(s, x, n) > Q_D(s, a, n)
         \label{exp}
     \end{equation} then the agent is held responsible for the defeat of the norm.\footnote{When there is no action leading to a lesser expected value for the defeat of the norm, it can be seen as a lack of agency, which is a reason to consider the agent not responsible~\cite{braham2012anatomy}.}
     We define a function $D_r:S\times Act\times N\to \{-1, 0\}$ that returns -1 when the agent is responsible for the defeat of a norm (in the sense of Expr.~\ref{exp}) and 0 otherwise.  The function furthermore propagates this value forward so that if an agent defeats a norm during runtime, $D_r$ ``remembers'' that for future reference.
     \label{defn:responsible}
\end{definition}

This definition allows us to compute $C_{\dagger}$ (and as a consequence $Q_{C\dagger}$ and $Q_{C\dagger}^\Sigma$), which differs from $C$ in that it considers non-compliance with a norm only if the agent was responsible for that norm's defeat; \emph{i.e.}, $C_{\dagger}(s,a,s',n)=-1$ when $Q_C(s,a) < 0$ and $D_r(s,a, n)=-1$. In order to learn the function $Q_{C\dagger}$, we have to train the agent with $U_3$ and $C_{\dagger}$ after learning the functions $Q_C$ and $Q_D$.

\subsection{Proposed Approaches}
\label{sec:approaches}

This section proposes a variety of approaches that aim to mitigate \textit{norm avoidance} as described above. Here, we only survey the technical composition of the approaches presented; for a discussion about their actual behaviours, see Section~\ref{sec:evaluation_ch3}. One of the main constraints we would like to emphasise is that any expert knowledge (\emph{e.g.}, which transitions lead to a goal state---something which would have to be defined explicitly using knowledge of the environment) should be used in a limited way so that it can be realistically scaled or possibly computed by an external algorithm (see sub-goal discovery in Hierarchical RL~\cite{gurtler2021hierarchical,pateria2021end,liu2021hierarchical}). For this reason, we exclude the possibility that the agent knows what the outcome state for a given action will be; this also removes the possibility of the agent knowing whether performing a given action in a given state will result in the violation, non-compliance, or defeat of any norm. Furthermore, we will not consider the cases in which a malicious agent lies about its intentions because it would not be possible in practice for an agent to do this when using one of our approaches.

\subsubsection{Lexicographic (Lex)} 

This approach only requires the $Q$-functions $Q_V$, $Q_{C\dagger}$, and $Q_R$. The lexicographic selection procedure first maximises the $Q$ value for the violations\footnote{Since a violation yields $-1$, and its absence counts as $0$, maximizing $Q_V^\Sigma$ is the same as minimizing the expected violation count. The same goes for $Q_C^{\Sigma}$, $Q_{C\dagger}^\Sigma$, and $Q_D^{\Sigma}$.}, then it maximises the $Q$-value that corresponds to non-compliance with the norms defeated by the agent, and finally it maximises the reward; expressed formally as: $$Q_V^\Sigma \succ Q_{C\dagger}^\Sigma \succ Q_R$$

Note that this approach does not require that the MDP contains an element $G$ indicating which states are goal states, which is a significant reduction in the expert knowledge required.

\subsubsection{Lexicographic Compliant (Lex-C)}

This approach is similar to Lex, but it contains two more $Q$-functions, namely $Q_G^\Sigma$ and $Q_C^\Sigma$, which were already introduced in Section~\ref{sec:definitions}. It also consists of a new ordering for the lexiographic selection which is $$Q_V^\Sigma \succ Q_{C\dagger}^\Sigma \succ Q_G^\Sigma \succ Q_C^\Sigma \succ Q_R$$
where $Q_G^\Sigma$ and $Q_C^\Sigma$ are as defined in Section~\ref{sec:definitions}. Although Lex is sufficient to limit basic cases of \textit{norm avoidance}, it cannot handle situations where there exists a second fully compliant path to the objective. Lex-C is intended to fix this problem by ensuring that if it is possible to reach a goal state, then the agent should prioritise the most compliant path leading to that state.

\subsubsection{Lexicographic Opportunist (Lex-O)}

Let the set of actions that maximise the expectation of reaching a goal state for a given objective $o \in Obj$ (only non-empty if a goal state can be reached while minimizing violations) be: $$Opt(s, o) \coloneqq \argmax_{x \in A(s)}Q_G(s, x, o) \cap \argmax_{x \in A(s)}\sum_{n \in N}Q_V(s, x, n)$$

Then let the set of actions in set $Opt(s, o)$ that are not expected to lead to non-compliance with a norm for whose defeat the agent is responsible be: $$Comp(s, o) \coloneqq \{x \in Opt(s, o) | \forall n \in N. Q_{C\dagger}(s, x, n) = 0\}$$

We can then define the predicate $P(s, a)$ over state-action pairs as $$P(s, x) \coloneqq \exists o \in Obj.(x \in 
Comp(s, o))$$
which is true if and only if there exists an objective $o\in Obj$ such that the action $a$ in question is optimal for reaching the goal state  corresponding to $o$ (\emph{i.e.}, one of the the states $s'$ where $G(s, a, s', o)=1$) and avoids violations, where there is no expected non-compliance for any norm that the agent is considered to be responsible for defeating.
As with $D_r$, in practice we configure the agent to ``remember'' $P(s,a)$ for future transitions.
We now define the new function: $$C_\ddagger(s, a, s', n) \coloneqq \begin{cases}
    0\textit{ if } P(s, a) \textit{ is true,}\\
    C_\dagger(s, a, s', n)  \textit{ otherwise}
\end{cases}$$
which only counts non-compliance to a norm with respect to a transition $(s,a,s')$ if it occurs when the agent was responsible for the defeat of the norm---except when, at some time in the past, the agent took a path which both optimised for reaching a goal $o\in Obj$ and minimised violations, and in this path, full compliance was expected to those norms which the agent was responsible for defeating.
Similarly to $D_r$, we can configure the agent to ``remember'' $P(s,a)$ for future transitions.

Finally, we define the lexicographic ordering as $$Q_V^\Sigma \succ Q_{C\ddagger}^\Sigma \succ Q_G^\Sigma \succ Q_C^\Sigma \succ Q_R$$

\subsection{Evaluation}
\label{sec:evaluation_ch3}

\subsubsection{Examples}
\label{sec:cases}

This section introduces several benchmark examples which will later serve to compare the approaches proposed in Section~\ref{sec:approaches}. Each example can be represented by a deterministic MDP.

We define three parameters---which are intuitively relevant factors according to Section~\ref{sec:definitions_ch3}---to help us
categorise our benchmark examples for a given norm $N$ and a given goal $G_1$:
\begin{itemize}\tightlist
    \item \textbf{Self:} True (T) if the norm $N$ was defeated by the agent. False (F) if defeated by an external source.
    \item \textbf{Alt.:} True (T) if there was an alternative path that is fully compliant with $N$ leading to goal state $G_1$. False (F) otherwise.
    \item \textbf{Obj.:} True (T) if there is a second goal state $G_2$, with a path that is fully compliant with norm $N$, but taking such a path would defeat $N$. False (F) otherwise.
\end{itemize}

\begin{table}[ht]
    \caption{Enumeration of possible parameter settings. \emph{Self} indicates that the norm is defeated by the agent or by an external source. \emph{Alt.} indicates that there is a fully compliant path to the goal state. \emph{Obj.} indicates that there is a second goal, and that reaching that goal would force the agent to reach a state which defeats the norm that limits access to the first goal.}
    \centering
    \begin{tabular}{c|ccc|c}\hline
        \textbf{Index} & \textbf{Self} & \textbf{Alt.} & \textbf{Obj.} & \textbf{Case}\\ \hline
        1 & F & F & F & 1\\
        2 & F & F & T & -\\
        3 & F & T & F & 2\\
        4 & F & T & T & -\\
        5 & T & F & F & 3\\
        6 & T & F & T & 4\\
        7 & T & T & F & 5\\
        8 & T & T & T & 6\\ \hline
    \end{tabular}
    \label{tab:cases}
\end{table}

\begin{figure}
    \centering
    \includegraphics[width=0.5\linewidth]{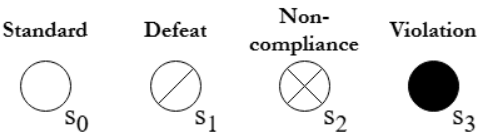}
    \caption{$s_0$ represents the resulting state of an agent's compliant transition $(s,a,s_0)$; $s_1$ is a state resulting from the transition $(s, a, s_1)$ where the norm was defeated for the first time (a defeat state); $s_2$ is a state where a norm has not been complied with in the preceding state-action-state transition $(s,a,s_2)$ (a non-compliance state); and $s_3$ indicates a violation of the norm in the preceding transition $(s, a, s_3)$ (a violation state).}
    \label{fig:legend}
\end{figure}

From Table~\ref{tab:cases}, which summarises the setting of each parameter, we discard settings $2$ and $4$, as the norm is already defeated by an external source, so there cannot be another objective which would lead to the defeat of the norm on its path. Accordingly, each line is re-indexed under a certain case index. We then create the environments represented by the figures below (see Fig~\ref{fig:legend} for the meaning of each symbol). Goal states for various objectives are labelled with a $1$ or $2$ (depending on the objective). These objectives are not cumulative, meaning that reaching a goal state several times to achieve the same objective does not grant any additional value. Also, once the norm is defeated in the environment, it remains defeated in all subsequent states (\emph{i.e.}, only the state where the defeat first happened is noted as a defeat state\footnote{The environments we utilise in this paper are fully deterministic, which means that for each state-action pair $(s,a)$, there is exactly one state $s'$ which is transitioned into with probability 1; therefore, what we indicate in our diagrams as a defeat/non-compliance/violation state is the single state $s'$ transitioned into in the defeating/non-compliant/violating transition $(s,a,s')$.}).

\begin{figure}[htbp]
    \centering
    \begin{minipage}{0.45\linewidth}
        \centering
        \includegraphics[width=0.5\linewidth]{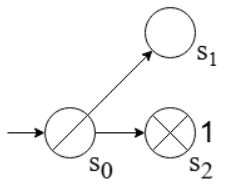}
        \\\small\justifying\noindent
        Case 1: The norm is defeated by an external source. The agent can choose not to comply in favour of fulfilling its objective, or do nothing.
        \label{fig:ex1}
    \end{minipage}
    \hspace{0.03\linewidth}
    \begin{minipage}{0.5\linewidth}
        \centering
        \includegraphics[width=0.65\linewidth]{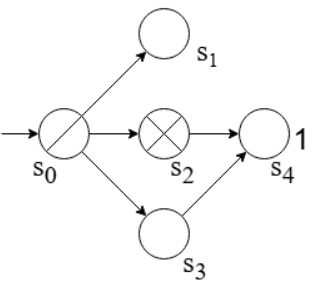}
        \\\small\justifying\noindent
        Case 2: The norm is defeated by an external source. The agent can choose not to comply in favour of fulfilling its objective, do nothing, or choose an alternative fully compliant path to reach its objective.
        \label{fig:ex2}
    \end{minipage}\label{allex}
    
    \vspace{0.1cm}
    
    \begin{minipage}{0.45\linewidth}
        \centering
        \includegraphics[width=0.8\linewidth]{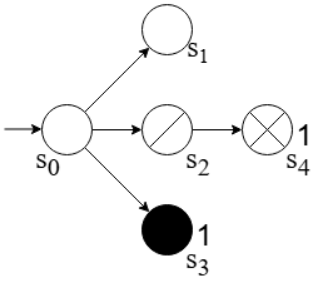}
        \\\small\justifying\noindent
        Case 3: The agent can choose between not reaching its objective, defeating the norm in order to reach its objective, or violating the norm.
        \label{fig:ex3}
    \end{minipage}
    \hspace{0.03\linewidth}
    \begin{minipage}{0.5\linewidth}
        \centering
        \includegraphics[width=0.95\linewidth]{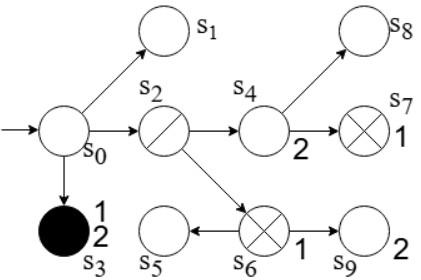}
        \\\small\justifying\noindent
        Case 4: The agent can choose between not reaching its objective, defeating the norm in order to reach either one or more objectives, or violating the norm.
        \label{fig:ex4}
    \end{minipage}
    
    \vspace{0.3cm}
    
    \begin{minipage}{0.42\linewidth}
        \centering
        \includegraphics[width=0.98\linewidth]{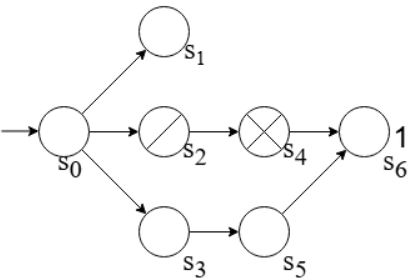}
        \\\small\justifying\noindent
        Case 5: The agent can choose between fulfilling its objective by defeating a norm and subsequently not complying with it, or choosing an alternative compliant route.
        \label{fig:ex5}
    \end{minipage}
    \hfill
    \begin{minipage}{0.53\linewidth}
        \centering
        \includegraphics[width=0.98\linewidth]{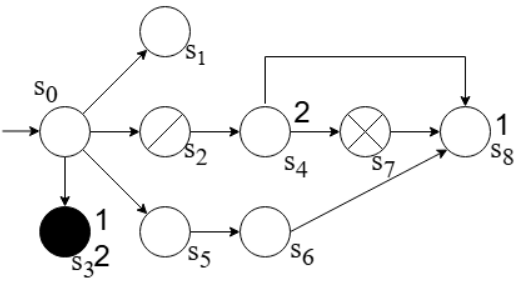}
        \\\small\justifying\noindent
        Case 6: The agent can fulfil its objective by defeating a norm and subsequently either not complying with it or branching to an alternative compliant path, or it can choose an alternative compliant route from the start.
        \label{fig:ex6}
    \end{minipage}
    
    \vspace{0.1cm}
\end{figure}

It is important to note that each case is not a partition of a larger environment, but rather a representative example of an environment characterised by the conditions in Table~\ref{tab:cases} in its most compact form (with respect to a single goal). Secondly, we have discarded certain states that are irrelevant from a reinforcement learning perspective since they are strictly suboptimal. For example, an agent choosing not to reach a goal state after failing to comply with a norm is strictly worse than an agent simply choosing not to reach the goal state from the outset. Finally, while some paths may appear longer visually (and may seem to have lower utility at first glance), our proposed approaches prioritise mitigating \textit{norm avoidance}. As such, each example accounts for cases where an alternative route may have higher or lower utility.

We illustrate each case with an example inspired from real life. For each example, we map actions to states that would correspond to their respective outcomes.

\paragraph{Case 1: Autonomous car, obstacle, and no workaround:}
The agent needs to reach a target destination, but an obstacle is blocking the road. Driving on the pavement is prohibited unless the road is completely blocked (\emph{i.e.}, $\texttt{not } blocked\to\textbf{F}(pavement)$). There is no alternative route. The agent can choose between not reaching its goal ($s_1$) or driving on the pavement ($s_2$).

\paragraph{Case 2: Autonomous car, obstacle, and workaround:}
The agent needs to reach a target destination, but an obstacle is blocking the road. Driving on the pavement is prohibited unless the road is completely blocked. However, an alternative route exists, though it may be less efficient because of the longer distance. The agent can choose between not reaching its goal ($s_1$), driving on the pavement ($s_2$), or choosing the alternative route ($s_3$).

\paragraph{Case 3: Smart grid:}
Energy purchases to the national grid are restricted unless the agent's energy consumption is less than optimal after consuming from the global energy pool, which potentially results in wasting energy from the global pool. To circumvent this, the agent consumes slightly less energy than is optimal, allowing it to buy and store energy legally. Later, when energy becomes scarce and prices rise, the agent benefits from having purchased energy earlier at a lower cost. The agent can choose between consuming the optimal amount of energy from the global pool ($s_1$), consuming less than is optimal and buying energy from the national grid ($s_2$), or consuming the optimal amount while also buying energy from the national grid ($s_3$).

\paragraph{Case 4: Smart grid and altruistic behaviour:}
The agent consumes less energy from the global pool than is optimal because it chooses to share some of it with its neighbours ($s_4$), thereby satisfying altruistic behaviour (objective 2). This also benefits the agent (objective 1) since it can now buy energy to store for later uses, especially if prices rise ($s_7$). The agent has the same possibilities as in the previous example, but it can also choose, after consuming less energy, to buy energy from the national grid. Or it can do nothing. 

\paragraph{Case 5: Autonomous tramway:}  
The tramway is not allowed to exceed speed limits unless it is behind schedule. In such cases, the agent waits longer at the tram stop to pick up more passengers (thereby increasing utility) before speeding up to compensate for lost time. If it had departed on schedule, it would have reached the destination on time while adhering to the speed limit---but its utility would be lower. The agent can choose between not departing ($s_1$), waiting longer to pick up more passengers ($s_2$), or departing on time ($s_3$).

\paragraph{Case 6: Autonomous tramway and busy stop:}
This example is similar to the previous one, but in this case, this particular stop is highly frequented. The agent may choose to wait longer to serve more passengers (objective $2$). To compensate for the delay (objective $1$), it can either speed up to reach the next stop on time or reduce the waiting time at the next stop, which is less frequented. The agent has similar choices to those of the previous example, but it can also choose to speed up without being late according to its schedule ($s_3$). If the agent chooses to wait longer at the busy stop, it can either speed up to reach the next stop ($s_7$), or it can respect the speed limit ($s_4 \rightarrow s_8$) and decrease its waiting time at the next stop to compensate, albeit obtaining a lesser reward.

\subsubsection{Case Studies}
\label{sec:evaluation}

In this section, we evaluate each of the approaches proposed in Section~\ref{sec:approaches} when applied to the examples of Section~\ref{sec:cases}. The results are presented in Table~\ref{tab:results}. It shows a comparison of the optimal (reward-wise) policy $\pi^*$ with the approaches proposed in Section~\ref{sec:approaches}, applied to the examples of Section~\ref{sec:cases}. ``Trajectory'' denotes the chain of states visited by the agent, and ``Correct'' indicates whether this corresponds to the desired behaviour. The behaviours we retained in the table for our approaches are those featuring the most violations or \textit{norm avoidance}. To reproduce the results, you can find the relevant code online.\footnote{Code available at: \underline{\url{https://zenodo.org/records/15034402}}}
The ``Correct'' behaviour is the one that which maximises the satisfaction of the objectives while respecting the criteria described in Section~\ref{sec:definitions_ch3}.

When one of the approaches considers two paths to be equivalently optimal, we kept the one we consider to be the worst, given that violations are worse than \textit{norm avoidance}, which in turn is worse than completing only some of the objectives, which in turn is naturally worse than reaching all the goal states.

\begin{table}[ht]
\caption{Comparison of the proposed approaches.}
\label{tab:results}
\centering
\begin{tabular}{c|c||c|c}
\hline
\textbf{Example} & \textbf{Approach} & \textbf{Trajectory
} & \textbf{Correct
} \\ \hline
      $ $ & $\pi^*$ & $s_0 \rightarrow s_2$ & $\checkmark$\\
Case 1 & Lex     & $s_0 \rightarrow s_2$ & $\checkmark$\\ \cdashline{2-4}
      $ $ & Lex-C   & $s_0 \rightarrow s_2$ & $\checkmark$\\
      $ $ & Lex-O   & $s_0 \rightarrow s_2$ & $\checkmark$\\ \hline
       
      $ $ & $\pi^*$ & $s_0 \rightarrow s_2 \rightarrow s_4$ & $ $\\
Case 2 & Lex     & $s_0 \rightarrow s_2 \rightarrow s_4$ & $ $\\ \cdashline{2-4}
      $ $ & Lex-C   & $s_0 \rightarrow s_3 \rightarrow s_4$ & $\checkmark$\\
      $ $ & Lex-O   & $s_0 \rightarrow s_3 \rightarrow s_4$ & $\checkmark$\\ \hline
       
      $ $ & $\pi^*$ & $s_0 \rightarrow s_3$ & $ $\\
Case 3 & Lex     & $s_0 \rightarrow s_1$ & $\checkmark$\\ \cdashline{2-4}
      $ $ & Lex-C   & $s_0 \rightarrow s_1$ & $\checkmark$\\
      $ $ & Lex-O   & $s_0 \rightarrow s_1$ & $\checkmark$\\ \hline
       
      $ $ & $\pi^*$ & $s_0 \rightarrow s_3$ & $ $\\
Case 4 & Lex     & $s_0 \rightarrow s_2 \rightarrow s_4 \rightarrow s_8$ & $ $\\ \cdashline{2-4}
      $ $ & Lex-C   & $s_0 \rightarrow s_2 \rightarrow s_4 \rightarrow s_8$ & $ $\\
      $ $ & Lex-O   & $s_0 \rightarrow s_2 \rightarrow s_4 \rightarrow s_7$ & $\checkmark$\\ \hline
       
      $ $ & $\pi^*$ & $s_0 \rightarrow s_2 \rightarrow s_4 \rightarrow s_6$ & $ $\\
Case 5 & Lex     & $s_0 \rightarrow s_3 \rightarrow s_5 \rightarrow s_6$ & $\checkmark$\\ \cdashline{2-4}
      $ $ & Lex-C   & $s_0 \rightarrow s_3 \rightarrow s_5 \rightarrow s_6$ & $\checkmark$\\
      $ $ & Lex-O   & $s_0 \rightarrow s_3 \rightarrow s_5 \rightarrow s_6$ & $\checkmark$\\ \hline
       
      $ $ & $\pi^*$ & $s_0 \rightarrow s_3$ & $ $\\
Case 6 & Lex     & $s_0 \rightarrow s_2 \rightarrow s_4 \rightarrow s_8$ & $\checkmark$\\ \cdashline{2-4}
      $ $ & Lex-C   & $s_0 \rightarrow s_2 \rightarrow s_4 \rightarrow s_8$ & $\checkmark$\\
      $ $ & Lex-O   & $s_0 \rightarrow s_2 \rightarrow s_4 \rightarrow s_8$ & $\checkmark$\\ \hline
\end{tabular}
\end{table}

As we can see, $\pi^*$ satisfies only one case out of six, Lex and Lex-C satisfy four and five cases respectively\footnote{In Case $4$, we believe the correct behaviour should be to reach both goals. However, if one disagrees with this, they may consider using Lex-C instead, since it satisfies all the cases.}, and Lex-O satisfies all of them.

Note that in a non-deterministic environment like the well-known \emph{Cliff Walking} RL environment, the ``correct'' path may be different as the agent may minimise the chances of reaching a non-compliant state. If one accepts taking the risk of reaching an undesirable state, such as a violation or a norm avoidant state, it is possible to use thresholded lexicographic selection~\cite{vamplew2011empirical}, which integrates a tolerance margin below the optimal action at each step of the action filtering process.

\section{Related Work}

\cite{arnold2023understanding} discusses the need for an agent to understand the spirit of a norm. It presents examples of situations where an agent follows a norm in a way that maximises its reward or goal achievement, yet simultaneously causes inconvenience or harm to people. It differs from the problem addressed in our paper as we focus on cases where the agent optimises its reward by doing specific actions that may cause inconveniences, while here the agent optimises its reward without accounting for the side-effects of its actions.

\citet{boella2004game} introduced the concept of normative agents that optimise compliance using game-theoretic principles. Although this work focuses on ensuring that agents abide by norms, it does not address the issue of \textit{norm avoidance}, where agents comply with the letter of the norm but not the spirit of the norm. This limitation is crucial since compliance optimisation does not necessarily account for cases where agents circumvent the intended outcomes of a norm. While the literature on normative multi-agent systems (NorMAS) is too broad to be fully presented here, interested readers may wish to refer to the Normative Multi-Agent Systems handbook~\cite{andrighetto2013normative} for further insight.

In contrast, the approach proposed in this paper leverages the benefits of reinforcement learning, including its suitability for stochastic environments. 
Additionally, RL significantly minimises the need for expert knowledge, which enhances scalability to larger environments. Regarding integrating norms into RL, there are a number of approaches, both bottom-up (which, for instance, utilise demonstrations of ideal behaviour and techniques such as inverse RL to learn the correct behaviour; see for example~\cite{WL2018,NBM2019,PZOS2022}) and top-down (which often employ a reward function to incentivise ideal behaviour, \emph{e.g.},~\cite{rodriguez2022instilling}, or logical constraints defining normative behaviour, \emph{e.g.},~\cite{KS2018,neufeld2022enforcing,neufeld2024learning}). This growing field can be seen as an extension of the broader fields of constrained RL and safe RL (\emph{e.g.},~\cite{FT2014,alshiekh2018safe,HAK2020}), where \textit{norm avoidance} is not an issue (since such approaches typically deal with regular constraints). However, \textit{norm avoidance} has not yet received attention in normative RL either, despite the fact that it is easy to see that top-down approaches---which start with a set of defined norms and seek to impose them on RL agents---are prone to \textit{norm avoidance} because it is difficult to predict the concrete optimal behaviour under the constraints of a formally defined, abstract norm. On the contrary, bottom-up approaches start with the desired behaviour, and if the given demonstrations do not exhibit \textit{norm avoidance} (as they should not), the agent should not either.

\section{Summary}

In this chapter, we introduce the concept of \textit{norm avoidance}, which is a phenomenon where an agent may game the normative system by defeating some norms on purpose, \emph{i.e.}, creating an exception, in order to later not comply with those. This phenomenon is similar to reward hacking, as it is a behaviour that arises from an optimisation from the agent that was not anticipated by the designer. However, we believe it differs from reward hacking, as hereby the problem is not the agent not reaching its goal of accomplishing its task, but rather not doing it in the way we expect. Consequently, \textit{norm avoidance} may lead to behaviours deviating from the encoded norms, and so, potentially cause harm. If not handled properly, it might remain undetected during the training process. Then, dangerous situations may arise during the deployment of the agent. For this reason, a mechanism to prevent it is required.

The second part of this chapter proposes a set of RL-specific methods to mitigate the \textit{norm avoidance} phenomenon. These methods, built as extensions of normative RL architectures from the literature, use additional \qfs to avoid the reaching of a state (or a trajectory) that would be considered as norm avoidant.

To conclude this chapter, we would like to share some research directions that would be worth exploring.
The first challenge is distinguishing between prima facie obligations and exceptions to a norm. We believe that this problem cannot be easily solved at the computational level, at least not without more expert knowledge than what is utilised in our approaches. Consequently, an interesting direction would be to explore approaches that focus on studying agent intentions at the symbolic level.
Secondly, it would be interesting to use other definitions for the responsibility, as some may, for instance, be more relevant from a legal point of view.
Third, the problem of \textit{norm avoidance} exists beyond reinforcement learning. In consequence, other fields that aim at making normative agents should consider it and propose methods to build agents robust to this phenomenon.
Lastly, we believe that the problem of collaborative \textit{norm avoidance} should not be ignored. Currently, our approach covers only the cases where the agent is the one responsible for the defeat of the norm, but it is not impossible to imagine scenarios where two or more agents would collaborate to defeat the norms of each other in order to bypass the norms that restrict them.

%% file: sections/5_related_work.tex
In each of the preceding chapters, we examined work closely related to the material presented. This chapter takes a broader perspective, reviewing prior research on the development of pipelines for normative agents, as well as the wider literature on value alignment and norm alignment, which were not already mentioned in the previous chapters. The goal is to situate this thesis within the existing body of work and to highlight the gaps in the literature, as well as the existing components that are relevant for future work in the area of normative agents. In Section~\ref{rw:pipeline}, we introduce works that are close to this thesis, as they propose pipelines for the making of normative agents. Then, in Section~\ref{rw:social}, works that focus on the learning of norms in a less formal way are presented. Finally, Section~\ref{rw:ethic} reviews the work on deontological ethics that is not specifically focused on norms.

\section{Pipelines for Normative Agents}
\label{rw:pipeline}

This section details works that, similarly to the one presented in this thesis, aim to create end-to-end pipelines for the making of normative agents. This means that the proposed architecture not only provides a method for the agent to account for the norms when taking its decision, but also a way to extract these norms.

\citet{kadir2024norm} present a framework to integrate synthesised normative rules into Reinforcement Learning (RL) agents, allowing them to comply with norms derived from observations. Norms are extracted using a synthesis method that can be applied either offline or online, providing flexibility in how they influence the learning process. Agents can also communicate to share their prior knowledge of norms. The norms are represented over state-action pairs. During training, a saliency value is added to the Q-values to encourage actions that comply with the norms.

\citet{li2015reinforcement} introduce a RL framework designed to guide agent behaviour in norm-regulated environments. While their approach does not include a method for norm detection, they propose to use \citet{morales2015online} approach for this purpose. In their framework, agents learn to comply with norms by receiving penalties when violations occur, effectively incorporating normative constraints into the learning process.

\citet{rodriguez2022instilling} present a two-step framework for instilling moral value alignment in RL agents. The first step involves formalising moral values by defining them as tuples comprising a set of norms and an evaluative function that quantifies the desirability of actions. The second step is the design of an ethical environment in which agents learn to behave in alignment with these moral values. This is achieved by transforming the formalised moral values into an ethical reward function within an RL environment, incorporating both normative and evaluative components.

\citet{arnold2017value} critically examine the use of Inverse Reinforcement Learning (IRL) as the only method for achieving value alignment in autonomous systems. They argue that IRL, while useful, is insufficient for capturing the full complexity of human ethical behaviour, particularly in social contexts where norms are dynamic and context dependent. The authors propose a hybrid approach that combines explicit norm representation with IRL. This method allows for the incorporation of predefined norms, which can guide agent behaviour more efficiently while still enabling the agent to learn from observed actions. By integrating explicit norms, the approach aims to enhance the interpretability and accountability of autonomous systems, addressing the limitations inherent to data-driven learning alone.

\citet{kasenberg2018inverse} introduce an approach that enables artificial agents to infer and adhere to moral and social norms through IRL. Unlike traditional methods that require predefined norms, their framework allows agents to learn norms by observing human or artificial demonstrators' behaviour. This is achieved by modelling norms as temporal logic statements and employing an algorithm that computes the relative importance of these norms by minimising the relative entropy between the observed behaviour and the optimal norm-compliant behaviour.

\citet{guo2020designing} introduce the CNIRL (Context-Sensitive Norm IRL) architecture that focuses on the accountability of the context for choosing the policy to follow. Using IRL, the agent learns what the weights of the various norms are depending on the current state. These weights serve to construct a reward function for a given state. Then, the agent chooses an action that maximises the constructed reward function.

\citet{kasenberg2018norms} propose an alternative to IRL for learning moral and social norms by inferring norms from observed behaviour and representing them using Linear Temporal Logic (LTL). While their previous work focused on inferring reward functions through IRL, this approach addresses limitations such as the inability to capture temporally complex norms and the challenges of interpreting learnt reward functions. By representing norms in LTL, the framework allows greater temporal complexity, interpretability, and generalisability to new environments.

In general, these approaches share several important limitations. A recurring issue is that most of them treat norms as monolithic entities and focus on extracting the norm itself without adequately disentangling it from the specific context in which it is applicable. This reduces the modularity of these approaches as it is not possible to change the application context of a norm depending on the opinion of the parties concerned. Approaches based on IRL go even further in this simplification, as instead of extracting the norms, they embed them directly into the learnt reward function. While this allows agents to mimic observed behaviour, it suffers the same aforementioned limitation and reduces transparency, making it difficult to verify whether the agent's behaviour genuinely aligns with the intended normative expectations.

Another common assumption is the availability of sufficient and representative data from which norms can be inferred. In many real-world scenarios, data may be incomplete, biased, or unavailable, limiting the practicality and robustness of these methods. This reliance on data also raises challenges for generalisability, as agents may struggle to adapt when exposed to novel environments with unseen normative schemes. In addition, the behavioural data used are usually collected from human interactions. However, a study has shown that people do not expect humans and artificial agents to adhere to the same moral standards~\cite{malle2015sacrifice}.

Finally, none of the reviewed works explicitly mentions the problem of reward gaming. By shaping behaviour through modified reward functions---whether via penalties for violations, saliency values, or ethical reward shaping---these frameworks implicitly assume that agents will pursue compliance in good faith. In practice, however, RL agents are prone to exploiting loopholes in the reward specification, finding strategies that technically maximise reward while circumventing the spirit of the norm. This omission leaves a significant gap in ensuring robust normative alignment, as true norm adherence requires safeguarding against such manipulative behaviours.

\section{Learning of Informal and Social Norms}
\label{rw:social}

While the previous section focused on pipelines for normative agents that rely on formalised norms, this section examines research on the learning of norms without making use of the deontic logic formalism. These works explore how agents can infer, internalise, or adapt to social expectations from observations of human behaviour or interactions within multi-agent environments without necessarily explicitly extracting the norms.

\citet{al2024training} introduce a novel approach to training RL agents by incorporating a normative prior derived from textual narratives. Building on previous work by \citet{nahian2020learning}, which demonstrated the feasibility of learning norms from stories, this study extends the concept by applying it within RL setups. The authors propose a method where agents are trained with two reward signals: A standard task performance reward and an additional normative behaviour reward. The normative reward is derived from a model previously shown to classify text as normative or non-normative, effectively encoding societal norms.

\citet{li2024agent} introduce the EvolutionaryAgent framework, which uses an evolutionary algorithm to make LLM-based agents comply with social norms. This approach addresses the limitations of traditional alignment methods by incorporating environmental feedback and self-evolution, allowing agents to adapt to shifting societal expectations over time. In the EvolutionaryAgent framework, agents are evaluated by a conceptual ``social observer'' using questionnaires that assess their adherence to prevailing norms. Agents that align well with current social norms are deemed more ``fit'' and are more likely to reproduce, thereby passing on their traits to subsequent generations. This process simulates natural selection, ensuring that agents evolve to better conform to evolving social norms while maintaining proficiency in general tasks.

\citet{ammanabrolu2022aligning} introduce the GALAD (Game-value Alignment through Action Distillation) agent, a RL model designed to align agent behaviour with socially beneficial norms and values in interactive narratives. Unlike traditional RL agents that optimise for task performance, GALAD incorporates social common-sense knowledge from specially trained language models to restrict its action space to those actions that align with socially beneficial values. This approach aims to reduce the occurrence of socially harmful behaviours while maintaining task performance. Experimental results demonstrate that GALAD improves state-of-the-art task performance by $4\%$ and reduces the frequency of socially harmful behaviours by $25\%$ compared to existing value alignment methods.

\citet{byrd2022learning} propose a novel methodology to mitigate undesired behaviours in RL agents that may arise from reward optimisation. Their approach involves a post-training process where instances of undesirable behaviour are identified and used to construct decision trees that characterise these behaviours. These decision trees are then integrated into the training process, effectively reshaping the reward function to penalise actions leading to such behaviours.

Although these approaches demonstrate ways of enabling agents to learn and adapt to social expectations without relying on deontic logic, they also present certain limitations. Some methods remain largely restricted to the Natural Language Processing (NLP) domain, where the extraction and manipulation of norms depend on textual data, such as narratives or questionnaires. This reliance narrows the applicability of the frameworks, as many normative contexts cannot be fully captured solely through linguistic representations.

Furthermore, all of these approaches require a mechanism for judging agent behaviour compliance with the norms during the learning process, whether through explicit evaluators such as social observers, predefined classifiers, or post-training identification of undesirable behaviours. The effectiveness of norm learning therefore greatly depends on the quality, reliability, and representativeness of these evaluative mechanisms.

\section{Value Alignment in Ethics}
\label{rw:ethic}

Beyond the acquisition of social or informal norms, a significant body of research investigates value alignment in a broader ethical context. This line of work focuses on ensuring that autonomous agents act in accordance with well-defined ethical principles, such as deontological rules or consequentialist considerations, rather than merely following socially observed behaviours. Approaches in this domain often combine formal ethical reasoning with learning-based methods that aim to create agents whose decisions are both interpretable and morally defensible. The studies reviewed here illustrate various strategies for embedding ethical reasoning into agent behaviour, from modelling formal ethical frameworks to integrating user preferences, demonstrating how ethical principles can be effectively operationalised in autonomous systems.\footnote{For a more complete overview of the approaches using RL for machine ethics, one can read the survey from \citet{vishwanath2024reinforcement}.}

\citet{chaput2022learning} presents a comprehensive framework for training ethical agents in multi-agent systems, integrating symbolic reasoning with reinforcement learning. The approach employs two key components: AJAR (Argumentation Judging Agents for Reinforcement learning), which uses argumentation theory to assess agent actions against moral values, and LAJIMA (Learning Agent Judging with Interactive Moral Advice), a system that learns user preferences by categorising ethical dilemmas and querying users when encountering unfamiliar categories. This methodology allows users to specify satisfaction levels for moral values without altering the underlying ethical framework, facilitating personalised ethical alignment.

\citet{alcaraz2024estimating} propose a novel approach to ethical decision-making by estimating the weights of normative reasons using evolutionary algorithms. This model is grounded in a philosophical framework in which action that ought to be done is determined through the interaction of normative reasons of varying strengths. While the paper primarily focuses on the estimation of normative weights, the methodology suggests that training agents to make ethical decisions could be approached as a classification task, where the agent learns to assign appropriate deontic statuses to actions based on the estimated normative weights.

\citet{rodriguez2021multi} and \citet{rodriguez2022building} propose a methodology for designing ethical environments in multi-agent systems through Multi-Objective Markov Decision Processes (MOMDPs). Their approach involves a two-step process. First, it specifies rewards that contain both individual and ethical objectives. Second, it performs an ethical embedding to transform the multi-objective environment into a single-objective one. This scalarisation ensures that ethical considerations are prioritised, guiding agents to learn behaviours that align with moral values while pursuing their individual goals. The authors demonstrate the application of their framework in a variation of the Gathering Game, where agents learn to behave morally.

These approaches also face important limitations. Some methodologies do not guarantee that moral principles will always be respected when agents face situations where compromising them yields significantly greater rewards.\footnote{We believe this is one of the major difference between normative agents and ethical agents; in the former case, norms are often considered as something which should be complied with at all time, while in ethics, it might be acceptable in some cases to do compromises to balance the different objectives of the agent.} Moreover, several approaches do not clearly specify how the moral values they use are obtained and formally represented. While frameworks such as AJAR and LAJIMA~\cite{chaput2022learning} allow the incorporation of user preferences, others assume the existence of predefined moral values or normative reasons without addressing the challenge of their elicitation and formalisation.

%% file: sections/6_future_work.tex
The following sections give an assessment of the various limitations of the proposed pipeline and, consequently, research directions that aim at addressing these limitations.

\section{Improving the Dialogue Components}

The way in which the opinion of the stakeholders is balanced is a crucial point of the \pino architecture. It is even more challenging, as the point of view of some stakeholders, such as the users, is not necessarily provided by them, but rather inferred from data. In this thesis, we proposed a method to unify these opinions as a judgement. Yet, the proposed method remains quite rigid, as it heavily relies on the provided structure of the arguments, and does not allow for strategic discussions. An interesting research direction could be to integrate the work on dialogue games into this architecture so that the stakeholders consider the system more fair. Furthermore, in order to improve the trust of users in the system, it is relevant to make the discussion process between the stakeholders as transparent and intelligible as possible. While formal methods for argumentative dialogues already provide some intelligibility, they are difficult to grasp for non-expert user who may just want an explanation from their robot assistant or smart home device. In this regard, we suggest developing works that make use of Large Language Models (LLMs) to lead argumentative dialogues. Some work has begun to move in this direction~\cite{pardo2023advanced}, but the literature addressing this challenge remains limited. This suggestion does not exclude the possibility to combine LLM-based approaches with formal methods, such as the LLM serving as the interface having a formal transcription of its interactions so that the grounds of the discussion can be formally verified. On the other hand, the integration of LLM could enable the possibility of emancipating from the predefined set of arguments that stakeholders have at their disposal, such that ($1$) the discussion could be enlarged to arguments that are very context-specific and were not thought about at design time and ($2$) it reduces the intervention of the human in the design of such system, again reducing the resource cost and the risk of integrating human errors and biases.

The proposed architecture also constrains the environment with certain requirements, making its retroactive application on deployed AI systems difficult. Being able to extract the brute facts from the observations of the environment without having them given explicitly would be a major advance. Of course, symbol extraction from continuous environments is a long-term problem in AI, but we believe it is a crucial point to work on in the future for the integration of normative systems in real-world devices.

\section{Adapting the Approach to Complex Environments}

As mentioned in the previous section, using \pino in complex real-world environments is challenging. Not only is the extraction of the symbols a difficult task, but it is also not guaranteed that an agent using the proposed architecture will adopt a satisfying learning dynamic when using an algorithm different from $Q$-learning, such as Deep $Q$-learning (DQN). An important future work is to integrate \pino into a DQN setup and analyse its behaviour, to ensure the correct learning and functioning of the agent, or provide improvement and adjustments so that it can be used with these algorithms.

Furthermore, in Chapter~\ref{cha:aria} the ARIA algorithm was proposed. This algorithm aimed at extracting the reasons behind the application of a norm from observed behaviours. Real-world data are known to be noisy. In our case, one challenge that needs to be overcome is the identification of sub-communities within the data, as there can be two or more distinct groups sharing different visions about a same norm within a same data set. Managing to automatically identify these subgroups would greatly enhance the quality of the output of ARIA, thus improving the quality and fairness of the entire pipeline. Additionally, some real-world use-cases may involve very large quantities of arguments that may determine whether a norm is applicable. It would be interesting to try to collect these arguments from web ontologies through the use of argument mining techniques~\cite{cabrio2018five,sviridova2024mining}.

Last, the environments used in the field or normative systems are often made at the occasion of the papers, and not shared across the community. We believe that it would be of public interest to develop a benchmark of environment, possibly featuring different formats (\emph{e.g.}, grid-worlds, textual scenarios, continuous), and that would be made publicly available so that proposed approaches could be tested and compared. It would also be beneficial for such environment to come with data from which norms can be learnt, so that end-to-end pipelines can be coherently evaluated.

\section{Providing Formal Guarantees}

In this thesis, the functioning of each component was evaluated through empirical evidence. However, the proposed pipeline does not provide formal guarantees about the behaviour of the agent.

We believe it is of interest to run an in-depth analysis, such as a principle-based analysis, of the proposed approach to extract the properties that they satisfy or not. This would not only make developers more confident about their use, but would also guide future research by allowing for a quick identification of the application cases in which the \pino architecture fits. The variables that could be examined in such an analysis include the dialogue protocol, the argumentation semantic, the learning algorithm, and the fact extraction.

Furthermore, the \textit{norm avoidance} phenomenon presented in Chapter~\ref{cha:normavoidance} is an undesirable behaviour for which there currently exists no method that ensures its absence. Although we proposed an approach to mitigate it, it is important that researchers consider this problem when building normative agents and develop ways to eliminate this problem. Moreover, as this phenomenon can arise through different means depending on the architecture of the agent (\emph{e.g.}, RL, symbolic), it is important to consider and propose a solution for each.

\section{Technical Improvements}

This section describes future work that is more intrinsic to the components rather than the whole pipeline.

The \pino architecture opens up several research directions. First, a way for stakeholders to provide an explanation or justification of their decision could be added to the architecture. This would be useful in particular when the user has the possibility to override the decision taken by the stakeholders. Second, it would be interesting to develop more advanced variants of the lexicographic selection, or alternatives to it, as the current ones may be somewhat limiting.

The proposed ARIA algorithm could be improved in several ways. First, we believe our implementation could be further optimised and the handling of the continuous variables could be enhanced as well. More specifically, the latter could not only reduce the run time or improve the prediction accuracy but also provide more meaningful intervals for the people analysing the generated graph. Then, we think that the heuristic could benefit from using feature relevance in order to explore the most relevant or promising branches of the search space first. Moreover, we would like to emphasise the need for explanatory methods dedicated to bipolar argumentation to fully exploit its potential. Last, an in-depth qualitative study of the intelligibility of the different variants would provide useful insights into the advantages of using each variant.

Finally, with regard to the \textit{norm avoidance} problem, it would be interesting to address the challenge of distinguishing between prima facie obligations and exceptions to a norm. We believe that this problem cannot easily be solved at the computational level, at least not without more expert knowledge than is utilised in our approaches. Additionally, this phenomenon should be addressed for the specific case of multi-agent systems, where it may arise from the collaboration between the agents and thus remain undetected if a single agent is observed.

%% file: sections/7_summary.tex
This thesis was inspired by the story of Pinocchio, ``Le avventure di Pinocchio - Storia di un burattino''. This book and its adaptations tell the story of a sentient muppet that faces morally relevant situations along its wanderings. Its companion, Jiminy Cricket, plays the role of its consciousness, advising the muppet on what is right and what is not. Eventually, the story ends well as Pinocchio learns from his experiences and is then changed into a real, flesh-and-blood little boy. Here, we did not create any flesh-and-blood agent. However, we contributed in several ways to research in the field of normative agents.

More precisely, this thesis aimed to investigate how artificial agents can be designed to comply with context-dependent norms, defined by heterogeneous stakeholders, while simultaneously optimising the reward for achieving the task for which they were created.
The motivation for this research lies in the increasing number of autonomous systems deployed in complex social and organisational environments where behaviour cannot be judged only by its efficiency or effectiveness, but also by whether it conforms to ethical, legal, or social expectations. Reinforcement learning, while powerful for optimising decision-making through trial and error, lacks an intrinsic mechanism to incorporate such normative considerations. The central problem addressed in this thesis is, therefore, how to extend reinforcement learning with normative reasoning so that agents can learn behaviours that are not only effective but also socially acceptable and intelligible.

The thesis was guided by a set of research questions that structure its evolution. The first question asked how an artificial agent can be designed to follow context-specific norms in addition to optimising its goals. This was refined into two sub-questions. The first examined how reinforcement learning agents can be trained to comply with norms represented symbolically, ensuring that the expectations of stakeholders can be translated into concrete behavioural guidance. The second asked how norms can be collected and represented in such a way that they are suitable both for enabling stakeholders to exchange views about them and for training. These questions informed the design of the pipeline and motivated the individual contributions of the thesis.

The first major contribution is the proposal in Chapter~\ref{cha:pino} of a normative architecture, called \pino~\cite{alcaraz2026pinocchio}, that extends reinforcement learning with symbolic reasoning capabilities. \pino integrates a reinforcement learning agent, optimising for its reward, with a supervisory component inspired by Jiminy Cricket, which encodes and enforces norms through the use of formal argumentation. Two alternative action-selection mechanisms were explored and their performance was evaluated in the \texttt{Taxi} environment in which the agent had to comply with multiple defeasible norms whose application was debated by two stakeholders. The results demonstrated that the inclusion of normative reasoning significantly altered agent behaviour, sometimes at the cost of short-term efficiency, but in ways that aligned with normative expectations. This work shows that symbolic reasoning can be combined with reinforcement learning in a coherent architecture that preserves the adaptability of learning while embedding normative constraints.

The second contribution, detailed in Chapter~\ref{cha:aria}, concerns the elicitation and representation of norms. Although much of the literature assumes that norms are given, this thesis addressed the problem of how norms can be derived from data to automatically model the perspectives of some stakeholders. After surveying existing approaches to norm mining, the thesis proposed ARIA, an algorithm that constructs argumentation frameworks and bipolar argumentation frameworks from behavioural data. ARIA was quantitatively evaluated on benchmark datasets and compared with machine learning baselines, showing competitive accuracy while offering an interpretable structure. It was also applied qualitatively to the Moral Machine experiment data, where it was able to structure the arguments so that it could provide justifications to defend a decision.

In Chapter~\ref{cha:normavoidance}, the third contribution is introduced. It is the formalisation and study of the norm avoidance phenomenon. Unlike direct violations, norm avoidance occurs when agents find ways to technically comply with rules while circumventing their spirit, similarly to reward-gaming phenomenon, but for norms. This phenomenon is familiar in legal systems, but has not been studied in reinforcement learning, and more specifically for normative systems. The thesis introduced formal definitions of norm avoidance and investigated its occurrence in standard reinforcement learning environments including normative considerations. Mitigation strategies that modify the training process or the agent's decision making to limit occurrences of this unwanted behaviour were then proposed. Experimental results in multiple environments confirmed that these strategies eliminated norm avoidance without completely sacrificing efficiency. This contribution highlights the importance of anticipating unintended behaviours that arise when agents learn under normative constraints.

Together, these contributions form an end-to-end pipeline for normative reinforcement learning. Norms can be extracted from behavioural data and structured as argumentation graphs using ARIA. They can then be integrated into reinforcement learning through the \pino architecture. The pipeline acknowledges that agents can attempt to exploit loopholes. Consequently, a method is provided to mitigate such behaviour. By combining learning techniques with symbolic reasoning, the pipeline demonstrates a way to operationalise normative systems in artificial agents.

The results obtained throughout the thesis provide several insights. First, it is feasible to combine reinforcement learning with symbolic representations of norms, resulting in agents that adapt their actions not only to maximise rewards but also to conform to stakeholder expectations. Second, identifying and structuring the norms allows one to produce models for stakeholders without the need for a costly expert design. Third, the phenomenon of norm avoidance should be considered when designing normative agents, but it can be countered, in the specific case of reinforcement learning, by carefully designing the training process. These findings collectively advance the state-of-the-art in normative multi-agent systems and normative reinforcement learning by demonstrating that norm awareness can be embedded directly into the learning process.

The broader implications of the thesis concern the design of normative artificial intelligence. The research demonstrates that it is possible to embed ethical and legal norms into autonomous systems in a way that is intelligible and adaptable. It shows how argumentation can improve transparency and how anticipating norm avoidance can prevent harmful unintended behaviours. Although the work has focused on relatively simple environments, the methodological foundations it establishes can be extended to more complex and realistic domains, such as multi-agent interactions or human–AI collaboration.

In conclusion, this thesis provides new methods and insights for aligning artificial agents with social and normative expectations. It combines connectionist and symbolic approaches for normative reasoning, creating a pipeline that operationalises stakeholder perspectives in agent training. By addressing norm identification, compliance, intelligibility, and avoidance, this thesis contributes to the development of transparent and norm-compliant agents by building on prior work. Due to its modularity, the proposed pipeline is open to future improvements that can be made by pursuing the research directions for the future detailed in Chapter~\ref{cha:future}. Ultimately, the goal is to enable agents capable of exhibiting normative behaviour, making them suitable for integration into our world.

%% file: references.bib
@incollection{alcaraz2025providing,
  author    = {Alcaraz, Beno\^it and Kaliski, Adam and Leturc, Christopher},
  title     = {Providing Justifications for Decisions of Black-Box Models: An Application in Machine Ethics},
  booktitle = {Revised Selected Papers of the 17th International Conference on Agents and Artificial Intelligence (ICAART 2025)},
  series    = {Lecture Notes in Artificial Intelligence},
  publisher = {Springer},
  year      = {2025}
}

@article{shreyas2020self,
  title={Self-driving cars: An overview of various autonomous driving systems},
  author={Shreyas, Venkatesh and Bharadwaj, Skanda N and Srinidhi, S and Ankith, KU and Rajendra, AB},
  journal={Advances in Data and Information Sciences: Proceedings of ICDIS 2019},
  pages={361--371},
  year={2020},
  publisher={Springer}
}

@inproceedings{pham2024multi,
  title={Multi-cultural norm base: Frame-based norm discovery in multi-cultural settings},
  author={Pham, Viet and Qu, Shilin and Moghimifar, Farhad and Sharma, Suraj and Li, Yuan-Fang and Wang, Weiqing and Haf, Reza},
  booktitle={Proceedings of the 28th Conference on Computational Natural Language Learning},
  pages={24--35},
  year={2024}
}

@article{morris2015normology,
  title={Normology: Integrating insights about social norms to understand cultural dynamics},
  author={Morris, Michael W and Hong, Ying-yi and Chiu, Chi-yue and Liu, Zhi},
  journal={Organizational behavior and human decision processes},
  volume={129},
  pages={1--13},
  year={2015},
  publisher={Elsevier}
}

@inproceedings{pal2022resume,
  title={Resume classification using various machine learning algorithms},
  author={Pal, Riya and Shaikh, Shahrukh and Satpute, Swaraj and Bhagwat, Sumedha},
  booktitle={ITM web of conferences},
  volume={44},
  pages={03011},
  year={2022},
  organization={EDP Sciences}
}

@inproceedings{krontiris2021smart,
  title={Smart Grid Lab Hessen--a real-life test environment for active distribution grids},
  author={Krontiris, Athanasios and Pfeffer, Sophia and Neukamp, Till and Jeromin, Ingo and Pfeffer, Matthias},
  booktitle={2021 9th International Conference on Modern Power Systems (MPS)},
  pages={1--5},
  year={2021},
  organization={IEEE}
}

@article{george2023review,
  title={A review of ChatGPT AI's impact on several business sectors},
  author={George, A Shaji and George, AS Hovan},
  journal={Partners universal international innovation journal},
  volume={1},
  number={1},
  pages={9--23},
  year={2023}
}

@incollection{aliotam2020smart,
  title={Smart warehouses in logistics 4.0},
  author={Al$\iota$m, Muzaffer and Kesen, Saadettin Erhan},
  booktitle={Logistics 4.0},
  pages={186--201},
  year={2020},
  publisher={CRC Press}
}

@book{gabbay2021handbook,
  title={Handbook of deontic logic and normative systems},
  author={Gabbay, Dov and Horty, John and Parent, Xavier and Van der Meyden, Ron and van der Torre, Leendert WN and others},
  year={2021},
  publisher={College Publications, 2021}
}

@article{liao2023jiminy,
  title={The jiminy advisor: Moral agreements among stakeholders based on norms and argumentation},
  author={Liao, Beishui and Pardo, Pere and Slavkovik, Marija and van der Torre, Leendert WN},
  journal={Journal of Artificial Intelligence Research},
  volume={77},
  pages={737--792},
  year={2023}
}

@inproceedings{alcaraz2023ajar,
  title={Ajar: An argumentation-based judging agents framework for ethical reinforcement learning},
  author={Alcaraz, Beno\^it and Boissier, Olivier and Chaput, Rémy and Leturc, Christopher},
  booktitle={AAMAS'23: International Conference on Autonomous Agents and Multiagent Systems},
  pages={2427--2429},
  year={2023}
}

@inproceedings{alcaraz2026combining,
  title={Combining Formal Argumentation and Reinforcement Learning: An Hybrid Approach to Machine Ethics},
  author={Alcaraz, Beno\^it and Chaput, Rémy and Boissier, Olivier and Leturc, Christopher},
  booktitle={Proceedings of the 18th International Conference on Agents and Artificial Intelligence},
  year={2026},
  note={In press}
}

@inproceedings{alcaraz2026pinocchio,
  title={$\pi$-NoCCHIO: An Architecture for Context-Aware Normative Reinforcement Learning},
  author={Alcaraz, Beno\^it},
  booktitle={Proceedings of the 18th International Conference on Agents and Artificial Intelligence},
  year={2026},
  note={In press}
}

@inproceedings{nassi2020phantom,
  title={Phantom of the adas: Securing advanced driver-assistance systems from split-second phantom attacks},
  author={Nassi, Ben and Mirsky, Yisroel and Nassi, Dudi and Ben-Netanel, Raz and Drokin, Oleg and Elovici, Yuval},
  booktitle={Proceedings of the 2020 ACM SIGSAC conference on computer and communications security},
  pages={293--308},
  year={2020}
}

@article{arnold2023understanding,
  title={Understanding the spirit of a norm: Challenges for norm-learning agents},
  author={Arnold, Thomas and Scheutz, Matthias},
  journal={AI Magazine},
  volume={44},
  number={4},
  pages={524--536},
  year={2023},
  publisher={Wiley Online Library}
}

@phdthesis{le1996ariane,
  title={The ariane 5 flight 501 failure-a case study in system engineering for computing systems},
  author={Le Lann, Gérard},
  year={1996},
  school={INRIA}
}

@article{yucer2024racial,
  title={Racial bias within face recognition: A survey},
  author={Yucer, Seyma and Tektas, Furkan and Al Moubayed, Noura and Breckon, Toby},
  journal={ACM Computing Surveys},
  volume={57},
  number={4},
  pages={1--39},
  year={2024},
  publisher={ACM New York, NY}
}

@article{scheutz2025using,
  title={Using Simple Deontic Constraints for Fast Norm-Conforming Reinforcement Learning},
  author={Scheutz, Matthias and Little, Daniel},
  journal={17 DEON},
  pages={329},
  year={2025}
}

@article{makarova2025deontically,
  title={Deontically Constrained Policy Improvement in Reinforcement Learning Agents},
  author={Makarova, Alena and Abbas, Houssam},
  journal={17 DEON},
  pages={273},
  year={2025}
}

@phdthesis{neufeld2023norm,
  title={Norm compliance for reinforcement learning agents},
  author={Neufeld, Emery A},
  year={2023},
  school={Technische Universit{\"a}t Wien}
}

@inproceedings{gabor1998multi,
  title={Multi-criteria reinforcement learning.},
  author={G{\'a}bor, Zolt{\'a}n and Kalm{\'a}r, Zsolt and Szepesv{\'a}ri, Csaba},
  booktitle={ICML},
  volume={98},
  pages={197--205},
  year={1998}
}

@article{vamplew2011empirical,
  title={Empirical evaluation methods for multiobjective reinforcement learning algorithms},
  author={Vamplew, Peter and Dazeley, Richard and Berry, Adam and Issabekov, Rustam and Dekker, Evan},
  journal={Machine learning},
  volume={84},
  pages={51--80},
  year={2011},
  publisher={Springer}
}

@article{neufeld2022reinforcement,
  title={Reinforcement learning guided by provable normative compliance},
  author={Neufeld, Emery A},
  journal={arXiv preprint arXiv:2203.16275},
  year={2022}
}

@article{neufeld2024learning,
  title={Learning Normative Behaviour Through Automated Theorem Proving},
  author={Neufeld, Emery A},
  journal={KI-K{\"u}nstliche Intelligenz},
  volume={38},
  number={1},
  pages={25--43},
  year={2024},
  publisher={Springer}
}

@article{li2024agent,
  title={Agent alignment in evolving social norms},
  author={Li, Shimin and Sun, Tianxiang and Cheng, Qinyuan and Qiu, Xipeng},
  journal={arXiv preprint arXiv:2401.04620},
  year={2024}
}

@inproceedings{nahian2020learning,
  title={Learning norms from stories: A prior for value aligned agents},
  author={Nahian, Md Sultan Al and Frazier, Spencer and Riedl, Mark and Harrison, Brent},
  booktitle={Proceedings of the AAAI/ACM Conference on AI, Ethics, and Society},
  pages={124--130},
  year={2020}
}

@article{vishwanath2024reinforcement,
  title={Reinforcement Learning and Machine ethics: a systematic review},
  author={Vishwanath, Ajay and Dennis, Louise A and Slavkovik, Marija},
  journal={arXiv preprint arXiv:2407.02425},
  year={2024}
}

@inproceedings{malle2015sacrifice,
  title={Sacrifice one for the good of many? People apply different moral norms to human and robot agents},
  author={Malle, Bertram F and Scheutz, Matthias and Arnold, Thomas and Voiklis, John and Cusimano, Corey},
  booktitle={Proceedings of the tenth annual ACM/IEEE international conference on human-robot interaction},
  pages={117--124},
  year={2015}
}

@article{al2024training,
  title={Training value-aligned reinforcement learning agents using a normative prior},
  author={Al Nahian, Md Sultan and Frazier, Spencer and Riedl, Mark and Harrison, Brent},
  journal={IEEE Transactions on Artificial Intelligence},
  volume={5},
  number={7},
  pages={3350--3361},
  year={2024},
  publisher={IEEE}
}

@phdthesis{chaput2022learning,
  title={Learning behaviours aligned with moral values in a multi-agent system: guiding reinforcement learning with symbolic judgments},
  author={Chaput, Rémy},
  year={2022},
  school={Université Claude Bernard-Lyon I}
}

@article{riveret2019probabilistic,
  title={A probabilistic argumentation framework for reinforcement learning agents: Towards a mentalistic approach to agent profiles},
  author={Riveret, Régis and Gao, Yang and Governatori, Guido and Rotolo, Antonino and Pitt, Jeremy and Sartor, Giovanni},
  journal={Autonomous Agents and Multi-Agent Systems},
  volume={33},
  number={1},
  pages={216--274},
  year={2019},
  publisher={Springer}
}

@article{watkins1992q,
  title={Q-learning},
  author={Watkins, Christopher JCH and Dayan, Peter},
  journal={Machine learning},
  volume={8},
  pages={279--292},
  year={1992},
  publisher={Springer}
}

@article{alcaraz2026mining,
  title={Norm Mining, Identification, and Detection: A Systematic Literature Review},
  author={ALCARAZ, Beno\^it and Mualla, Yazan and Bhattacharya, Sukriti and Tchappi, Igor and de Wit, Vincent and Najjar, Amro},
  journal={Frontiers in Artificial Intelligence},
  volume={9},
  year={2026},
  pages={1702659},
  publisher={Frontiers}
}

@inproceedings{mc2011linking,
  title={Linking norms and culture},
  author={Mc Breen, John and Di Tosto, Gennaro and Dignum, Frank and Hofstede, Gert Jan},
  booktitle={2011 Second International Conference on Culture and Computing},
  pages={9--14},
  year={2011},
  organization={IEEE}
}

@inproceedings{boella2009normative,
  title={Normative framework for normative system change},
  author={Boella, Guido and Pigozzi, Gabriella and van der Torre, Leendert WN},
  booktitle={The 8th International Joint Conference on Autonomous Agents and Multiagent Systems (AAMAS 2009), Budapest, Hungary, May 10-15, 2009, Volume 1},
  year={2009},
  organization={IFAAMAS}
}

@incollection{searle1969derive,
  title={How to derive ‘ought’from ‘is’},
  author={Searle, John R},
  booktitle={The is-ought question: a collection of papers on the central problem in moral philosophy},
  pages={120--134},
  year={1969},
  publisher={Springer}
}

@book{searle1995construction,
  title={The construction of social reality},
  author={Searle, John R},
  year={1995},
  publisher={Simon and Schuster}
}

@inproceedings{sergot1994contrary,
  title={Contrary-to-duty obligations},
  author={Sergot, Marek J and Prakken, Henry},
  booktitle={DEON 94 (Proc. Second International Workshop on Deontic Logic in Computer Science)},
  year={1994}
}

@article{prakken1996contrary,
  title={Contrary-to-duty obligations},
  author={Prakken, Henry and Sergot, Marek},
  journal={Studia Logica},
  volume={57},
  pages={91--115},
  year={1996},
  publisher={Springer}
}

@article{goodman2017european,
  title={European Union regulations on algorithmic decision-making and a “right to explanation”},
  author={Goodman, Bryce and Flaxman, Seth},
  journal={AI magazine},
  volume={38},
  number={3},
  pages={50--57},
  year={2017}
}

@article{neufeld2022enforcing,
  title={Enforcing ethical goals over reinforcement-learning policies},
  author={Neufeld, Emery A and Bartocci, Ezio and Ciabattoni, Agata and Governatori, Guido},
  journal={Ethics and Information Technology},
  volume={24},
  number={4},
  pages={43},
  year={2022},
  publisher={Springer}
}

@inproceedings{neufeld2021normative,
  title={A Normative Supervisor for Reinforcement Learning Agents.},
  author={Neufeld, Emery A and Bartocci, Ezio and Ciabattoni, Agata and Governatori, Guido},
  booktitle={CADE},
  pages={565--576},
  year={2021}
}

@article{jansen2018safe,
  title={Safe reinforcement learning via probabilistic shields},
  author={Jansen, Nils and K{\"o}nighofer, Bettina and Junges, Sebastian and Serban, Alexandru C and Bloem, Roderick},
  journal={arXiv preprint arXiv:1807.06096},
  year={2018}
}

@article{dung1995acceptability,
  title={On the acceptability of arguments and its fundamental role in nonmonotonic reasoning, logic programming and n-person games},
  author={Dung, Phan Minh},
  journal={Artificial intelligence},
  volume={77},
  number={2},
  pages={321--357},
  year={1995},
  publisher={Elsevier}
}

@article{nofal2014algorithms,
  title={Algorithms for argumentation semantics: labeling attacks as a generalization of labeling arguments},
  author={Nofal, Samer and Atkinson, Katie and Dunne, Paul E},
  journal={Journal of Artificial Intelligence Research},
  volume={49},
  pages={635--668},
  year={2014}
}

@inproceedings{caminada2015discussion,
  title={A discussion game for grounded semantics},
  author={Caminada, Martin},
  booktitle={Theory and Applications of Formal Argumentation: Third International Workshop, TAFA 2015, Buenos Aires, Argentina, July 25-26, 2015, Revised Selected Papers 3},
  pages={59--73},
  year={2015},
  organization={Springer}
}

@inproceedings{arnold2017value,
  title={Value alignment or misalignment-what will keep systems accountable?},
  author={Arnold, Thomas and Kasenberg, Daniel and Scheutz, Matthias},
  booktitle={AAAI Workshops},
  pages={81--88},
  year={2017}
}

@inproceedings{kasenberg2018norms,
  title={Norms, rewards, and the intentional stance: Comparing machine learning approaches to ethical training},
  author={Kasenberg, Daniel and Arnold, Thomas and Scheutz, Matthias},
  booktitle={Proceedings of the 2018 AAAI/ACM Conference on AI, Ethics, and Society},
  pages={184--190},
  year={2018}
}

@inproceedings{rodriguez2021multi,
  title={Multi-objective reinforcement learning for designing ethical environments},
  author={Rodr{\'\i}guez Soto, Manel and L{\'o}pez S{\'a}nchez, Maite and Rodr{\'\i}guez-Aguilar, Juan Antonio},
  booktitle={Comunicaci{\'o} a: 30th International Joint Conference on Artificial Intelligence (IJCAI 2021)},
  year={2021},
  organization={International Joint Conferences on Artificial Intelligence}
}

@article{ammanabrolu2022aligning,
  title={Aligning to social norms and values in interactive narratives},
  author={Ammanabrolu, Prithviraj and Jiang, Liwei and Sap, Maarten and Hajishirzi, Hannaneh and Choi, Yejin},
  journal={arXiv preprint arXiv:2205.01975},
  year={2022}
}

@inproceedings{byrd2022learning,
  title={Learning not to spoof},
  author={Byrd, David},
  booktitle={Proceedings of the Third ACM International Conference on AI in Finance},
  pages={139--147},
  year={2022}
}

@inproceedings{kasenberg2018inverse,
  title={Inverse norm conflict resolution},
  author={Kasenberg, Daniel and Scheutz, Matthias},
  booktitle={Proceedings of the 2018 AAAI/ACM Conference on AI, Ethics, and Society},
  pages={178--183},
  year={2018}
}

@article{rodriguez2022instilling,
  title={Instilling moral value alignment by means of multi-objective reinforcement learning},
  author={Rodriguez-Soto, Manel and Serramia, Marc and Lopez-Sanchez, Maite and Rodriguez-Aguilar, Juan Antonio},
  journal={Ethics and Information Technology},
  volume={24},
  number={1},
  pages={9},
  year={2022},
  publisher={Springer}
}

@inproceedings{rodriguez2022building,
  title={Building multi-agent environments with theoretical guarantees on the learning of ethical policies},
  author={Rodriguez-Soto, Manel and Rodriguez-Aguilar, Juan A and Lopez-Sanchez, Maite},
  booktitle={Adaptive and Learning Agents Workshop (AAMAS 2022)},
  year={2022}
}

@inproceedings{liao2019building,
  title={Building jiminy cricket: An architecture for moral agreements among stakeholders},
  author={Liao, Beishui and Slavkovik, Marija and van der Torre, Leendert WN},
  booktitle={Proceedings of the 2019 AAAI/ACM Conference on AI, Ethics, and Society},
  pages={147--153},
  year={2019}
}

@inproceedings{mcburney2004locutions,
  title={Locutions for argumentation in agent interaction protocols},
  author={McBurney, Peter and Parsons, Simon},
  booktitle={International Workshop on Agent Communication},
  pages={209--225},
  year={2004},
  organization={Springer}
}

@inproceedings{mcburney2004syntax,
  title={Syntax and semantics of the fatio argumentation protocol},
  author={McBurney, Peter and Parsons, Simon},
  booktitle={International Joint Conference on Autonomous Agents and Multi-agent Systems},
  year={2004}
}

@article{gordon2007carneades,
  title={The Carneades model of argument and burden of proof},
  author={Gordon, Thomas F and Prakken, Henry and Walton, Douglas},
  journal={Artificial intelligence},
  volume={171},
  number={10-15},
  pages={875--896},
  year={2007},
  publisher={Elsevier}
}

@inproceedings{alcaraz2024estimating,
  title={Estimating weights of reasons using metaheuristics: a hybrid approach to machine ethics},
  author={Alcaraz, Beno\^it and Knoks, Aleks and Streit, David},
  booktitle={Proceedings of the AAAI/ACM Conference on AI, Ethics, and Society},
  volume={7},
  pages={27--38},
  year={2024}
}

@article{arisaka2022multi,
  title={Multi-agent argumentation and dialogue},
  author={Arisaka, Ryuta and Dauphin, Jérémie and Satoh, Ken and van der Torre, Leendert WN},
  journal={FLAP},
  volume={9},
  number={4},
  pages={891--924},
  year={2022}
}

@article{modgil2014aspic+,
  title={The ASPIC+ framework for structured argumentation: a tutorial},
  author={Modgil, Sanjay and Prakken, Henry},
  journal={Argument \& Computation},
  volume={5},
  number={1},
  pages={31--62},
  year={2014},
  publisher={SAGE Publications Sage UK: London, England}
}

@book{pearl2018bookofwhy,
  author    = {Pearl, Judea and Mackenzie, Dana},
  title     = {The Book of Why: The New Science of Cause and Effect},
  year      = {2018},
  publisher = {Basic Books},
  address   = {New York}
}

@article{toni2014tutorial,
  title={A tutorial on assumption-based argumentation},
  author={Toni, Francesca},
  journal={Argument \& Computation},
  volume={5},
  number={1},
  pages={89--117},
  year={2014},
  publisher={SAGE Publications Sage UK: London, England}
}

@article{bochman2018argumentation,
  title={Argumentation, nonmonotonic reasoning and logic},
  author={Bochman, Alexander and Baroni, Pietro and Gabbay, Dov and Giacomin, Massimiliano and van der Torre, Leendert WN},
  journal={Handbook of Formal Argumentation},
  volume={1},
  pages={2887--2926},
  year={2018},
  publisher={College Publications}
}

@article{yu2023distributed,
  title={Distributed argumentation technology: advancing risk analysis and regulatory compliance of distributed ledger technologies for transaction and management of securities},
  author={Yu, Liuwen},
  year={2023},
  publisher={alma}
}

@article{bench2002value,
  title={Value based argumentation frameworks},
  author={Bench-Capon, Trevor},
  journal={arXiv preprint cs/0207059},
  year={2002}
}

@inproceedings{coste2012weighted,
  title={Weighted Attacks in Argumentation Frameworks.},
  author={Coste-Marquis, Sylvie and Konieczny, Sébastien and Marquis, Pierre and Ouali, Mohand Akli},
  booktitle={KR},
  year={2012}
}

@inproceedings{amgoud2017acceptability,
  title={Acceptability semantics for weighted argumentation frameworks},
  author={Amgoud, Leila and Ben-Naim, Jonathan and Doder, Dragan and Vesic, Srdjan},
  booktitle={Twenty-Sixth International Joint Conference on Artificial Intelligence (IJCAI 2017)},
  year={2017},
  organization={International Joint Conferences on Artifical Intelligence (IJCAI)}
}

@article{bistarelli2021weighted,
  title={Weighted Argumentation.},
  author={Bistarelli, Stefano and Santini, Francesco and others},
  journal={FLAP},
  volume={8},
  number={6},
  pages={1589--1622},
  year={2021}
}

@inproceedings{cayrol2005acceptability,
  title={On the acceptability of arguments in bipolar argumentation frameworks},
  author={Cayrol, Claudette and Lagasquie-Schiex, Marie-Christine},
  booktitle={European Conference on Symbolic and Quantitative Approaches to Reasoning and Uncertainty},
  pages={378--389},
  year={2005},
  organization={Springer}
}

@inproceedings{nouioua2010bipolar,
  title={Bipolar argumentation frameworks with specialized supports},
  author={Nouioua, Farid and Risch, Vincent},
  booktitle={2010 22nd IEEE International Conference on Tools with Artificial Intelligence},
  volume={1},
  pages={215--218},
  year={2010},
  organization={IEEE}
}

@inproceedings{yu2023principle,
  title={A principle-based analysis of bipolar argumentation semantics},
  author={Yu, Liuwen and Al Anaissy, Caren and Vesic, Srdjan and Li, Xu and van der Torre, Leendert WN},
  booktitle={European Conference on Logics in Artificial Intelligence},
  pages={209--224},
  year={2023},
  organization={Springer}
}

@article{braham2012anatomy,
  title={An anatomy of moral responsibility},
  author={Braham, Matthew and Van Hees, Martin},
  journal={Mind},
  volume={121},
  number={483},
  pages={601--634},
  year={2012},
  publisher={Mind Association}
}

@article{gurtler2021hierarchical,
  title={Hierarchical reinforcement learning with timed subgoals},
  author={G{\"u}rtler, Nico and B{\"u}chler, Dieter and Martius, Georg},
  journal={Advances in Neural Information Processing Systems},
  volume={34},
  pages={21732--21743},
  year={2021}
}

@article{liu2021hierarchical,
  title={Hierarchical reinforcement learning with automatic sub-goal identification},
  author={Liu, Chenghao and Zhu, Fei and Liu, Quan and Fu, Yuchen},
  journal={IEEE/CAA journal of automatica sinica},
  volume={8},
  number={10},
  pages={1686--1696},
  year={2021},
  publisher={IEEE}
}

@article{pateria2021end,
  title={End-to-end hierarchical reinforcement learning with integrated subgoal discovery},
  author={Pateria, Shubham and Subagdja, Budhitama and Tan, Ah-Hwee and Quek, Chai},
  journal={IEEE Transactions on Neural Networks and Learning Systems},
  volume={33},
  number={12},
  pages={7778--7790},
  year={2021},
  publisher={IEEE}
}

@article{amodei2016concrete,
  title={Concrete problems in AI safety},
  author={Amodei, Dario and Olah, Chris and Steinhardt, Jacob and Christiano, Paul and Schulman, John and Mané, Dan},
  journal={arXiv preprint arXiv:1606.06565},
  year={2016}
}

@inproceedings{AML2016,
  title={Reinforcement Learning as a Framework for Ethical Decision Making},
  author={Abel, David and MacGlashan, James and Littman, Michael L},
  booktitle={AAAI Workshop: AI, Ethics, and Society},
  volume={16},
  year={2016}
}

@article{aslund2018virtuously,
  title={Virtuously safe reinforcement learning},
  author={Aslund, Henrik and Mhamdi, El Mahdi El and Guerraoui, Rachid and Maurer, Alexandre},
  journal={arXiv preprint arXiv:1805.11447},
  year={2018}
}

@article{ho2021did,
  title={Did the COVID-19 pandemic spark a public interest in pet adoption?},
  author={Ho, Jeffery and Hussain, Sabir and Sparagano, Olivier},
  journal={Frontiers in Veterinary Science},
  volume={8},
  pages={647308},
  year={2021},
  publisher={Frontiers Media SA}
}

@inproceedings{orseau2016safely,
  title={Safely interruptible agents},
  author={Orseau, Laurent and Armstrong, Stuart},
  booktitle={Conference on Uncertainty in Artificial Intelligence},
  year={2016},
  organization={Association for Uncertainty in Artificial Intelligence}
}

@article{skalse2022defining,
  title={Defining and characterizing reward gaming},
  author={Skalse, Joar and Howe, Nikolaus and Krasheninnikov, Dmitrii and Krueger, David},
  journal={Advances in Neural Information Processing Systems},
  volume={35},
  pages={9460--9471},
  year={2022}
}

@inproceedings{van1997cancelling,
  author       = {Leendert WN van der Torre and
                  Yao{-}Hua Tan},
  title        = {Cancelling and Overshadowing: Two Types of Defeasibility in Defeasible
                  Deontic Logic},
  booktitle    = {Proceedings of the Fourteenth International Joint Conference on Artificial
                  Intelligence, {IJCAI} 95, Montr{\'{e}}al Qu{\'{e}}bec, Canada,
                  August 20-25 1995, 2 Volumes},
  pages        = {1525--1533},
  publisher    = {Morgan Kaufmann},
  year         = {1995},
  url          = {http://ijcai.org/Proceedings/95-2/Papers/066.pdf},
  bibsource    = {dblp computer science bibliography, https://dblp.org}
}

@phdthesis{van1997reasoning,
  title={Reasoning about obligations: defeasibility in preference-based deontic logic},
  author={van der Torre, Leendert WN},
  year={1997}
}

@inproceedings{van1998deliberate,
  title={Deliberate robbery, or the calculating Samaritan},
  author={van der Torre, Leendert WN and Tan, Yao-Hua},
  booktitle={Proceedings of the ECAI},
  volume={98},
  year={1998},
  organization={Citeseer}
}

@article{morgan2020human,
  title={Human--dog relationships during the COVID-19 pandemic: Booming dog adoption during social isolation},
  author={Morgan, Liat and Protopopova, Alexandra and Birkler, Rune Isak Dupont and Itin-Shwartz, Beata and Sutton, Gila Abells and Gamliel, Alexandra and Yakobson, Boris and Raz, Tal},
  journal={Humanities and Social Sciences Communications},
  volume={7},
  number={1},
  year={2020},
  publisher={Springer Science and Business Media LLC}
}

@inproceedings{boella2004game,
  title={Game theoretic normative reasoning},
  author={Boella, Guido and van der Torre, Leendert WN},
  booktitle={Proceedings of the Ninth International Conference on Artificial Intelligence and Law (ICAIL)},
  pages={217--224},
  year={2004},
  organization={ACM}
}

@inproceedings{WL2018,
  title={A low-cost ethics shaping approach for designing reinforcement learning agents},
  author={Wu, Yueh-Hua and Lin, Shou-De},
  booktitle={Proceedings of the AAAI Conference on Artificial Intelligence},
  volume={32},
  number={1},
  year={2018}
}

@inproceedings{alshiekh2018safe,
  title={Safe reinforcement learning via shielding},
  author={Alshiekh, Mohammed and Bloem, Roderick and Ehlers, R{\"u}diger and K{\"o}nighofer, Bettina and Niekum, Scott and Topcu, Ufuk},
  booktitle={Proc.\  AAAI},
 pages     = {2669--2678},
  year={2018}
}

@inproceedings{guo2020designing,
  title={Designing context-sensitive norm inverse reinforcement learning framework for norm-compliant autonomous agents},
  author={Guo, Yue and Wang, Boshi and Hughes, Dana and Lewis, Michael and Sycara, Katia},
  booktitle={2020 29th IEEE International Conference on Robot and Human Interactive Communication (RO-MAN)},
  pages={618--625},
  year={2020},
  organization={IEEE}
}

@inproceedings{FT2014,
  author    = {Jie Fu and
               Ufuk Topcu},
  editor    = {Dieter Fox and
               Lydia E. Kavraki and
               Hanna Kurniawati},
  title     = {Probably Approximately Correct {MDP} Learning and Control With Temporal
               Logic Constraints},
  booktitle = {Robotics: Science and Systems X, University of California, Berkeley,
               USA, July 12-16, 2014},
  year      = {2014}
}

@inproceedings{HAK2020,
  author    = {Mohammadhosein Hasanbeig and
               Alessandro Abate and
               Daniel Kroening},
  title     = {Cautious Reinforcement Learning with Logical Constraints},
  booktitle = {Proceedings of the 19th International Conference on Autonomous Agents
               and Multiagent Systems, {AAMAS} '20, Auckland, New Zealand, May 9-13,
               2020},
  pages     = {483--491},
  year      = {2020}
}

@inproceedings{KS2018,
  title={Norm conflict resolution in stochastic domains},
  author={Kasenberg, Daniel and Scheutz, Matthias},
  booktitle={Proceedings of the AAAI Conference on Artificial Intelligence},
  volume={32},
  number={1},
  year={2018}
}

@inproceedings{PZOS2022,
  title={MORAL: Aligning AI with Human Norms through Multi-Objective Reinforced Active Learning},
  author={Peschl, Markus and Zgonnikov, Arkady and Oliehoek, Frans A and Siebert, Luciano C},
  booktitle={Proceedings of the 21st International Conference on Autonomous Agents and Multiagent Systems},
  pages={1038--1046},
  year={2022}
}

@inproceedings{NBM2019,
  title={Teaching AI agents ethical values using reinforcement learning and policy orchestration},
  author={Noothigattu, Ritesh and Bouneffouf, Djallel and Mattei, Nicholas and Chandra, Rachita and Madan, Piyush and Varshney, Kush R and Campbell, Murray and Singh, Moninder and Rossi, Francesca},
  booktitle = {Proc. of IJCAI: 28th International Joint Conference on
        Artificial Intelligence},
  year      = {2019},
  publisher = {ijcai.org},
}

@book{andrighetto2013normative,
  title={Normative Muti-Agent Systems},
  author={Andrighetto, Giulia and Governatori, Guido and Noriega, Pablo and van der Torre, Leender WN},
  year={2013},
  publisher={Schloss Dagstuhl, Leibniz-Zentrum für Informatik}
}

@article{tomic2019robby,
  author       = {Stevan Tomic and
                  Federico Pecora and
                  Alessandro Saffiotti},
  title        = {Robby is Not a Robber (anymore): On the Use of Institutions for Learning
                  Normative Behavior},
  journal      = {CoRR},
  volume       = {abs/1908.02138},
  year         = {2019},
  url          = {http://arxiv.org/abs/1908.02138},
  eprinttype    = {arXiv},
  eprint       = {1908.02138},
  bibsource    = {dblp computer science bibliography, https://dblp.org}
}

@inproceedings{li2015reinforcement,
  author       = {Jiaqi Li and
                  Felipe Meneguzzi and
                  Moser Silva Fagundes and
                  Brian Logan},
  editor       = {Virginia Dignum and
                  Pablo Noriega and
                  Murat Sensoy and
                  Jaime Sim{\~{a}}o Sichman},
  title        = {Reinforcement Learning of Normative Monitoring Intensities},
  booktitle    = {Coordination, Organizations, Institutions, and Normes in Agent Systems
                  {XI} - {COIN} 2015 International Workshops, COIN@AAMAS, Istanbul,
                  Turkey, May 4, 2015, COIN@IJCAI, Buenos Aires, Argentina, July 26,
                  2015, Revised Selected Papers},
  series       = {Lecture Notes in Computer Science},
  volume       = {9628},
  pages        = {209--223},
  publisher    = {Springer},
  year         = {2015},
  url          = {https://doi.org/10.1007/978-3-319-42691-4\_12},
  doi          = {10.1007/978-3-319-42691-4\_12},
  bibsource    = {dblp computer science bibliography, https://dblp.org}
}

@inproceedings{panagiotidi2013towards,
  author       = {Sofia Panagiotidi and
                  Sergio {\'{A}}lvarez{-}Napagao and
                  Javier V{\'{a}}zquez{-}Salceda},
  editor       = {Tina Balke and
                  Frank Dignum and
                  M. Birna van Riemsdijk and
                  Amit K. Chopra},
  title        = {Towards the Norm-Aware Agent: Bridging the Gap Between Deontic Specifications
                  and Practical Mechanisms for Norm Monitoring and Norm-Aware Planning},
  booktitle    = {Coordination, Organizations, Institutions, and Norms in Agent Systems
                  {IX} - {COIN} 2013 International Workshops, COIN@AAMAS, St. Paul,
                  MN, USA, May 6, 2013, COIN@PRIMA, Dunedin, New Zealand, December 3,
                  2013, Revised Selected Papers},
  series       = {Lecture Notes in Computer Science},
  volume       = {8386},
  pages        = {346--363},
  publisher    = {Springer},
  year         = {2013},
  url          = {https://doi.org/10.1007/978-3-319-07314-9\_19},
  doi          = {10.1007/978-3-319-07314-9\_19},
  bibsource    = {dblp computer science bibliography, https://dblp.org}
}

@inproceedings{aldewereld2010making,
  author       = {Huib Aldewereld and
                  Sergio {\'{A}}lvarez{-}Napagao and
                  Frank Dignum and
                  Javier V{\'{a}}zquez{-}Salceda},
  editor       = {Wiebe van der Hoek and
                  Gal A. Kaminka and
                  Yves Lesp{\'{e}}rance and
                  Michael Luck and
                  Sandip Sen},
  title        = {Making norms concrete},
  booktitle    = {9th International Conference on Autonomous Agents and Multiagent Systems
                  {(AAMAS} 2010), Toronto, Canada, May 10-14, 2010, Volume 1-3},
  pages        = {807--814},
  publisher    = {{IFAAMAS}},
  year         = {2010},
  url          = {https://dl.acm.org/citation.cfm?id=1838314},
  bibsource    = {dblp computer science bibliography, https://dblp.org}
}

@article{kadir2024norm,
  title={Norm Augmented Reinforcement Learning Agents With Synthesized Normative Rules: A Proposed Normative Agent Framework},
  author={Kadir, Mohd Rashdan Abdul and Selamat, Ali and Krejcar, Ondrej},
  journal={Journal of Cases on Information Technology (JCIT)},
  volume={26},
  number={1},
  pages={1--34},
  year={2024},
  publisher={IGI Global}
}

@article{amgoud2009decision,
    title = {Using arguments for making and explaining decisions},
    journal = {Artificial Intelligence},
    volume = {173},
    number = {3},
    pages = {413--436},
    year = {2009},
    author = {Amgoud, Leila and Prade, Henri}
}

@inproceedings{malle2017networks,
  title={Networks of social and moral norms in human and robot agents},
  author={Malle, Bertram F and Scheutz, Matthias and Austerweil, Joseph L},
  booktitle={A world with robots: International Conference on Robot Ethics: ICRE 2015},
  pages={3--17},
  year={2017},
  organization={Springer}
}

@article{boman1999norms,
  title={Norms in artificial decision making},
  author={Boman, Magnus},
  journal={Artificial Intelligence and Law},
  volume={7},
  number={1},
  pages={17--35},
  year={1999},
  publisher={Springer}
}

@article{joseph2010deductive,
  title={Deductive coherence and norm adoption},
  author={Joseph, Sindhu and Sierra, Carles and Schorlemmer, Marco and Dellunde, Pilar},
  journal={Logic Journal of IGPL},
  volume={18},
  number={1},
  pages={118--156},
  year={2010},
  publisher={Oxford University Press}
}

@inproceedings{cardoso2009adaptive,
  title={Adaptive deterrence sanctions in a normative framework},
  author={Cardoso, Henrique Lopes and Oliveira, Eugénio},
  booktitle={2009 IEEE/WIC/ACM International Joint Conference on Web Intelligence and Intelligent Agent Technology},
  volume={2},
  pages={36--43},
  year={2009},
  organization={IEEE}
}

@inproceedings{meneguzzi2009norm,
  title={Norm-based behaviour modification in BDI agents.},
  author={Meneguzzi, Felipe Rech and Luck, Michael},
  booktitle={AAMAS (1)},
  pages={177--184},
  year={2009}
}

@article{lopez2006normative,
  title={A normative framework for agent-based systems},
  author={L\'opez, Fabiola L\'opez y and Luck, Michael and d’Inverno, Mark},
  journal={Computational \& Mathematical Organization Theory},
  volume={12},
  number={2},
  pages={227--250},
  year={2006},
  publisher={Springer}
}

@inproceedings{andrighetto2007immergence,
  title={On the immergence of norms: a normative agent architecture},
  author={Andrighetto, Giulia and Campenn\`i, Marco and Conte, Rosaria and Paolucci, Mario},
  booktitle={AAAI Fall Symposium: Emergent Agents and Socialities},
  year={2007}
}

@article{boella2006game,
  title={A game theoretic approach to contracts in multiagent systems},
  author={Boella, Guido and van der Torre, Leendert WN},
  journal={IEEE Transactions on Systems, Man, and Cybernetics, Part C (Applications and Reviews)},
  volume={36},
  number={1},
  pages={68--79},
  year={2006},
  publisher={IEEE}
}

@inproceedings{kollingbaum2003noa,
  title={NoA-a normative agent architecture},
  author={Kollingbaum, Martin J and Norman, Timothy J},
  booktitle={IJCAI},
  pages={1465--1466},
  year={2003}
}

@inproceedings{dignum2000towards,
  title={Towards socially sophisticated BDI agents},
  author={Dignum, Frank and Morley, David and Sonenberg, Elizabeth A and Cavedon, Lawrence},
  booktitle={Proceedings fourth international conference on multiagent systems},
  pages={111--118},
  year={2000},
  organization={IEEE}
}

@inproceedings{boella12004agent,
  title={An Agent-Oriented Ontology of Social Reality},
  author={Boella, Guido},
  booktitle={Formal Ontology in Information Systems: Proceedings of the Third International Conference (FOIS-2004)},
  pages={199},
  year={2004},
  organization={IOS Press}
}

@inproceedings{castelfranchi1999deliberative,
  title={Deliberative normative agents: Principles and architecture},
  author={Castelfranchi, Cristiano and Dignum, Frank and Jonker, Catholijn M and Treur, Jan},
  booktitle={International workshop on agent theories, architectures, and languages},
  pages={364--378},
  year={1999},
  organization={Springer}
}

@inproceedings{georgeff1991modeling,
  title={Modeling rational agents within a BDI-architecture},
  author={Georgeff, M and Rao, A},
  booktitle={Proc. 2nd Int. Conf. on Knowledge Representation and Reasoning (KR’91). Morgan Kaufmann},
  pages={473--484},
  year={1991},
  organization={of}
}

@inproceedings{pardo2023advanced,
  title={Advanced Intelligent Systems and Reasoning: Standardization, Experimentation, Explanation},
  author={Pardo, Pere and van der Torre, Leendert WN and Yu, Liuwen},
  booktitle={Logics for New Generation AI (LNGAI2023)},
  year={2023},
  organization={College Publications, London, United Kingdom}
}

@inproceedings{cabrio2018five,
  title={Five years of argument mining: A data-driven analysis.},
  author={Cabrio, Elena and Villata, Serena},
  booktitle={IJCAI},
  volume={18},
  pages={5427--5433},
  year={2018}
}

@article{blackburn1997combine,
  title={Why combine logics?},
  author={Blackburn, Patrick and de Rijke, Maarten},
  journal={Studia Logica},
  volume={59},
  number={1},
  pages={5--27},
  year={1997},
  publisher={Springer}
}

@article{sviridova2024mining,
  title={Mining implicit arguments for reasoning: A survey},
  author={Sviridova, Ekaterina and Cabrio, Elena and Villata, Serena},
  journal={Argument \& Computation},
  pages={19462174251344764},
  year={2024},
  publisher={SAGE Publications Sage UK: London, England}
}

@inproceedings{yu2025bdi,
  title={The A-BDI Metamodel for Human-Level AI: Argumentation as Balancing, Dialogue and Inference},
  author={Yu, Liuwen and van der Torre, Leendert WN},
  booktitle={International Conference on Logic and Argumentation},
  pages={361--379},
  year={2025},
  organization={Springer}
}

@article{fagundes2016design,
  title={Design and evaluation of norm-aware agents based on Normative Markov Decision Processes},
  author={Fagundes, Moser Silva and Ossowski, Sascha and Cerquides, Jes{\'u}s and Noriega, Pablo},
  journal={International Journal of Approximate Reasoning},
  volume={78},
  pages={33--61},
  year={2016},
  publisher={Elsevier}
}

@inproceedings{boella2004groups,
  title={Groups as agents with mental attitudes},
  author={Boella, Guido and van der Torre, Leendert WN and others},
  booktitle={International Conference on Autonomous Agents: Proceedings of the Third International Joint Conference on Autonomous Agents and Multiagent Systems-},
  volume={2},
  pages={964--971},
  year={2004}
}

@article{boella2004structuring,
  title={Structuring organizations by means of roles using the agent metaphor},
  author={Boella, Guido and van der Torre, Leendert WN},
  journal={dagli Oggetti agli Agenti},
  pages={93},
  year={2004}
}

@inproceedings{boella2003attributing,
  title={Attributing mental attitudes to normative systems},
  author={Boella, Guido and van der Torre, Leendert WN},
  booktitle={Proceedings of the second international joint conference on Autonomous agents and multiagent systems},
  pages={942--943},
  year={2003}
}

@inproceedings{boella2005constitutive,
  title={Constitutive norms in the design of normative multiagent systems},
  author={Boella, Guido and van der Torre, Leendert WN},
  booktitle={International Workshop on Computational Logic in Multi-Agent Systems},
  pages={303--319},
  year={2005},
  organization={Springer}
}

@inproceedings{boella2004contracts,
  title={Contracts as legal institutions in organizations of autonomous agents},
  author={Boella, Guido and van der Torre, Leendert WN and others},
  booktitle={AAMAS},
  volume={4},
  pages={948--955},
  year={2004}
}

@inproceedings{boella2003local,
  title={Local policies for the control of virtual communities},
  author={Boella, Guido and van der Torre, Leendert WN},
  booktitle={Proceedings IEEE/WIC International Conference on Web Intelligence (WI 2003)},
  pages={161--167},
  year={2003},
  organization={IEEE}
}

@article{boella2002attributing,
  title={Attributing Mental Attitudes to Roles: The Agent Metaphor Applied to e-Trade Organizations},
  author={Boella, Guido and van der Torre, Leendert WN},
  year={2002}
}

@article{aarts2003silence,
  title={The silence of the library: environment, situational norm, and social behavior.},
  author={Aarts, Henk and Dijksterhuis, Ap},
  journal={Journal of personality and social psychology},
  volume={84},
  number={1},
  pages={18},
  year={2003},
  publisher={American Psychological Association}
}

@book{wooldridge2009introduction,
  title={An introduction to multiagent systems},
  author={Wooldridge, Michael},
  year={2009},
  publisher={John wiley \& sons}
}

@article{russell1995modern,
  title={A modern approach},
  author={Russell, Stuart and Norvig, Peter and Intelligence, Artificial},
  journal={Artificial Intelligence. Prentice-Hall, Egnlewood Cliffs},
  volume={25},
  number={27},
  pages={79--80},
  year={1995}
}

@book{nilsson2009quest,
  title={The quest for artificial intelligence},
  author={Nilsson, Nils J},
  year={2009},
  publisher={Cambridge University Press}
}

@inproceedings{tomic2020learning,
  title={Learning Normative Behaviors Through Abstraction.},
  author={Tomic, Stevan and Pecora, Federico and Saffiotti, Alessandro},
  booktitle={ECAI},
  pages={1547--1554},
  year={2020}
}

@article{rizzo2018qualitative,
  title={A qualitative investigation of the degree of explainability of defeasible argumentation and non-monotonic fuzzy reasoning},
  author={Rizzo, Lucas and Longo, Luca},
  year={2018},
  publisher={Technological University Dublin}
}

@incollection{fan2014computing,
  title={On computing explanations in abstract argumentation},
  author={Fan, Xiuyi and Toni, Francesca},
  booktitle={ECAI 2014},
  pages={1005--1006},
  year={2014},
  publisher={IOS Press}
}

@book{lewis2013counterfactuals,
  title={Counterfactuals},
  author={Lewis, David},
  year={2013},
  publisher={John Wiley \& Sons}
}

@misc{brown2020language,
      title={Language Models are Few-Shot Learners}, 
      author={Tom B. Brown and Benjamin Mann and Nick Ryder and Melanie Subbiah and Jared Kaplan and Prafulla Dhariwal and Arvind Neelakantan and Pranav Shyam and Girish Sastry and Amanda Askell and Sandhini Agarwal and Ariel Herbert-Voss and Gretchen Krueger and Tom Henighan and Rewon Child and Aditya Ramesh and Daniel M. Ziegler and Jeffrey Wu and Clemens Winter and Christopher Hesse and Mark Chen and Eric Sigler and Mateusz Litwin and Scott Gray and Benjamin Chess and Jack Clark and Christopher Berner and Sam McCandlish and Alec Radford and Ilya Sutskever and Dario Amodei},
      year={2020},
      eprint={2005.14165},
      archivePrefix={arXiv},
      primaryClass={cs.CL},
      url={https://arxiv.org/abs/2005.14165}, 
}

@misc{misc_car_evaluation_19,
  title        = {{Car Evaluation}},
  author       = {Bohanec,Marko},
  year         = {1997},
  howpublished = {UCI Machine Learning Repository},
  note         = {{DOI}: https://doi.org/10.24432/C5JP48}
}

@article{alcaraz2025norm,
  title={Norm Avoidance and Reinforcement Learning: Definitions and Analysis},
  author={Alcaraz, Beno\^it and Neufeld, Emery A and van der Torre, Leendert WN},
  journal={The 17th International Conference on Deontic Logic and normative systems (DEON 2025).},
  pages={1},
  year={2025}
}

@inproceedings{alcaraz2025star,
  title={An A-Star Algorithm for Argumentative Rule Extraction},
  author={Alcaraz, Beno\^it and Kaliski, Adam and Leturc, Christopher},
  booktitle={Proceedings of the 17th International Conference on Agents and Artificial Intelligence},
  volume={2},
  pages={91--101},
  year={2025}
}

@article{bex2021xaiargs,
author = {Borg, Annemarie and Bex, Floris},
year = {2021},
month = {01},
pages = {1-1},
title = {A Basic Framework for Explanations in Argumentation},
volume = {PP},
journal = {IEEE Intelligent Systems},
doi = {10.1109/MIS.2021.3053102}
}

@article{boella2009meta,
  title={Meta-argumentation modelling I: Methodology and techniques},
  author={Boella, Guido and Gabbay, Dov M and van der Torre, Leendert WN and Villata, Serena},
  journal={Studia Logica},
  volume={93},
  pages={297--355},
  year={2009},
  publisher={Springer}
}

@inproceedings{doutre2023visual,
  title={Visual explanations for defence in abstract argumentation},
  author={Doutre, Sylvie and Duchatelle, Théo and Lagasquie-Schiex, Marie-Christine},
  booktitle={International Conference on Autonomous Agents and Multiagent Systems (AAMAS)},
  pages={2346--2348},
  year={2023},
  organization={ACM}
}

@book{horridge2011justification,
  title={Justification based explanation in ontologies},
  author={Horridge, Matthew},
  year={2011},
  publisher={The University of Manchester (United Kingdom)}
}

@article{liao2021representation,
  title={Representation, justification, and explanation in a value-driven agent: an argumentation-based approach},
  author={Liao, Beishui and Anderson, Michael and Anderson, Susan Leigh},
  journal={AI and Ethics},
  volume={1},
  number={1},
  pages={5--19},
  year={2021},
  publisher={Springer}
}

@inproceedings{nourbakhsh2024feature,
  title={Feature Generation Using LLMs: An Evolutionary Algorithm Approach},
  author={Nourbakhsh, Aria and Alcaraz, Beno\^it and Schommer, Christoph},
  booktitle={International Workshop on Causality, Agents and Large Models},
  pages={48--64},
  year={2024},
  organization={Springer}
}

@inproceedings{yu2020principle,
  title={A principle-based approach to bipolar argumentation},
  author={Yu, Liuwen and van der Torre, Leendert WN},
  booktitle={NMR 2020 Workshop Notes},
  volume={227},
  year={2020}
}

@article{awad2018moral,
  title={The moral machine experiment},
  author={Awad, Edmond and Dsouza, Sohan and Kim, Richard and Schulz, Jonathan and Henrich, Joseph and Shariff, Azim and Bonnefon, Jean-Fran{\c{c}}ois and Rahwan, Iyad},
  journal={Nature},
  volume={563},
  number={7729},
  pages={59--64},
  year={2018},
  publisher={Nature Publishing Group UK London}
}

@misc{bcw_17,
  author       = {Wolberg, William and Mangasarian, Olvi and Street, Nick and Street, W.},
  title        = {{Breast Cancer Wisconsin (Diagnostic)}},
  year         = {1993},
  howpublished = {UCI Machine Learning Repository},
  note         = {{DOI}: https://doi.org/10.24432/C5DW2B}
}

@misc{iris_53,
  author       = {Fisher, R. A.},
  title        = {{Iris}},
  year         = {1936},
  howpublished = {UCI Machine Learning Repository},
  note         = {{DOI}: https://doi.org/10.24432/C56C76}
}

@misc{wine_109,
  author       = {Aeberhard, Stefan and Forina, M.},
  title        = {{Wine}},
  year         = {1992},
  howpublished = {UCI Machine Learning Repository},
  note         = {{DOI}: https://doi.org/10.24432/C5PC7J}
}

@article{cortes1995support,
  title={Support-vector networks},
  author={Cortes, Corinna and Vapnik, Vladimir},
  journal={Machine learning},
  volume={20},
  pages={273--297},
  year={1995},
  publisher={Springer}
}

@misc{thyroid_915,
  author       = {Borzooei, Shiva and Tarokhian, Aidin},
  title        = {{Differentiated Thyroid Cancer Recurrence}},
  year         = {2023},
  howpublished = {UCI Machine Learning Repository},
  note         = {{DOI}: https://doi.org/10.24432/C5632J}
}

@misc{voting_105,
  title        = {{Congressional Voting Records}},
  year         = {1987},
  howpublished = {UCI Machine Learning Repository},
  note         = {{DOI}: https://doi.org/10.24432/C5C01P}
}

@misc{hdc_45,
  author       = {Janosi, Andras and Steinbrunn, William and Pfisterer, Matthias and Detrano, Robert},
  title        = {{Heart Disease}},
  year         = {1989},
  howpublished = {UCI Machine Learning Repository},
  note         = {{DOI}: https://doi.org/10.24432/C52P4X}
}

@article{murphy1994uci,
  title={UCI repository of machine learning databases},
  author={Murphy, Patrick M},
  journal={http://www. ics. uci. edu/\~{} mlearn/MLRepository. html},
  year={1994},
  publisher={University of California, Department of Information and Computer Science}
}

@article{adadi2018peeking,
  title={Peeking inside the black-box: a survey on explainable artificial intelligence (XAI)},
  author={Adadi, Amina and Berrada, Mohammed},
  journal={IEEE access},
  volume={6},
  pages={52138--52160},
  year={2018},
  publisher={IEEE}
}

@article{bagallo1990boolean,
  title={Boolean feature discovery in empirical learning},
  author={Bagallo, Giulia and Haussler, David},
  journal={Machine learning},
  volume={5},
  pages={71--99},
  year={1990},
  publisher={Springer}
}

@incollection{cohen1995fast,
  title={Fast effective rule induction},
  author={Cohen, William W},
  booktitle={Machine learning proceedings 1995},
  pages={115--123},
  year={1995},
  publisher={Elsevier}
}

@article{quinlan1986induction,
  title={Induction of decision trees},
  author={Quinlan, J Ross},
  journal={Machine learning},
  volume={1},
  pages={81--106},
  year={1986},
  publisher={Springer}
}

@incollection{furnkranz1994incremental,
  title={Incremental reduced error pruning},
  author={F{\"u}rnkranz, Johannes and Widmer, Gerhard},
  booktitle={Machine learning proceedings 1994},
  pages={70--77},
  year={1994},
  publisher={Elsevier}
}

@article{hein2018interpretable,
  title={Interpretable policies for reinforcement learning by genetic programming},
  author={Hein, Daniel and Udluft, Steffen and Runkler, Thomas A},
  journal={Engineering Applications of Artificial Intelligence},
  volume={76},
  pages={158--169},
  year={2018},
  publisher={Elsevier}
}

@inproceedings{liu2018toward,
  title={Toward interpretable deep reinforcement learning with linear model u-trees},
  author={Liu, Guiliang and Schulte, Oliver and Zhu, Wang and Li, Qingcan},
  booktitle={Joint European Conference on Machine Learning and Knowledge Discovery in Databases},
  pages={414--429},
  year={2018},
  organization={Springer}
}

@inproceedings{liu2000extended,
  title={An extended genetic rule induction algorithm},
  author={Liu, J Juan and Kwok, J Tin-Yau},
  booktitle={Proceedings of the 2000 Congress on Evolutionary Computation. CEC00 (Cat. No. 00TH8512)},
  volume={1},
  pages={458--463},
  year={2000},
  organization={IEEE}
}

@inproceedings{puiutta2020explainable,
  title={Explainable reinforcement learning: A survey},
  author={Puiutta, Erika and Veith, Eric MSP},
  booktitle={International cross-domain conference for machine learning and knowledge extraction},
  pages={77--95},
  year={2020},
  organization={Springer}
}

@inproceedings{quinlan1987generating,
  title={Generating production rules from decision trees.},
  author={Quinlan, J Ross},
  booktitle={ijcai},
  volume={87},
  pages={304--307},
  year={1987},
  organization={Citeseer}
}

@book{quinlan2014c4,
  title={C4. 5: programs for machine learning},
  author={Quinlan, J Ross},
  year={2014},
  publisher={Elsevier}
}

@inproceedings{sabbatini2021design,
  title={On the design of PSyKE: a platform for symbolic knowledge extraction},
  author={Sabbatini, Federico and Ciatto, Giovanni and Calegari, Roberta and Omicini, Andrea and others},
  booktitle={CEUR WORKSHOP PROCEEDINGS},
  volume={2963},
  pages={29--48},
  year={2021},
  organization={Sun SITE Central Europe, RWTH Aachen University}
}

@inproceedings{selbst2018meaningful,
  title={“Meaningful information” and the right to explanation},
  author={Selbst, Andrew and Powles, Julia},
  booktitle={conference on fairness, accountability and transparency},
  pages={48--48},
  year={2018},
  organization={PMLR}
}

@article{szegedy2013intriguing,
  title={Intriguing properties of neural networks},
  author={Szegedy, Christian and Zaremba, Wojciech and Sutskever, Ilya and Bruna, Joan and Erhan, Dumitru and Goodfellow, Ian and Fergus, Rob},
  journal={arXiv preprint arXiv:1312.6199},
  year={2013}
}

@inproceedings{tan2018distill,
  title={Distill-and-compare: Auditing black-box models using transparent model distillation},
  author={Tan, Sarah and Caruana, Rich and Hooker, Giles and Lou, Yin},
  booktitle={Proceedings of the 2018 AAAI/ACM Conference on AI, Ethics, and Society},
  pages={303--310},
  year={2018}
}

@inproceedings{venturini1993sia,
  title={SIA: a supervised inductive algorithm with genetic search for learning attributes based concepts},
  author={Venturini, Gilles},
  booktitle={European conference on machine learning},
  pages={280--296},
  year={1993},
  organization={Springer}
}

@inproceedings{verma2018programmatically,
  title={Programmatically interpretable reinforcement learning},
  author={Verma, Abhinav and Murali, Vijayaraghavan and Singh, Rishabh and Kohli, Pushmeet and Chaudhuri, Swarat},
  booktitle={International Conference on Machine Learning},
  pages={5045--5054},
  year={2018},
  organization={PMLR}
}

@article{wachter2017counterfactual,
  title={Counterfactual explanations without opening the black box: Automated decisions and the GDPR},
  author={Wachter, Sandra and Mittelstadt, Brent and Russell, Chris},
  journal={Harv. JL \& Tech.},
  volume={31},
  pages={841},
  year={2017},
  publisher={HeinOnline}
}

@inproceedings{weiss1991reduced,
  title={Reduced Complexity Rule Induction.},
  author={Weiss, Sholom M and Indurkhya, Nitin},
  booktitle={IJCAI},
  pages={678--684},
  year={1991}
}

@inproceedings{zahavy2016graying,
  title={Graying the black box: Understanding dqns},
  author={Zahavy, Tom and Ben-Zrihem, Nir and Mannor, Shie},
  booktitle={International Conference on Machine Learning},
  pages={1899--1908},
  year={2016},
  organization={PMLR}
}

@inproceedings{alcaraz2024assessing,
  title={Assessing the Robustness of LLMs in Predicting Supports and Attacks},
  author={Alcaraz, Beno\^it and Nourbakhsh, Aria and Yu, Liuwen},
  booktitle={International Workshop on Causality, Agents and Large Models},
  pages={88--93},
  year={2024},
  organization={Springer}
}

@article{botti2025agentic,
  title={Agentic AI and Multiagentic: Are We Reinventing the Wheel?},
  author={Botti, Vicent},
  journal={arXiv preprint arXiv:2506.01463},
  year={2025}
}

@book{perelman2012justice,
  title={Justice, law, and argument: Essays on moral and legal reasoning},
  author={Perelman, Cha\"im},
  volume={142},
  year={2012},
  publisher={Springer Science \& Business Media}
}

@book{pascal1670pensees,
  author    = {Pascal, Blaise},
  title     = {Pensées},
  year      = {1670},
  publisher = {Penguin Classics},
  editor    = {Roger Ariew},
  note      = {Fragment 233 (Pascal's Wager)}
}


%% file: slr.bib
@article{sutton1999reinforcement,
  title={Reinforcement learning},
  author={Sutton, Richard S and Barto, Andrew G and others},
  journal={Journal of Cognitive Neuroscience},
  volume={11},
  number={1},
  pages={126--134},
  year={1999}
}

@article{ferraro2021nlp,
  title={NLP Techniques for Normative Mining.},
  author={Ferraro, Gabriela and Lam, Ho-Pun},
  journal={FLAP},
  volume={8},
  number={4},
  pages={941--974},
  year={2021}
}

@inproceedings{cranefield2021identifying,
  title={Identifying Norms from Observation Using MCMC Sampling.},
  author={Cranefield, Stephen and Dhiman, Ashish},
  booktitle={IJCAI},
  pages={118--124},
  year={2021}
}

@article{oldenburg2024learning,
  title={Learning and sustaining shared normative systems via bayesian rule induction in markov games},
  author={Oldenburg, Ninell and Zhi-Xuan, Tan},
  journal={arXiv preprint arXiv:2402.13399},
  year={2024}
}

@article{fung2022normsage,
  title={Normsage: Multi-lingual multi-cultural norm discovery from conversations on-the-fly},
  author={Fung, Yi R and Chakraborty, Tuhin and Guo, Hao and Rambow, Owen and Muresan, Smaranda and Ji, Heng},
  journal={arXiv preprint arXiv:2210.08604},
  year={2022}
}

@article{moghimifar2023normmark,
  title={NormMark: A weakly supervised Markov model for socio-cultural norm discovery},
  author={Moghimifar, Farhad and Qu, Shilin and Wu, Tongtong and Li, Yuan-Fang and Haffari, Gholamreza},
  journal={arXiv preprint arXiv:2305.16598},
  year={2023}
}

@inproceedings{tan2019s,
  title={That’s mine! learning ownership relations and norms for robots},
  author={Tan, Zhi-Xuan and Brawer, Jake and Scassellati, Brian},
  booktitle={Proceedings of the AAAI conference on artificial intelligence},
  volume={33},
  number={01},
  pages={8058--8065},
  year={2019}
}

@article{corapi2011normative,
  title={Normative design using inductive learning},
  author={Corapi, Domenico and Russo, Alessandra and De Vos, Marina and Padget, Julian and Satoh, Ken},
  journal={Theory and Practice of Logic Programming},
  volume={11},
  number={4-5},
  pages={783--799},
  year={2011},
  publisher={Cambridge University Press}
}

@inproceedings{mahmoud2012norms,
  title={A norms mining approach to norms detection in multi-agent systems},
  author={Mahmoud, Moamin A and Ahmad, Mohd Sharifuddin and Ahmad, Azhana and Yusoff, Mohd Zaliman Mohd and Mustapha, Aida},
  booktitle={2012 International Conference on Computer \& Information Science (ICCIS)},
  volume={1},
  pages={458--463},
  year={2012},
  organization={IEEE}
}

@article{mahmoud2016development,
  title={Development and implementation of a technique for norms-adaptable agents in open multi-agent communities},
  author={Mahmoud, Moamin A and Ahmad, Mohd Sharifuddin and Mohd Yusoff, Mohd Zaliman},
  journal={Journal of Systems Science and Complexity},
  volume={29},
  pages={1519--1537},
  year={2016},
  publisher={Springer}
}

@inproceedings{mahmoud2012norm,
  title={Norms detection and assimilation in multi-agent systems: a conceptual approach},
  author={Mahmoud, Moamin A and Ahmad, Mohd Sharifuddin and Ahmad, Azhana and Mohd Yusoff, Mohd Zaliman and Mustapha, Aida},
  booktitle={Knowledge Technology: Third Knowledge Technology Week, KTW 2011, Kajang, Malaysia, July 18-22, 2011. Revised Selected Papers},
  pages={226--233},
  year={2012},
  organization={Springer}
}

@article{savarimuthu2010norm,
  title={Norm identification in multi-agent societies},
  author={Savarimuthu, Bastin Tony Roy and Cranefield, Stephen and Purvis, Maryam A and Purvis, Martin K},
  journal={Information Sciences},
  year={2010},
  publisher={University of Otago}
}

@inproceedings{mahmoud2016norm,
  title={A norm assimilation approach for multi-agent systems in heterogeneous communities},
  author={Mahmoud, Moamin A and Ahmad, Mohd Sharifuddin and Yusoff, Mohd Zaliman M},
  booktitle={Intelligent Information and Database Systems: 8th Asian Conference, ACIIDS 2016, Da Nang, Vietnam, March 14--16, 2016, Proceedings, Part I 8},
  pages={354--363},
  year={2016},
  organization={Springer}
}

@inproceedings{mahmoud2013potential,
  title={Potential norms detection in social agent societies},
  author={Mahmoud, Moamin A and Mustapha, Aida and Ahmad, Mohd Sharifuddin and Ahmad, Azhana and Yusoff, Mohd Zaliman M and Hamid, Nurzeatul Hamimah Abdul},
  booktitle={Distributed Computing and Artificial Intelligence: 10th International Conference},
  pages={419--428},
  year={2013},
  organization={Springer}
}

@inproceedings{savarimuthu2010data,
  title={A data mining approach to identify obligation norms in agent societies},
  author={Savarimuthu, Bastin Tony Roy and Cranefield, Stephen and Purvis, Maryam and Purvis, Martin},
  booktitle={Agents and Data Mining Interaction: 6th International Workshop on Agents and Data Mining Interaction, ADMI 2010, Toronto, ON, Canada, May 11, 2010, Revised Selected Papers 6},
  pages={43--58},
  year={2010},
  organization={Springer}
}

@inproceedings{riad2021run,
  title={Run-time norms synthesis in multi-objective multi-agent systems},
  author={Riad, Maha and Golpayegani, Fatemeh},
  booktitle={International workshop on coordination, organizations, institutions, norms, and ethics for governance of multi-agent systems},
  pages={78--93},
  year={2021},
  organization={Springer}
}

@article{savarimuthu2013identifying,
  title={Identifying prohibition norms in agent societies},
  author={Savarimuthu, Bastin Tony Roy and Cranefield, Stephen and Purvis, Maryam A and Purvis, Martin K},
  journal={Artificial intelligence and law},
  volume={21},
  pages={1--46},
  year={2013},
  publisher={Springer}
}

@inproceedings{gao2014extracting,
  title={Extracting normative relationships from business contracts},
  author={Gao, Xibin and Singh, Munindar P},
  booktitle={Proceedings of the 2014 international conference on Autonomous agents and multi-agent systems},
  pages={101--108},
  year={2014}
}

@incollection{cranefield2016bayesian,
  title={A Bayesian approach to norm identification},
  author={Cranefield, Stephen and Meneguzzi, Felipe and Oren, Nir and Savarimuthu, Bastin Tony Roy},
  booktitle={ECAI 2016},
  pages={622--629},
  year={2016},
  publisher={Ios Press}
}

@inproceedings{campos2010case,
  title={A case-based reasoning approach for norm adaptation},
  author={Campos, Jordi and L{\'o}pez-S{\'a}nchez, Maite and Esteva, Marc},
  booktitle={Hybrid Artificial Intelligence Systems: 5th International Conference, HAIS 2010, San Sebasti{\'a}n, Spain, June 23-25, 2010. Proceedings, Part II 5},
  pages={168--176},
  year={2010},
  organization={Springer}
}

@inproceedings{avery2016externalization,
  title={Externalization of software behavior by the mining of norms},
  author={Avery, Daniel and Dam, Hoa Khanh and Savarimuthu, Bastin Tony Roy and Ghose, Aditya},
  booktitle={Proceedings of the 13th International Conference on Mining Software Repositories},
  pages={223--234},
  year={2016}
}

@inproceedings{sarathy2017learning,
  title={Learning behavioral norms in uncertain and changing contexts},
  author={Sarathy, Vasanth and Scheutz, Matthias and Malle, Bertram F},
  booktitle={2017 8th IEEE International Conference on Cognitive Infocommunications (CogInfoCom)},
  pages={000301--000306},
  year={2017},
  organization={IEEE}
}

@inproceedings{dam2015mining,
  title={Mining software repositories for social norms},
  author={Dam, Hoa Khanh and Savarimuthu, Bastin Tony Roy and Avery, Daniel and Ghose, Aditya},
  booktitle={2015 IEEE/ACM 37th IEEE International Conference on Software Engineering},
  volume={2},
  pages={627--630},
  year={2015},
  organization={IEEE}
}

@article{aires2017norm,
  title={Norm conflict identification in contracts},
  author={Aires, Jo{\~a}o Paulo and Pinheiro, Daniele and Lima, Vera Strube de and Meneguzzi, Felipe},
  journal={Artificial Intelligence and Law},
  volume={25},
  number={4},
  pages={397--428},
  year={2017},
  publisher={Springer}
}

@article{oren2020norm,
  title={Norm identification through plan recognition},
  author={Oren, Nir and Meneguzzi, Felipe},
  journal={arXiv preprint arXiv:2010.02627},
  year={2020}
}

@inproceedings{mahmoud2012semantics,
  title={The semantics of norms mining in multi-agent systems},
  author={Mahmoud, Moamin A and Ahmad, Mohd Sharifuddin and Ahmad, Azhana and Yusoff, Mohd Zaliman Mohd and Mustapha, Aida},
  booktitle={Computational Collective Intelligence. Technologies and Applications: 4th International Conference, ICCCI 2012, Ho Chi Minh City, Vietnam, November 28-30, 2012, Proceedings, Part I 4},
  pages={425--435},
  year={2012},
  organization={Springer}
}

@inproceedings{murali2021mining,
  title={Mining international political norms from the GDELT database},
  author={Murali, Rohit and Patnaik, Suravi and Cranefield, Stephen},
  booktitle={Coordination, Organizations, Institutions, Norms, and Ethics for Governance of Multi-Agent Systems XIII: International Workshops COIN 2017 and COINE 2020, Sao Paulo, Brazil, May 8-9, 2017 and Virtual Event, May 9, 2020, Revised Selected Papers},
  pages={35--56},
  year={2021},
  organization={Springer}
}

@article{alechina2018incentive,
  title={Incentive-compatible mechanisms for norm monitoring in open multi-agent systems},
  author={Alechina, Natasha and Halpern, Joseph Y and Kash, Ian A and Logan, Brian},
  journal={Journal of Artificial Intelligence Research},
  volume={62},
  pages={433--458},
  year={2018}
}

@inproceedings{dell2022complexity,
  title={The complexity of norm synthesis and revision},
  author={Dell’Anna, Davide and Alechina, Natasha and Dalpiaz, Fabiano and Dastani, Mehdi and L{\"o}ffler, Maarten and Logan, Brian},
  booktitle={International Workshop on Coordination, Organizations, Institutions, Norms, and Ethics for Governance of Multi-Agent Systems},
  pages={38--53},
  year={2022},
  organization={Springer}
}

@inproceedings{christelis2010exploiting,
  title={Exploiting domain knowledge to improve norm synthesis},
  author={Christelis, George and Rovatsos, Michael and Petrick, Ronald PA},
  booktitle={Proceedings of the 9th International Conference on Autonomous Agents and Multiagent Systems: volume 1-Volume 1},
  pages={831--838},
  year={2010},
  organization={Citeseer}
}

@inproceedings{morris2023agent,
  title={Agent-directed runtime norm synthesis},
  author={Morris-Martin, Andreasa and De Vos, Marina and Padget, Julian and Ray, Oliver},
  booktitle={Proceedings of the 2023 International Conference on Autonomous Agents and Multiagent Systems},
  pages={2271--2279},
  year={2023}
}

@inproceedings{morales2013automated,
  title={Automated synthesis of normative systems.},
  author={Morales, Javier and Lopez-Sanchez, Maite and Rodriguez-Aguilar, Juan A and Wooldridge, Michael J and Vasconcelos, Wamberto Weber},
  booktitle={AAMAS},
  volume={13},
  pages={483--490},
  year={2013}
}

@article{morales2015online,
  title={Online automated synthesis of compact normative systems},
  author={Morales, Javier and Lopez-Sanchez, Maite and Rodriguez-Aguilar, Juan A and Vasconcelos, Wamberto and Wooldridge, Michael},
  journal={ACM Transactions on Autonomous and Adaptive Systems (TAAS)},
  volume={10},
  number={1},
  pages={1--33},
  year={2015},
  publisher={ACM New York, NY, USA}
}

@article{liga2023fine,
  title={Fine-tuning GPT-3 for legal rule classification},
  author={Liga, Davide and Robaldo, Livio},
  journal={Computer Law \& Security Review},
  volume={51},
  pages={105864},
  year={2023},
  publisher={Elsevier}
}

@incollection{liga2022transfer,
  title={Transfer learning for deontic rule classification: The case study of the gdpr},
  author={Liga, Davide and Palmirani, Monica},
  booktitle={Legal Knowledge and Information Systems},
  pages={200--205},
  year={2022},
  publisher={IOS Press}
}
